\documentclass{article}
\usepackage{enumitem}

\usepackage[preprint]{neurips_2026}
\usepackage{graphicx}

\usepackage[utf8]{inputenc} 
\usepackage[T1]{fontenc}    
\usepackage{hyperref}       
\usepackage{url}            
\usepackage{booktabs}       
\usepackage{amsfonts}       
\usepackage{nicefrac}       
\usepackage{microtype}      
\usepackage{xcolor}         
\usepackage{multirow}
\usepackage{amsmath} 

\usepackage[most]{tcolorbox}
\usepackage{listings}
\usepackage{enumitem}

\newtcolorbox{promptbox}[1][]{
    colback=gray!10,        
    colframe=gray!60,       
    title=#1,               
    fonttitle=\bfseries,
    coltitle=black,
    boxrule=0.8pt,
    arc=3pt,                
    left=6pt,
    right=6pt,
    top=6pt,
    bottom=6pt,
    enhanced,
    breakable               
}

\title{MAS-Algorithm: A Workflow for Solving Algorithmic Programming Problems with a Multi-Agent System}

\author{%
  Yuliang Xu$^{1*}$, Xiang Xu$^{2}$\thanks{Equal contribution.}, Yao Wan$^{3}$,Hu Wei$^{2\dag}$ ,Tong Jia$^{1}$\thanks{Corresponding Author.}\\
        $^{1}$Peking University \hspace{1em} $^{2}$Alibaba Group \hspace{1em} 
        $^{3}$Huazhong University of Science and Technology \hspace{1em}  \\
        \texttt{yanquan.xx@alibaba-inc.com}, \texttt{jia.tong@pku.edu.cn}
}

\begin{document}

\maketitle

\begin{abstract}
Algorithmic problem solving serves as a rigorous testbed for evaluating structured reasoning in AI coding systems, as it directly reflects a model’s ability to perform structured reasoning in complex scenarios.
Existing approaches predominantly rely on model-centric strategies, such as architectural modifications and data scaling, which are costly and offer limited interpretability. Alternative methods leveraging external tools or prompting techniques (e.g., chain-of-thought) are often fragmented and lack a unified framework.
In this paper, we propose MAS-Algorithm, a systematic multi-agent workflow for algorithmic problem solving inspired by the practices of competitive programmers and algorithm engineers. Our framework decomposes the end-to-end solving process into modular stages, enabling structured reasoning, tool integration, and flexible coordination among agents. The design emphasizes both rigor and extensibility, allowing it to generalize across diverse problem types.
Experimental results on a self-constructed benchmark demonstrate consistent improvements across multiple Qwen series models, achieving an average gain of 6.48\% in acceptance rate. In contrast, parameter-efficient fine-tuning on the same data yields only a marginal improvement of 0.89\%. We further observe a 4.72\% gain on LiveCodeBench-Pro, along with consistent improvements across additional accuracy and efficiency metrics.
Beyond performance gains, we conduct comprehensive analyses to better understand the reasoning process within the workflow, including error patterns and cross-scenario behaviors. We further perform customized replacement and ablation studies to explore the upper bound of the framework, showing that individual agents can contribute improvements of up to 27.7\%. These results highlight the strong potential of MAS-Algorithm for advancing AI-driven algorithmic reasoning.
\end{abstract}

\section{Introduction}

Algorithmic problem solving in competitive programming (e.g., ACM-ICPC, IOI, Codeforces) serves as a stringent testbed for AI coding systems. These tasks solve highly complex problems under strict efficiency constraints, often condensing years of algorithmic insight into concise implementations. As the minimal unit of such reasoning paradigms, input–output (IO) problems capture core software engineering challenges including abstraction, optimization, and correctness. Thus, performance on algorithmic problems reflects the upper bound of an LLM’s coding and reasoning capabilities.

A series of benchmarks has been developed to evaluate these capabilities. Early datasets such as HumanEval\cite{chen2021evaluatinglargelanguagemodels,liu2023is} and MBPP\cite{austin2021programsynthesislargelanguage} focus on functional correctness, while more recent benchmarks, including LiveCodeBench\cite{jain2024livecodebenchholisticcontaminationfree} and LiveCodeBench-Pro\cite{zheng2025livecodebenchproolympiadmedalists}, emphasize difficulty, timeliness, and competition-level rigor. Additional benchmarks such as CodeElo\cite{quan2025codeelobenchmarkingcompetitionlevelcode} further evaluate real-world performance. Despite their effectiveness in measuring outcomes via execution-based metrics, these benchmarks provide limited guidance for improving the reasoning process itself.

Existing approaches for improving algorithmic problem solving can be broadly categorized into two directions. The first is model-centric training, which relies on large-scale data and training paradigms such as pretraining, supervised fine-tuning, and reinforcement learning. Representative works include AlphaCode\cite{li2022competition} and X-Coder\cite{wu2026xcoderadvancingcompetitiveprogramming}, as well as approaches that explicitly model reasoning processes through structured chain-of-thought prompting\cite{li2023structuredchainofthoughtpromptingcode,pmlr-v267-liu25ah}. While effective, these methods are highly dependent on data quality and computational resources, and often struggle to generalize to evolving, high-difficulty benchmarks.
The second direction augments LLMs with external tools or environments, including algorithm selection\cite{kim2024problemsolvingguidepredictingalgorithm,wu2024asllm}, retrieval-based methods\cite{shi2024language,tao2026retrievalaugmentedcodegenerationsurvey}, and iterative feedback mechanisms such as PyCapsule\cite{adnan2025largelanguagemodelguided}. However, these approaches remain fragmented and lack a unified, systematic framework for algorithmic reasoning.

To address these limitations, we propose MAS-Algorithm, a multi-agent workflow for algorithmic problem solving. Our framework integrates structured process design, specialized agents, auxiliary tools, and a self-constructed dataset, together with a comprehensive evaluation and analysis pipeline. The significance of our approach lies in the following aspects:
\textbf{(i) Beyond single-LLM interaction.}
Single-LLM dialogue is insufficient for complex algorithmic tasks. In practice, problem solving involves iterative strategy selection, knowledge retrieval, comparison, and feedback-driven debugging. MAS-Algorithm introduces a multi-agent setting with tool interaction to better approximate this process.
\textbf{(ii) Process decomposition.}
We model problem solving as a structured pipeline—problem understanding, hypothesis formulation, information retrieval, reasoning, implementation, and refinement. This decomposition disentangles capabilities and enables fine-grained analysis via specialized agents.
\textbf{(iii) Method and data integration.}
We unify techniques including algorithm selection, retrieval, structured reasoning, and feedback correction. Our dataset further provides problem descriptions, test cases, labels, and user code, supporting detailed analysis.

We evaluate MAS-Algorithm on Qwen models of different scales, comparing direct prompting with workflow-based solving, and additionally validate on LiveCodeBench-Pro. Results show consistent and substantial improvements across models, surpassing supervised fine-tuning. We further conduct module-level replacement and ablation experiments to analyze component contributions and explore the framework’s upper bound, alongside systematic analyses of model behavior across tasks.
As a summary, MAS-Algorithm exhibits the following key characteristics:

\begin{itemize}[nolistsep, leftmargin=*]
\item \textbf{(i) Comprehensiveness.} It unifies a wide range of existing methods for LLM-based problem solving; its dataset is diverse and broadly sourced; and its analysis framework systematically examines the capabilities required in algorithmic scenarios.
\item \textbf{(ii) Process-oriented design.} The workflow decomposes the problem-solving process into explicit stages, making intermediate steps observable, improving interpretability, and enabling fine-grained analysis of model capabilities and limitations.
\item \textbf{(iii) Engineering practicality.} The modular structure and auxiliary tools are applicable across different models and problem types, while also supporting extensibility and maintainability for future integration of additional components.
\item \textbf{(iv) Standardization.} The framework establishes a standardized paradigm for algorithmic problem solving with LLMs, provides a set of analytical methods for future research and applications.
\end{itemize}



\section{Related Work}

\textbf{LLMs for Code Generation.} Early work, such as CodeBERT\cite{feng2020codebert}, CodeT5\cite{wang2021codet5}, Codex\cite{chen2021evaluatinglargelanguagemodels}, and AlphaCode\cite{li2022competition}, has demonstrated the feasibility of LLM for code generation tasks. Now, people are gradually extending code generation to the entire software engineering workflow\cite{yang2024sweagentagentcomputerinterfacesenable,zhang2024autocoderoverautonomousprogramimprovement}, using LLM from the agent's perspective for code generation\cite{dong2025surveycodegenerationllmbased}, debugging\cite{ashrafi2025enhancingllmcodegeneration,islam2025codesimmultiagentcodegeneration}, and repairing\cite{bouzenia2024repairagentautonomousllmbasedagent}.

\textbf{Algorithmic Problems.} Classic benchmarks for algorithmic scenarios also include MATH\cite{hendrycks2021measuringmathematicalproblemsolving}, CodeContests\cite{wang2025codecontestshighqualitytestcase}, APPS\cite{hendrycks2021measuringcodingchallengecompetence}, and classic methods also include DeepCoder\cite{balog2017deepcoder}, AlphaEvolve\cite{novikov2025alphaevolvecodingagentscientific}.

\textbf{Multi-Agent Systems.} Multi-agent frameworks such as CAMEL\cite{li2023camelcommunicativeagentsmind}, MetaGPT\cite{hong2024metagpt}, AutoGen\cite{wu2024autogen}, and AgentVerse\cite{chen2024agentverse} have gained increasing attention, and are applied across diverse scenarios\cite{ye2025masgpttrainingllmsbuild,tang2025autoagentfullyautomatedzerocodeframework,zeng2026aiforsciencelowcodeplatformbayesian}, including code generation\cite{islam-etal-2024-mapcoder,islam2025codesimmultiagentcodegeneration} and software engineering tasks\cite{qian-etal-2024-chatdev,nguyen2024agilecoderdynamiccollaborativeagents}. They also demonstrate strong performance in coding tasks: MapCoder\cite{islam-etal-2024-mapcoder} improves HumanEval\cite{chen2021evaluatinglargelanguagemodels} by 7.5\%, while AgentCoder\cite{huang2024agentcodermultiagentbasedcodegeneration} and MetaGPT\cite{hong2024metagpt} achieves 96.3\% and 85.9\% on the same benchmark respectively.

\section{MAS-Algorithm}

\subsection{Agents}

In the MAS-Algorithm system, we define a set of specialized agents, each responsible for a distinct stage in the problem-solving pipeline. The overall architecture is illustrated in Figure~\ref{fig:agents}, and detailed prompts are provided in Appendix~\ref{prompt agents}.

\begin{figure*}[t]
    \centering
    \includegraphics[width=\textwidth]{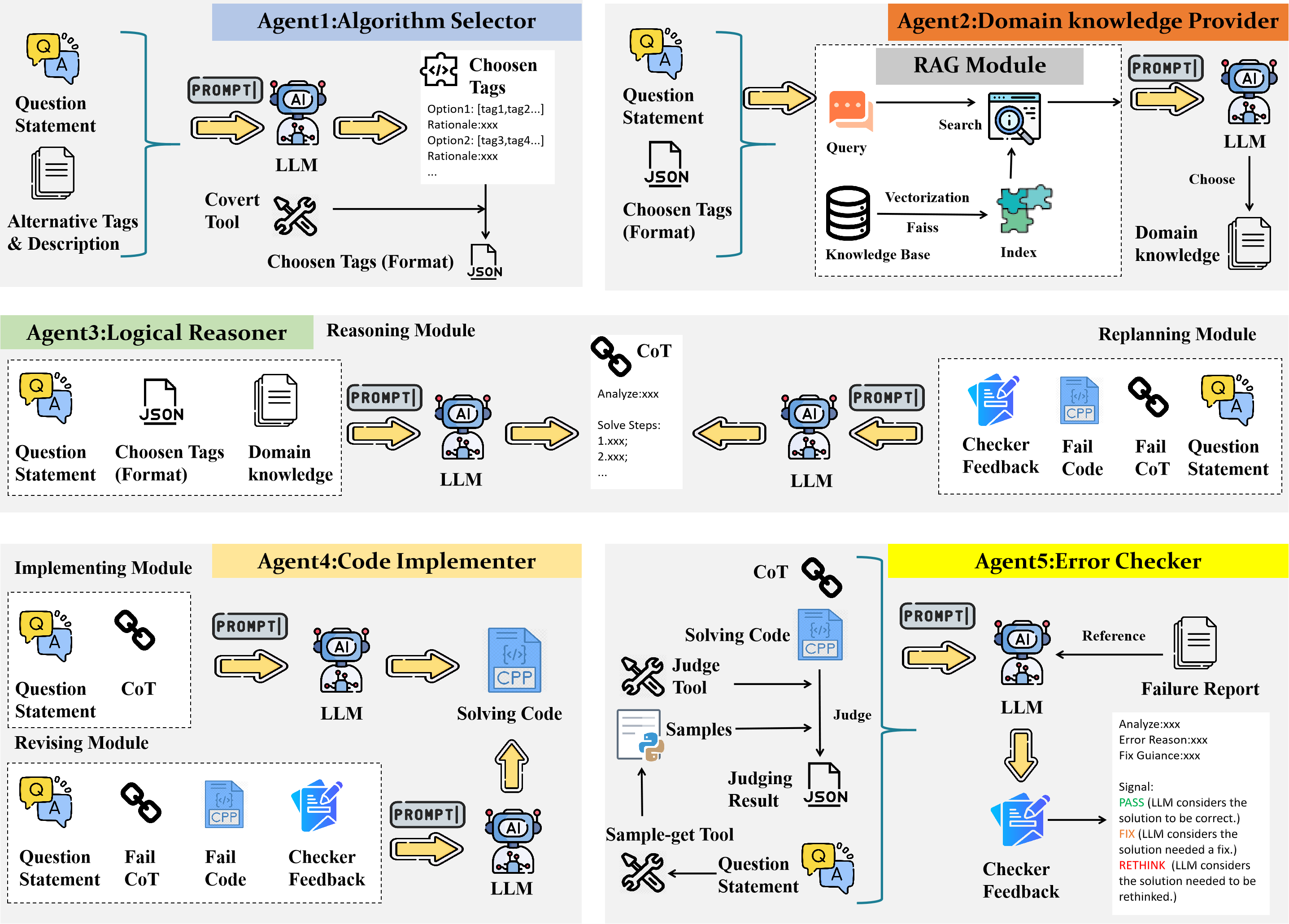}
    \caption{Structure diagram of our five Agent components.}
    \label{fig:agents}
\end{figure*}


\textbf{Agent1 – Algorithm Selector.}
Selecting appropriate algorithms and data structures is a critical step in solving algorithmic problems. An incorrect choice may lead to suboptimal complexity, flawed logic, or infeasible implementations. Inspired by the iterative strategy exploration adopted by human practitioners, we introduce an algorithm selection module to guide downstream reasoning.
Agent1 takes the problem description as input and predicts one or more relevant algorithm and data structure labels. This module is conceptually related to prior work on method selection, such as PSG\cite{kim2024problemsolvingguidepredictingalgorithm} and AS-LLM\cite{wu2024asllm}. In our implementation, we formulate the task as a multiple-choice selection problem: the model is provided with a predefined set of candidate labels (with descriptions) and is required to select the most relevant ones along with corresponding rationales.
To account for problems with multiple valid solution strategies, Agent1 supports multi-branch outputs, where each branch consists of a (label set, rationale) pair. These branches are processed in parallel in subsequent stages, enabling exploration of alternative solution paths.

\textbf{Agent2 – Domain Knowledge Provider.}
Effective problem solving often requires access to domain-specific knowledge beyond the model’s parametric memory. In practice, programmers frequently consult external resources (e.g., algorithm templates, standard techniques) before attempting a solution. To emulate this process, Agent2 provides retrieval-augmented domain knowledge conditioned on the problem and the selected labels.
We implement Agent2 using a Retrieval-Augmented Generation (RAG) pipeline, which retrieves relevant information from an external knowledge base and summarizes it for downstream use. Specifically, we adopt a standard pipeline consisting of embedding, indexing, and top-k retrieval, followed by LLM-based aggregation. We use the bge-zh embedding model~\cite{bge_embedding} and construct the knowledge base from a local copy of OI-WIKI~\cite{oiwiki}, a widely used repository of algorithmic knowledge.


\textbf{Agent3 – Logical Reasoner.}
Agent3 serves as the core reasoning module of the system. Given the problem description, selected algorithms, and retrieved knowledge, it generates a structured solution plan that specifies the reasoning process and implementation steps.
The output of Agent3 follows a structured format analogous to Chain-of-Thought (CoT), consisting of two components: problem analysis and solution steps. Each step is required to be explicit, actionable, and, when appropriate, supported by pseudocode. Importantly, this module focuses solely on logical reasoning and planning, intentionally decoupling high-level reasoning from code generation.
Agent3 operates in two modes:
(i) Reasoning mode, where it generates an initial solution plan;
(ii) Replanning mode, where it revises the plan based on feedback from downstream components (e.g., execution errors or logical inconsistencies).
This design enables iterative refinement of the reasoning process while maintaining a clear separation between planning and execution.


\textbf{Agent4 – Code Implementer.}
Agent4 is responsible for translating the structured solution plan into executable code. This module focuses on the mapping from natural language representations to programming language implementations (C++ in our setting).
It supports two modes:
(i) Implementation mode, which generates complete and executable code from the problem description and solution plan;
(ii) Revision mode, which updates previously generated code based on feedback (e.g., error reports or inconsistencies with the intended algorithm).
The generated code serves as an intermediate artifact that can be directly compiled and evaluated by the execution environment.

\textbf{Agent5 – Error Checker.}
Agent5 serves as a quality control and diagnostic module, responsible for validating intermediate outputs and guiding iterative refinement. Given the complexity of algorithmic problems, solutions generated in a single pass often contain errors, which may arise from incorrect algorithm selection, flawed reasoning, unhandled edge cases, violations of time or space constraints, or low-level implementation issues.
To address this, Agent5 performs structured analysis by leveraging multiple sources of information:
(i) the problem description;
(ii) the reasoning process and solution plan from Agent3;
(iii) the generated code from Agent4;
(iv) execution results, including test cases, expected outputs, and actual outputs; and
(v) an optional error taxonomy or checklist covering common failure patterns (e.g., boundary conditions, logical inconsistencies, and implementation bugs. An example of an failure report is shown in the Appendix \ref{samples files}.).
Based on these inputs, Agent5 produces a structured feedback report consisting of four components:
Analysis, which examines the consistency between the intended solution and the actual implementation, as well as discrepancies revealed by execution results;
Error Identification, which summarizes the root causes of failure, ranging from high-level strategy errors to low-level coding mistakes;
Remediation Guidance, which provides actionable suggestions for correction, either through local code fixes or adjustments to the solution strategy;
Control Signals, which determine the next step in the workflow.
We define three control signals:
PASS, indicating the solution is correct and meets all constraints;
FIX, indicating that implementation-level issues should be corrected without changing the overall strategy;
RETHINK, indicating that the solution is fundamentally flawed and requires replanning.
The distinction between FIX and RETHINK enables the system to differentiate between local implementation errors and global reasoning failures, allowing feedback to be routed adaptively to either Agent4 or Agent3. Through this design, Agent5 establishes a closed-loop refinement mechanism that improves both solution correctness and interpretability.re modes and their corresponding causes.

\subsection{Workflow}
\label{3.3}
To enable effective coordination among agents and ensure stable execution, we design auxiliary tools for data transformation, evaluation, and workflow control.

\textbf{Format Conversion Tool.}
This tool standardizes intermediate outputs by converting model-generated content into structured formats (e.g., JSON). For example, label selections from Agent1 are transformed into structured queries for Agent2, ensuring consistency across modules.

\textbf{Sample Extraction Tool.}
Algorithmic problem descriptions include task statements, I/O specifications, constraints, and sample test cases for correctness verification. To ensure reliability, we implement a deterministic extraction tool that parses samples directly from problem descriptions. We avoid LLM-based extraction due to potential inaccuracies. The tool uses customized regular expressions to support multiple formats (e.g., Nowcoder, HDU, Jisuanke, Codeforces) and outputs standardized input–output files.

\textbf{Judging Tools.}
We develop a custom evaluation module that functions as both a final verifier and an intermediate feedback mechanism, simulating online judge systems. Given code, the system runs it in a Docker container (gcc:latest), executes test cases, and compares outputs. Results include: (i) evaluation status (e.g., AC, WA, runtime errors); (ii) pass statistics, recording the number of test cases passed; (iii) detailed feedback such as execution time/memory, error messages, or output discrepancies. For workflow integration, the feedback additionally provides failed test indices, corresponding inputs, and mismatched outputs, enabling fine-grained diagnosis by Agent5.

\begin{figure*}[t]
    \centering
    \includegraphics[width=\textwidth]{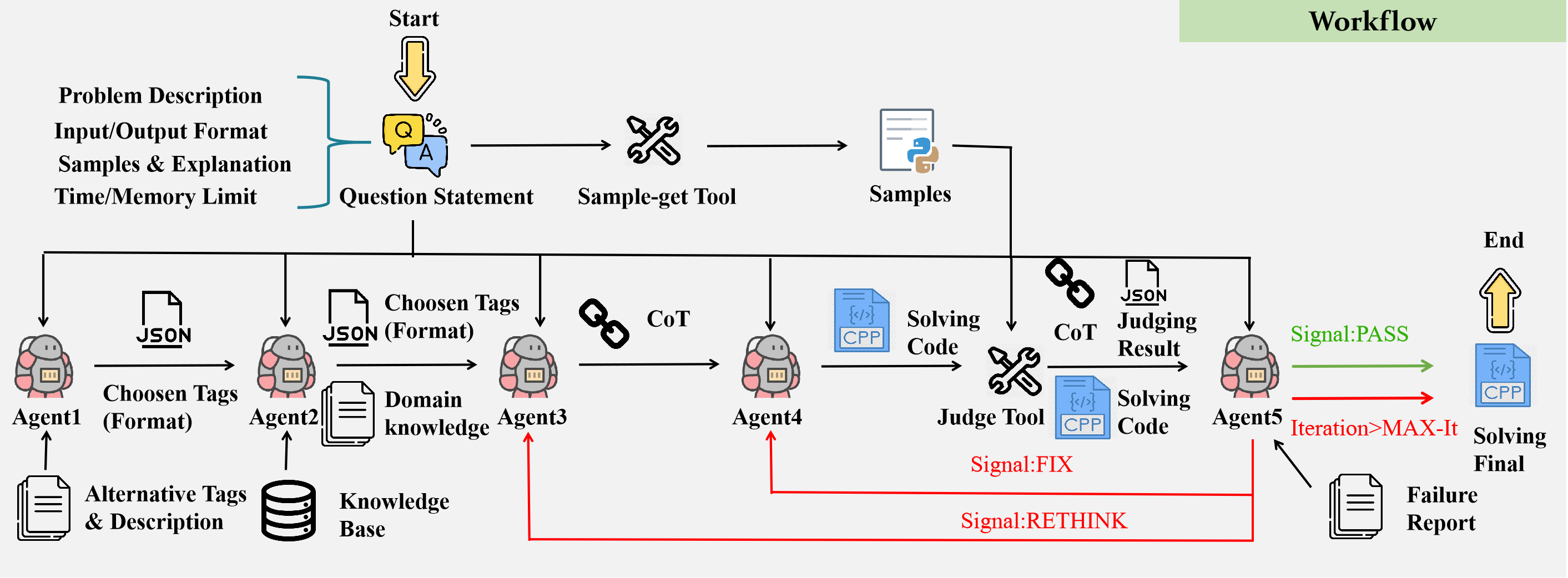}
    \caption{The execution process of the MAS-Algorithm workflow.}
    \label{fig:workflow}
\end{figure*}

\textbf{Workflow Execution.}
The overall workflow is illustrated in Figure~\ref{fig:workflow}. Given a problem-solving task, the system proceeds as follows:
Sample extraction;
Algorithm selection (Agent1);
Knowledge retrieval (Agent2);
Solution reasoning (Agent3);
Code implementation (Agent4);
Intermediate evaluation;
Error analysis (Agent5).
Subsequent actions are determined by the control signal from Agent5. If the signal is PASS or the iteration limit is reached, the process terminates. If FIX, the workflow returns to Agent4 for code revision; if RETHINK, it returns to Agent3 for solution replanning.
A complete execution example is provided in Appendix~\ref{process}.

\section{Experiement}
\subsection{Settings}

\textbf{Comparison.}
To evaluate the effectiveness of MAS-Algorithm, we conduct comparative experiments under two settings: (i) \emph{direct prompting}, where the model receives only the problem description and generates code via standard prompts; and (ii) \emph{workflow-based solving}, where the same model is integrated into the MAS-Algorithm pipeline described in Section~\ref{3.3}.  
We evaluate five Qwen models of different scales: Qwen3-8B, Qwen3-14B, Qwen3-Coder-30B-A3B-Instruct, Qwen3-Coder-480B-A35B-Instruct, and Qwen3-235B-A22B-Instruct-2507.

\textbf{Dataset.}
Experiments are primarily conducted on a self-constructed dataset derived from Alibaba's intern dataset, with the distribution of problem types shown in Figure~\ref{fig:dataset}.  
To further validate generalization, we additionally evaluate on LiveCodeBench-Pro (Q4 2024, Q1 2025, and Q2 2025).

\textbf{Fine-tuning.}
To compare against training-based approaches, we fine-tune Qwen3-Coder-30B-A3B-Instruct using accepted solutions from our dataset via LoRA, with prompts aligned to Agent4 for a fair comparison.
In addition, to reflect the training cost and extensibility of our framework, we further construct fine-tuned variants for multiple agents, including Agent1, Agent3, Agent4, and Agent5. These models are trained using a mixture of original annotations (e.g., labels and accepted code) and distilled data (e.g., reasoning traces and feedback generated by stronger models).
We note that these auxiliary fine-tuned agents are primarily used to assess the feasibility of training the workflow components, rather than to optimize final performance.

\textbf{Metrics.}
We evaluate all generated solutions using the judging tool described in Section~\ref{3.3}, and report two categories of metrics:

\emph{(i) Correctness.}
We measure (a) AC rate, (b) case pass rate, and (c) weighted case pass rate:
\begin{align*}
\text{AC rate} &= \frac{\text{Number of AC submissions}}{\text{Total number of submissions}} =\frac{\sum_{i=1}^{N} \mathbf{1}_{\{\text{status}_i = \text{AC}\}}}{N}, \\
\text{case pass rate} &= \frac{\sum_{i=1}^{N} \text{passed}_i}{\sum_{i=1}^{N} \text{total}_i}, \quad
\text{case pass rate (with weight)} = \frac{1}{N} \sum_{i=1}^{N} \frac{\text{passed}_i}{\text{total}_i}.
\end{align*}

\emph{(ii) Efficiency.}
We further report average execution time and memory usage over accepted solutions:
\[
\text{time aver} = \frac{1}{N_{\text{AC}}} \sum_{i \in \mathcal{A}} \text{time}_i, \quad
\text{mem aver} = \frac{1}{N_{\text{AC}}} \sum_{i \in \mathcal{A}} \text{memory}_i.
\]

\subsection{Result and Analysis}
\label{4.2}
\begin{table}[t!]
\scriptsize
  \caption{Judging results of models on the MAS-Algorithm and comparison with direct questioning.}
  \label{tab:overall_results}
  \renewcommand{\arraystretch}{1.2}
\setlength{\tabcolsep}{6pt}
  \centering
  \resizebox{\textwidth}{!}{
  \begin{tabular}{c|c|ccccc}
    \toprule
    \textbf{Model (Source, Dataset)}  & \textbf{Type}   & \textbf{AC rate}     & \textbf{case pass rate}  & \textbf{case pass rate}  & \textbf{time aver} & \textbf{mem aver} \\
          &    &     &   & \textbf{(with weight)}  & \textbf{(s)} & \textbf{(KB)} \\
        \midrule
Qwen3-8B  & Direct Ask & 26.21\%	& 26.41\%	& 29.22\%	& 0.48 & 5.71\\ 
(from dashscope, self dataset) & MAS-Algorithm & 32.25\% (+6.04\%) & 34.16\% (+7.75\%) & 36.41\% (+7.19\%) & 0.07 (-0.41) & 7.61 (+1.90) \\
    \midrule
Qwen3-14B  & Direct Ask & 28.98\%	& 29.34\%	& 32.13\%	& 0.67 & 5.98\\ 
(from dashscope, self dataset) & MAS-Algorithm & 37.98\% (+9.00\%) & 40.14\% (+10.80\%) & 42.51\% (+10.38\%) & 0.18 (-0.49) & 20.01 (+14.03) \\
    \midrule
Qwen3-Coder-30B-A3B-Instruct  & Direct Ask & 32.25\%	& 33.12\%	& 36.03\%	& 0.13 & 4.57\\ 
(from huggingface, self datase) & MAS-Algorithm & 38.87\% (+6.62\%) & 38.51\% (+5.39\%) & 42.28\% (+6.25\%) & 0.16 (+0.03) & 11.18 (+6.61) \\
    \midrule
Qwen3-Coder-480B-A35B-Instruct  & Direct Ask & 46.49\%	& 46.41\%	& 49.71\%	& 0.54 & 7.74\\ 
(from huggingface, self dataset) & MAS-Algorithm & 52.82\% (+6.33\%) & 52.30\% (+5.89\%) & 56.91\% (+7.20\%) & 0.24 (-0.30) & 13.39 (+5.65) \\
    \midrule
Qwen3-235B-A22B-Instruct-2507  & Direct Ask & 62.87\%	& 63.10\%	& 65.98\%	& 0.24 & 8.09\\ 
(from dashscope, self dataset) & MAS-Algorithm & 67.26\% (+4.39\%) & 67.89\% (+4.79\%) & 71.49\% (+5.51\%) & 0.20 (-0.04) & 10.12 (+2.03) \\
        \midrule
Qwen3-Coder-30B-A3B-Instruct  & Direct Ask & 33.14\% (+0.89\%)	& 33.30\% (+0.18\%)	& 36.72\% (+0.69\%)	& 0.18 (+0.05) & 5.33 (+0.76)\\ 
(LoRA Fituning, self dataset) & MAS-Algorithm & 23.84\% (-15.03\%) & 25.89\% (-12.62\%) & 27.87\% (-14.41\%) & 0.23 (+0.07) & 6.51 (-4.67) \\
       \midrule
Qwen3-Coder-30B-A3B-Instruct  & Direct Ask & 5.28\%	& 9.07\%	& 8.82\%	& 0.02 & 4.93\\ 
(from huggingface, LiveCodeBench-Pro) & MAS-Algorithm & 10.00\% (+4.72\%) & 14.05\% (+4.98\%) & 13.81\% (+4.99\%) & 0.04 (+0.02) & 5.24 (+0.31) \\
    
    \bottomrule
  \end{tabular}
  }
  \vspace{-0.55cm}
\end{table}

\textbf{Comparison.}
The overall results are summarized in Table~\ref{tab:overall_results}. Across all five evaluated models, the MAS-Algorithm consistently yields substantial improvements over direct prompting in all correctness metrics. On average, the AC rate increases by 6.48\%, with gains ranging from 4.39\% to 9.00\%. Notably, Qwen3-14B achieves the largest improvement, exceeding 9\% in AC rate and over 10\% in both case-based metrics, indicating that mid-scale models particularly benefit from structured workflow guidance.
This trend generalizes to more challenging benchmarks. On LiveCodeBench-Pro, MAS-Algorithm improves performance by 4.72\%–4.99\% across metrics, demonstrating its robustness beyond the in-domain dataset (See Table~\ref{tab:quarter} for details, also demonstrate the good generalization ability of MAS-Algorithm).
In contrast, training-based approaches provide limited gains. Fine-tuning Qwen3-Coder-30B-A3B-Instruct improves AC rate by only 0.89\%, which is substantially lower than the 6.62\% gain achieved by MAS-Algorithm. This suggests that, under comparable data conditions, structured inference-time workflows are more effective than parameter updates for improving performance in algorithmic reasoning tasks.
Regarding efficiency, MAS-Algorithm generally reduces execution time while increasing memory usage. This indicates a tendency toward more time-efficient but space-intensive solutions, likely due to the selection of more robust or generalized algorithmic strategies. Given that memory consumption remains within acceptable limits, this trade-off is practical and further supports the effectiveness of the proposed workflow.

\begin{figure*}[t]
    \centering
    \includegraphics[width=\textwidth]{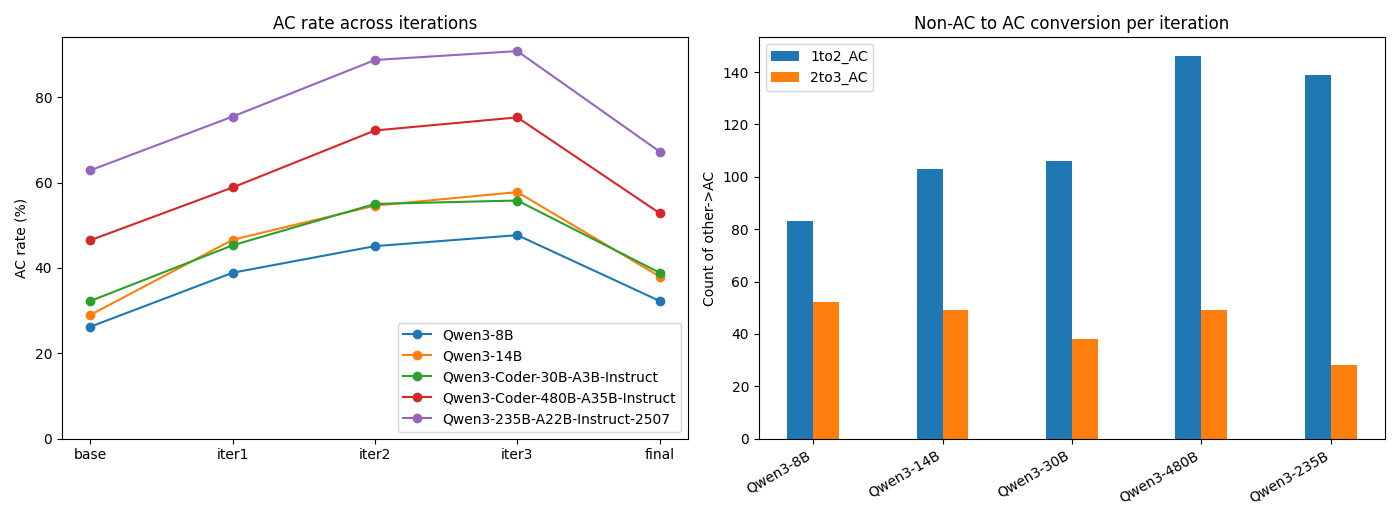}
    \caption{
Evolution of AC rate across iterative refinement and transition statistics. 
\textbf{Left:} AC rate for each model over iterations. “Base” denotes direct prompting without the workflow, while “Final” denotes the final evaluation result of MAS-Algorithm after all iterations. 
\textbf{Right:} Number of problems that transition from non-AC to AC between consecutive iterations, reflecting the effectiveness of iterative correction. (Model's abbreviations: Qwen3-Coder-30B-A3B-Instruct: Qwen3-30B, Qwen3-Coder-480B-A35B-Instruct: Qwen3-480B, Qwen3-235B-A22B-Instruct-2507: Qwen3-235B)
}
    \label{fig:iter_ac}
      \vspace{-0.4cm}
\end{figure*}

\textbf{Changes in AC rate during the iteration process.}
Beyond aggregate performance, we analyze intermediate behaviors of MAS-Algorithm to understand the role and limitations of each module. Figure~\ref{fig:iter_ac} shows the evolution of AC rate across iterations. Starting from direct prompting, all models exhibit consistent improvements from Iter1 to Iter3, confirming the effectiveness of the iterative feedback mechanism. However, a noticeable drop occurs from Iter3 to the final evaluation, indicating that many solutions pass sample tests but fail on full evaluation cases. This highlights a key limitation: while the workflow improves local correctness, ensuring generalization to more complex or hidden test cases remains challenging.
The right panel further quantifies transitions from non-AC to AC states. Although substantial corrections occur in early iterations, the number of successful transitions decreases in later rounds, suggesting diminishing returns. This indicates that iterative refinement converges to a performance plateau, where remaining errors are likely due to inherent model limitations or require fundamentally different reasoning strategies.

\begin{figure*}[t]
    \centering
    \includegraphics[width=\textwidth]{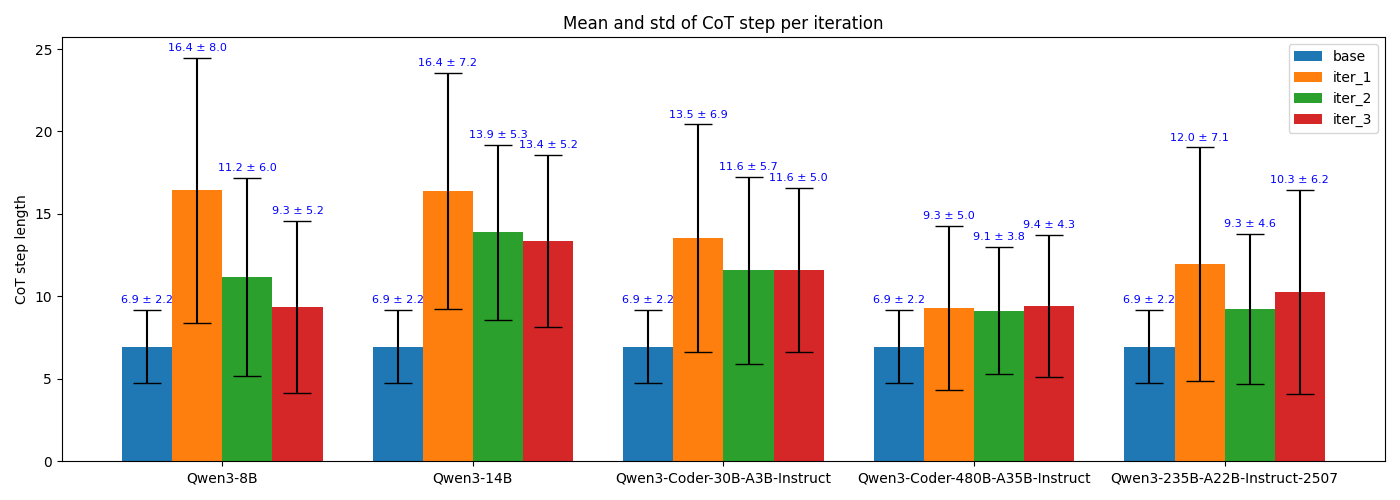}
\caption{
Statistics of Chain-of-Thought (CoT) length across iterations. 
“Base” refers to a distilled reference constructed by applying a strong model (GPT-5.2) to extract solution steps from ground-truth accepted code. 
Bars indicate mean CoT length, and error lines denote standard deviation. The figure shows how reasoning complexity and stability evolve during iterative refinement.
}
    \label{fig:cot_len}
    \vspace{-0.5cm}
\end{figure*}

\textbf{CoT during the iteration process.}
Figure~\ref{fig:cot_len} analyzes the reasoning behavior of Agent3 through the length of generated solution steps (CoT length). Compared to a distilled reference baseline, all models produce longer and more variable reasoning chains, indicating that LLM-generated reasoning is generally more verbose and less stable than human-like solutions. In some cases, CoT length exceeds 10–20 steps, reflecting redundancy or inefficient reasoning patterns.
Across iterations, CoT length tends to decrease and stabilize, particularly for larger models. This suggests that iterative feedback helps refine reasoning efficiency. At the same time, overly long reasoning chains may correlate with incorrect solutions, implying that excessive or unfocused reasoning can hinder performance. In extreme cases, models may generate spurious reasoning when unable to solve a problem, revealing a potential failure mode of current LLM-based reasoning systems.


\textbf{Failure analysis.}
Table~\ref{tab:failure_class} presents a taxonomy of errors in the final incorrect solutions, providing insight into the failure modes of LLMs in algorithmic tasks. The analysis is constructed using GPT-5.2\cite{openai2025gpt52} to compare model outputs against ground-truth accepted solutions and categorize errors across problems.
We identify thirteen error categories, with the majority concentrated in a few dominant types. In particular, core modeling errors, complexity misjudgment, and semantic misunderstanding together account for over half of all failures. This indicates that deficiencies in abstraction, constraint reasoning, and problem interpretation remain the primary bottlenecks for current models. Improving these capabilities could potentially address a large fraction of unsolved cases.
Based on these observations, we construct a structured error report as part of Agent5, enabling the model to explicitly check for common failure patterns. However, empirical results show that such prompt-level guidance does not always lead to effective self-correction. In some cases, models persist in incorrect interpretations despite repeated feedback, as illustrated in Appendix~\ref{misunder}. This suggests that certain failure modes are not easily mitigated through iterative prompting alone, and may require stronger reasoning or representation capabilities. Overall, these findings highlight that improving high-level reasoning and problem abstraction remains critical for advancing LLM performance.

\subsection{Replacement and Ablation}
\label{4.3}

While Section~\ref{4.2} validates the overall effectiveness of MAS-Algorithm, it does not isolate the contribution of individual components. To address this, we propose a replacement and ablation paradigm. Unlike conventional ablation—where modules are removed—we instead enhance or substitute specific Agents with oracle-like or augmented variants. This design is motivated by two considerations: (i) MAS-Algorithm represents an integrated problem-solving pipeline, where removing components (e.g., reasoning or coding) would break functional completeness; moreover, direct prompting already approximates a single-module (Agent4) baseline, making further removal redundant; (ii) replacing modules with stronger or idealized versions enables us to probe the upper bound of the framework, revealing both the potential of the overall system and the gap between current implementations and their optimal counterparts.
We define the following variants:

\begin{table}[t!]
\scriptsize
  \caption{Judging results of replacement and ablation experiments.}
  \label{tab:ablation results}
  \renewcommand{\arraystretch}{1.2}
\setlength{\tabcolsep}{6pt}
  \centering
  \resizebox{\textwidth}{!}{
  \begin{tabular}{c|ccccc}
    \toprule
       \textbf{Type}   & \textbf{AC rate}     & \textbf{case pass rate}  & \textbf{case pass rate}  & \textbf{time aver} & \textbf{mem aver} \\
              & \    &   & \textbf{(with weight)}  & \textbf{(s)} & \textbf{(KB)} \\
        \midrule
Agent0  & 52.13\% (+13.26\%)	& 53.33\% (+14.82\%)	& 55.83\% (+13.55\%)	& 0.05 (-0.11) & 8.20 (-2.98)\\ 
    \midrule
Agent1  & 38.38\% (-0.49\%)	& 38.59\% (+0.08\%)	& 42.13\% (-0.15\%)	& 0.24 (+0.08) & 6.31 (-4.87)\\ 
    \midrule
    Agent2  & 61.82\% (+22.95\%)	& 61.68\% (+23.17\%)	& 65.02\% (+22.74\%)	& 0.08 (-0.08) & 7.58 (-3.60)\\ 
    \midrule
    Agent3  & 66.57\% (+27.70\%)	& 66.15\% (+27.64\%)	& 69.57\% (+27.29\%)	& 0.07 (-0.09) & 8.78 (-2.40)\\ 
    \midrule
    Agent5  & 48.27\% (+9.40\%)	& 48.58\% (+10.07\%)	& 50.86\% (+8.58\%)	& 0.27 (+0.11) & 9.53 (-1.65)\\ 
    
    \bottomrule
  \end{tabular}
  }
  \vspace{-0.55cm}
\end{table}

\textbf{Agent0 (Semantic Interpreter):}
To mitigate semantic misunderstandings, we introduce a preprocessing module that rewrites problem statements. Using GPT-5.2\cite{openai2025gpt52}, the problem is clarified by removing irrelevant details, normalizing formats (e.g., formulas), and explaining ambiguous terms based on reference solutions. The refined statement replaces the original input.

\textbf{Agent1 (Oracle Labels):}
We replace predicted labels with ground-truth algorithm labels provided in the dataset, while keeping the rest of the pipeline unchanged.

\textbf{Agent2 (Enhanced Knowledge Base):}
Beyond OI-WIKI, we augment the knowledge base with a curated problem bank consisting of problem descriptions and accepted solutions. The RAG module retrieves both conceptual knowledge and top-k similar examples for reference.

\textbf{Agent3 (Distilled CoT):}
The reasoning output is replaced with distilled solution steps (analysis + procedure) derived from correct solutions, as described in Section~\ref{4.2}.

\textbf{Agent5 (Oracle Failure Report):}
We provide a customized error report for each problem, constructed from prior failure analysis. This report explicitly highlights common pitfalls and guides correction.


Table \ref{tab:ablation results} presents the results. All variants except Agent1 yield substantial improvements across correctness metrics, accompanied by reduced runtime and memory usage. Notably, replacing Agent3 leads to the largest gain (+27.7\% AC rate), underscoring the central role of structured reasoning in algorithmic problem solving. Similarly, the enhanced knowledge base (Agent2) contributes over 20\% improvement, highlighting the importance of high-quality external knowledge.
In contrast, replacing Agent1 yields marginal impact. We attribute this to the current limitations of the knowledge integration module (Agent2): noisy or overly broad label predictions may inadvertently introduce diverse but partially useful knowledge, whereas strict ground-truth labels restrict this diversity. As a result, the benefit of label accuracy is not fully realized in isolation.
Overall, these experiments demonstrate that MAS-Algorithm has substantial headroom for improvement. More importantly, they provide a principled way to identify which components are most critical and how far each remains from its potential upper bound, offering clear directions for future optimization.

\subsection{Other Analysis}
\label{4.4}

We conduct additional analyses to further examine the generality and limitations of MAS-Algorithm.

\textbf{Category Analysis.}
We evaluate performance across different problem categories. Taking Qwen3-Coder-30B-A3B-Instruct as a representative model, the results (Table \ref{tab:category_results}) show consistent improvements across all categories, demonstrating that MAS-Algorithm is not specialized for a narrow subset of tasks but provides broad gains. Notably, the largest improvements occur in geometry and matrix problems, while gains in mathematically intensive categories such as number theory and DP are relatively smaller. This suggests that tasks requiring deeper formal reasoning and precise modeling remain more challenging, highlighting key directions for future improvement.

\begin{table}[t!]
\scriptsize
  \caption{Judging results for Qwen3-Coder-30B-A3B-Instruct categorized by question type. We selected five representative scenarios; the complete results are shown in the Table~\ref{tab:category_results_all}.}
  \label{tab:category_results}
  \renewcommand{\arraystretch}{1.2}
  \setlength{\tabcolsep}{5pt}
  \centering
  \resizebox{\textwidth}{!}{
  \begin{tabular}{c|c|ccccc}
    \toprule
    \textbf{Category} & \textbf{Type} & \textbf{AC rate} & \textbf{case pass rate} & \textbf{case pass rate} & \textbf{time aver} & \textbf{mem aver} \\
     &  &  &  & \textbf{(with weight)} & \textbf{(s)} & \textbf{(KB)} \\
     \midrule
     Graph \& Network
      & Direct Ask & 11.49\% & 12.22\% & 14.63\% & 0.02 & 4.77 \\
      & MAS-Algorithm & 18.24\% (+6.76\%) & 17.90\% (+5.68\%) & 22.05\% (+7.42\%) & 0.15 (+0.13) & 9.38 (+4.61) \\
    \midrule
    DP \& Optimization
      & Direct Ask & 17.39\% & 18.33\% & 21.29\% & 0.74 & 9.61 \\
      & MAS-Algorithm & 22.22\% (+4.83\%) & 20.87\% (+2.53\%) & 26.80\% (+5.51\%) & 0.45 (-0.30) & 9.28 (-0.33) \\
    \midrule
    Math \& Number Theory
      & Direct Ask & 31.79\% & 35.10\% & 35.64\% & 0.10 & 4.29 \\
      & MAS-Algorithm & 35.41\% (+3.62\%) & 38.18\% (+3.08\%) & 38.80\% (+3.16\%) & 0.06 (-0.04) & 5.88 (+1.58) \\
    \midrule
    Geometry \& Matrix
      & Direct Ask & 24.00\% & 23.14\% & 24.48\% & 0.00 & 3.62 \\
      & MAS-Algorithm & 40.00\% (+16.00\%) & 44.15\% (+21.01\%) & 40.33\% (+15.86\%) & 0.05 (+0.05) & 6.43 (+2.80) \\
    \midrule
        Greedy \& Heuristics
      & Direct Ask & 35.23\% & 36.28\% & 39.11\% & 0.15 & 4.05 \\
      & MAS-Algorithm & 41.78\% (+6.55\%) & 41.62\% (+5.34\%) & 45.07\% (+5.96\%) & 0.13 (-0.02) & 6.00 (+1.95) \\
    \bottomrule
  \end{tabular}
  }
  \vspace{-0.55cm}
\end{table}

\textbf{Brute-force Analysis.}
In practice, brute-force solutions are often used as a baseline for verification. This motivates a potential extension of MAS-Algorithm with a “cross-checking” module. To assess feasibility, we evaluate the model’s ability to generate brute-force solutions (direct asking using different prompt)  under relaxed criteria (accepting both AC and TLE). As shown in Table \ref{tab:brute force}, most models exhibit minimal improvement in AC rate, and even when allowing TLE, a substantial portion of problems remains unsolved. This indicates that current LLMs struggle with correctness rather than efficiency: they often fail to produce even valid brute-force implementations. Consequently, incorporating brute-force-based verification is not yet reliable, and improving correctness—particularly problem understanding and modeling—remains the primary bottleneck.

\section{Conclusion}
We propose MAS-Algorithm, a multi-agent workflow for algorithmic problem solving with LLMs. By decomposing the process into modular stages—method selection, knowledge retrieval, structured reasoning, code generation, and feedback refinement—the framework enables more effective and interpretable problem solving.
Experiments on our dataset and LiveCodeBench-Pro show consistent performance improvements across models, outperforming direct prompting and surpassing supervised fine-tuning. Beyond performance, the workflow supports fine-grained analysis of intermediate reasoning, revealing key limitations in modeling accuracy and semantic understanding.
Through replacement and ablation studies, we further demonstrate the potential upper bound of the framework and identify critical components such as reasoning and knowledge integration. Additional analyses confirm its general applicability while highlighting challenges in reasoning-intensive tasks.
Overall, MAS-Algorithm provides a practical and extensible paradigm for LLM-based algorithmic reasoning, offering a foundation for future work on structured reasoning and multi-agent code generation.

\newpage
\section*{Limitations} MAS-Algorithm is an initial instantiation of our framework; both agent design and tool implementations remain preliminary, leaving a gap to the upper bounds observed in replacement experiments. Training under this paradigm is also challenging: multi-agent fine-tuning shows unexpected degradation, likely due to data quality issues and the lack of standardized supervision for components such as reasoning, label selection, and feedback. In addition, evaluation of intermediate outputs is still limited, and more scalable metrics are needed. Finally, although the framework may generalize to other reasoning-intensive domains, we do not validate this in the current work.

\bibliographystyle{unsrt}
\bibliography{reference}

@misc{chen2021evaluatinglargelanguagemodels,
      title={Evaluating Large Language Models Trained on Code}, 
      author={Mark Chen and Jerry Tworek and Heewoo Jun and Qiming Yuan and Henrique Ponde de Oliveira Pinto and Jared Kaplan and Harri Edwards and Yuri Burda and Nicholas Joseph and Greg Brockman and Alex Ray and Raul Puri and Gretchen Krueger and Michael Petrov and Heidy Khlaaf and Girish Sastry and Pamela Mishkin and Brooke Chan and Scott Gray and Nick Ryder and Mikhail Pavlov and Alethea Power and Lukasz Kaiser and Mohammad Bavarian and Clemens Winter and Philippe Tillet and Felipe Petroski Such and Dave Cummings and Matthias Plappert and Fotios Chantzis and Elizabeth Barnes and Ariel Herbert-Voss and William Hebgen Guss and Alex Nichol and Alex Paino and Nikolas Tezak and Jie Tang and Igor Babuschkin and Suchir Balaji and Shantanu Jain and William Saunders and Christopher Hesse and Andrew N. Carr and Jan Leike and Josh Achiam and Vedant Misra and Evan Morikawa and Alec Radford and Matthew Knight and Miles Brundage and Mira Murati and Katie Mayer and Peter Welinder and Bob McGrew and Dario Amodei and Sam McCandlish and Ilya Sutskever and Wojciech Zaremba},
      year={2021},
      eprint={2107.03374},
      archivePrefix={arXiv},
      primaryClass={cs.LG},
      url={https://arxiv.org/abs/2107.03374}, 
}

@inproceedings{
liu2023is,
title={Is Your Code Generated by Chat{GPT} Really Correct? Rigorous Evaluation of Large Language Models for Code Generation},
author={Jiawei Liu and Chunqiu Steven Xia and Yuyao Wang and LINGMING ZHANG},
booktitle={Thirty-seventh Conference on Neural Information Processing Systems},
year={2023},
url={https://openreview.net/forum?id=1qvx610Cu7}
}

@misc{austin2021programsynthesislargelanguage,
      title={Program Synthesis with Large Language Models}, 
      author={Jacob Austin and Augustus Odena and Maxwell Nye and Maarten Bosma and Henryk Michalewski and David Dohan and Ellen Jiang and Carrie Cai and Michael Terry and Quoc Le and Charles Sutton},
      year={2021},
      eprint={2108.07732},
      archivePrefix={arXiv},
      primaryClass={cs.PL},
      url={https://arxiv.org/abs/2108.07732}, 
}

@misc{jain2024livecodebenchholisticcontaminationfree,
      title={LiveCodeBench: Holistic and Contamination Free Evaluation of Large Language Models for Code}, 
      author={Naman Jain and King Han and Alex Gu and Wen-Ding Li and Fanjia Yan and Tianjun Zhang and Sida Wang and Armando Solar-Lezama and Koushik Sen and Ion Stoica},
      year={2024},
      eprint={2403.07974},
      archivePrefix={arXiv},
      primaryClass={cs.SE},
      url={https://arxiv.org/abs/2403.07974}, 
}

@misc{zheng2025livecodebenchproolympiadmedalists,
      title={LiveCodeBench Pro: How Do Olympiad Medalists Judge LLMs in Competitive Programming?}, 
      author={Zihan Zheng and Zerui Cheng and Zeyu Shen and Shang Zhou and Kaiyuan Liu and Hansen He and Dongruixuan Li and Stanley Wei and Hangyi Hao and Jianzhu Yao and Peiyao Sheng and Zixuan Wang and Wenhao Chai and Aleksandra Korolova and Peter Henderson and Sanjeev Arora and Pramod Viswanath and Jingbo Shang and Saining Xie},
      year={2025},
      eprint={2506.11928},
      archivePrefix={arXiv},
      primaryClass={cs.SE},
      url={https://arxiv.org/abs/2506.11928}, 
}

@misc{quan2025codeelobenchmarkingcompetitionlevelcode,
      title={CodeElo: Benchmarking Competition-level Code Generation of LLMs with Human-comparable Elo Ratings}, 
      author={Shanghaoran Quan and Jiaxi Yang and Bowen Yu and Bo Zheng and Dayiheng Liu and An Yang and Xuancheng Ren and Bofei Gao and Yibo Miao and Yunlong Feng and Zekun Wang and Jian Yang and Zeyu Cui and Yang Fan and Yichang Zhang and Binyuan Hui and Junyang Lin},
      year={2025},
      eprint={2501.01257},
      archivePrefix={arXiv},
      primaryClass={cs.CL},
      url={https://arxiv.org/abs/2501.01257}, 
}

@misc{wu2026xcoderadvancingcompetitiveprogramming,
      title={X-Coder: Advancing Competitive Programming with Fully Synthetic Tasks, Solutions, and Tests}, 
      author={Jie Wu and Haoling Li and Xin Zhang and Jiani Guo and Jane Luo and Steven Liu and Yangyu Huang and Ruihang Chu and Scarlett Li and Yujiu Yang},
      year={2026},
      eprint={2601.06953},
      archivePrefix={arXiv},
      primaryClass={cs.CL},
      url={https://arxiv.org/abs/2601.06953}, 
}

@misc{li2023structuredchainofthoughtpromptingcode,
      title={Structured Chain-of-Thought Prompting for Code Generation}, 
      author={Jia Li and Ge Li and Yongmin Li and Zhi Jin},
      year={2023},
      eprint={2305.06599},
      archivePrefix={arXiv},
      primaryClass={cs.SE},
      url={https://arxiv.org/abs/2305.06599}, 
}

@InProceedings{pmlr-v267-liu25ah,
  title = 	 {Revisiting Chain-of-Thought in Code Generation: Do Language Models Need to Learn Reasoning before Coding?},
  author =       {Liu, Ren-Biao and Li, Anqi and Yang, Chaoding and Sun, Hui and Li, Ming},
  booktitle = 	 {Proceedings of the 42nd International Conference on Machine Learning},
  pages = 	 {38809--38826},
  year = 	 {2025},
  editor = 	 {Singh, Aarti and Fazel, Maryam and Hsu, Daniel and Lacoste-Julien, Simon and Berkenkamp, Felix and Maharaj, Tegan and Wagstaff, Kiri and Zhu, Jerry},
  volume = 	 {267},
  series = 	 {Proceedings of Machine Learning Research},
  month = 	 {13--19 Jul},
  publisher =    {PMLR},
  url = 	 {https://proceedings.mlr.press/v267/liu25ah.html}
}

@misc{kim2024problemsolvingguidepredictingalgorithm,
      title={Problem-Solving Guide: Predicting the Algorithm Tags and Difficulty for Competitive Programming Problems}, 
      author={Juntae Kim and Eunjung Cho and Dongbin Na},
      year={2024},
      eprint={2310.05791},
      archivePrefix={arXiv},
      primaryClass={cs.CL},
      url={https://arxiv.org/abs/2310.05791}, 
}

@misc{
wu2024asllm,
title={{AS}-{LLM}: When Algorithm Selection Meets Large Language Model},
author={Xingyu Wu and Yan Zhong and Jibin Wu and KC Tan},
year={2024},
url={https://openreview.net/forum?id=l7aD9VMQUq}
}

@misc{shi2024language,
            title={Can Language Models Solve Olympiad Programming?}, 
            author={Quan Shi and Michael Tang and Karthik Narasimhan and Shunyu Yao},
            year={2024},
            eprint={2404.10952},
            archivePrefix={arXiv},
            primaryClass={cs.CL}
          }

@misc{tao2026retrievalaugmentedcodegenerationsurvey,
      title={Retrieval-Augmented Code Generation: A Survey with Focus on Repository-Level Approaches}, 
      author={Yicheng Tao and Yao Qin and Yepang Liu},
      year={2026},
      eprint={2510.04905},
      archivePrefix={arXiv},
      primaryClass={cs.SE},
      url={https://arxiv.org/abs/2510.04905}, 
}

@misc{adnan2025largelanguagemodelguided,
      title={Large Language Model Guided Self-Debugging Code Generation}, 
      author={Muntasir Adnan and Zhiwei Xu and Carlos C. N. Kuhn},
      year={2025},
      eprint={2502.02928},
      archivePrefix={arXiv},
      primaryClass={cs.SE},
      url={https://arxiv.org/abs/2502.02928}, 
}

@misc{huang2024agentcodermultiagentbasedcodegeneration,
      title={AgentCoder: Multi-Agent-based Code Generation with Iterative Testing and Optimisation}, 
      author={Dong Huang and Jie M. Zhang and Michael Luck and Qingwen Bu and Yuhao Qing and Heming Cui},
      year={2024},
      eprint={2312.13010},
      archivePrefix={arXiv},
      primaryClass={cs.CL},
      url={https://arxiv.org/abs/2312.13010}, 
}

@misc{bge_embedding,
      title={C-Pack: Packaged Resources To Advance General Chinese Embedding}, 
      author={Shitao Xiao and Zheng Liu and Peitian Zhang and Niklas Muennighoff},
      year={2023},
      eprint={2309.07597},
      archivePrefix={arXiv},
      primaryClass={cs.CL}
}

@misc{oiwiki,
  author = {OI Wiki Team},
  title = {OI Wiki},
  year = {2016},
  publisher = {GitHub},
  journal = {GitHub Repository},
  howpublished = {\url{https://github.com/OI-wiki/OI-wiki}},
}

@article{feng2020codebert,
  title={Codebert: A pre-trained model for programming and natural languages},
  author={Feng, Zhangyin and Guo, Daya and Tang, Duyu and Duan, Nan and Feng, Xiaocheng and Gong, Ming and Shou, Linjun and Qin, Bing and Liu, Ting and Jiang, Daxin and others},
  journal={arXiv preprint arXiv:2002.08155},
  year={2020}
}

@article{wang2021codet5,
  title={Codet5: Identifier-aware unified pre-trained encoder-decoder models for code understanding and generation},
  author={Wang, Yue and Wang, Weishi and Joty, Shafiq and Hoi, Steven CH},
  journal={arXiv preprint arXiv:2109.00859},
  year={2021}
}

@article{li2022competition,
  title={Competition-level code generation with alphacode},
  author={Li, Yujia and Choi, David and Chung, Junyoung and Kushman, Nate and Schrittwieser, Julian and Leblond, R{\'e}mi and Eccles, Tom and Keeling, James and Gimeno, Felix and Dal Lago, Agustin and others},
  journal={Science},
  volume={378},
  number={6624},
  pages={1092--1097},
  year={2022},
  publisher={American Association for the Advancement of Science}
}

@misc{bouzenia2024repairagentautonomousllmbasedagent,
      title={RepairAgent: An Autonomous, LLM-Based Agent for Program Repair}, 
      author={Islem Bouzenia and Premkumar Devanbu and Michael Pradel},
      year={2024},
      eprint={2403.17134},
      archivePrefix={arXiv},
      primaryClass={cs.SE},
      url={https://arxiv.org/abs/2403.17134}, 
}

@misc{zhang2024autocoderoverautonomousprogramimprovement,
      title={AutoCodeRover: Autonomous Program Improvement}, 
      author={Yuntong Zhang and Haifeng Ruan and Zhiyu Fan and Abhik Roychoudhury},
      year={2024},
      eprint={2404.05427},
      archivePrefix={arXiv},
      primaryClass={cs.SE},
      url={https://arxiv.org/abs/2404.05427}, 
}

@misc{islam2025codesimmultiagentcodegeneration,
      title={CODESIM: Multi-Agent Code Generation and Problem Solving through Simulation-Driven Planning and Debugging}, 
      author={Md. Ashraful Islam and Mohammed Eunus Ali and Md Rizwan Parvez},
      year={2025},
      eprint={2502.05664},
      archivePrefix={arXiv},
      primaryClass={cs.CL},
      url={https://arxiv.org/abs/2502.05664}, 
}

@misc{yang2024sweagentagentcomputerinterfacesenable,
      title={SWE-agent: Agent-Computer Interfaces Enable Automated Software Engineering}, 
      author={John Yang and Carlos E. Jimenez and Alexander Wettig and Kilian Lieret and Shunyu Yao and Karthik Narasimhan and Ofir Press},
      year={2024},
      eprint={2405.15793},
      archivePrefix={arXiv},
      primaryClass={cs.SE},
      url={https://arxiv.org/abs/2405.15793}, 
}

@misc{dong2025surveycodegenerationllmbased,
      title={A Survey on Code Generation with LLM-based Agents}, 
      author={Yihong Dong and Xue Jiang and Jiaru Qian and Tian Wang and Kechi Zhang and Zhi Jin and Ge Li},
      year={2025},
      eprint={2508.00083},
      archivePrefix={arXiv},
      primaryClass={cs.SE},
      url={https://arxiv.org/abs/2508.00083}, 
}

@misc{ashrafi2025enhancingllmcodegeneration,
      title={Enhancing LLM Code Generation: A Systematic Evaluation of Multi-Agent Collaboration and Runtime Debugging for Improved Accuracy, Reliability, and Latency}, 
      author={Nazmus Ashrafi and Salah Bouktif and Mohammed Mediani},
      year={2025},
      eprint={2505.02133},
      archivePrefix={arXiv},
      primaryClass={cs.SE},
      url={https://arxiv.org/abs/2505.02133}, 
}

@misc{novikov2025alphaevolvecodingagentscientific,
      title={AlphaEvolve: A coding agent for scientific and algorithmic discovery}, 
      author={Alexander Novikov and Ngân Vũ and Marvin Eisenberger and Emilien Dupont and Po-Sen Huang and Adam Zsolt Wagner and Sergey Shirobokov and Borislav Kozlovskii and Francisco J. R. Ruiz and Abbas Mehrabian and M. Pawan Kumar and Abigail See and Swarat Chaudhuri and George Holland and Alex Davies and Sebastian Nowozin and Pushmeet Kohli and Matej Balog},
      year={2025},
      eprint={2506.13131},
      archivePrefix={arXiv},
      primaryClass={cs.AI},
      url={https://arxiv.org/abs/2506.13131}, 
}

@misc{wang2025codecontestshighqualitytestcase,
      title={CodeContests+: High-Quality Test Case Generation for Competitive Programming}, 
      author={Zihan Wang and Siyao Liu and Yang Sun and Hongyan Li and Kai Shen},
      year={2025},
      eprint={2506.05817},
      archivePrefix={arXiv},
      primaryClass={cs.SE},
      url={https://arxiv.org/abs/2506.05817}, 
}

@misc{hendrycks2021measuringmathematicalproblemsolving,
      title={Measuring Mathematical Problem Solving With the MATH Dataset}, 
      author={Dan Hendrycks and Collin Burns and Saurav Kadavath and Akul Arora and Steven Basart and Eric Tang and Dawn Song and Jacob Steinhardt},
      year={2021},
      eprint={2103.03874},
      archivePrefix={arXiv},
      primaryClass={cs.LG},
      url={https://arxiv.org/abs/2103.03874}, 
}

@misc{hendrycks2021measuringcodingchallengecompetence,
      title={Measuring Coding Challenge Competence With APPS}, 
      author={Dan Hendrycks and Steven Basart and Saurav Kadavath and Mantas Mazeika and Akul Arora and Ethan Guo and Collin Burns and Samir Puranik and Horace He and Dawn Song and Jacob Steinhardt},
      year={2021},
      eprint={2105.09938},
      archivePrefix={arXiv},
      primaryClass={cs.SE},
      url={https://arxiv.org/abs/2105.09938}, 
}

@inproceedings{
balog2017deepcoder,
title={DeepCoder: Learning to Write Programs},
author={Matej Balog and Alexander L. Gaunt and Marc Brockschmidt and Sebastian Nowozin and Daniel Tarlow},
booktitle={International Conference on Learning Representations},
year={2017},
url={https://openreview.net/forum?id=ByldLrqlx}
}

@inproceedings{
hong2024metagpt,
title={Meta{GPT}: Meta Programming for A Multi-Agent Collaborative Framework},
author={Sirui Hong and Mingchen Zhuge and Jonathan Chen and Xiawu Zheng and Yuheng Cheng and Jinlin Wang and Ceyao Zhang and Zili Wang and Steven Ka Shing Yau and Zijuan Lin and Liyang Zhou and Chenyu Ran and Lingfeng Xiao and Chenglin Wu and J{\"u}rgen Schmidhuber},
booktitle={The Twelfth International Conference on Learning Representations},
year={2024},
url={https://openreview.net/forum?id=VtmBAGCN7o}
}

@inproceedings{
wu2024autogen,
title={AutoGen: Enabling Next-Gen {LLM} Applications via Multi-Agent Conversations},
author={Qingyun Wu and Gagan Bansal and Jieyu Zhang and Yiran Wu and Beibin Li and Erkang Zhu and Li Jiang and Xiaoyun Zhang and Shaokun Zhang and Jiale Liu and Ahmed Hassan Awadallah and Ryen W White and Doug Burger and Chi Wang},
booktitle={First Conference on Language Modeling},
year={2024},
url={https://openreview.net/forum?id=BAakY1hNKS}
}

@inproceedings{
chen2024agentverse,
title={AgentVerse: Facilitating Multi-Agent Collaboration and Exploring Emergent Behaviors},
author={Weize Chen and Yusheng Su and Jingwei Zuo and Cheng Yang and Chenfei Yuan and Chi-Min Chan and Heyang Yu and Yaxi Lu and Yi-Hsin Hung and Chen Qian and Yujia Qin and Xin Cong and Ruobing Xie and Zhiyuan Liu and Maosong Sun and Jie Zhou},
booktitle={The Twelfth International Conference on Learning Representations},
year={2024},
url={https://openreview.net/forum?id=EHg5GDnyq1}
}

@misc{li2023camelcommunicativeagentsmind,
      title={CAMEL: Communicative Agents for "Mind" Exploration of Large Language Model Society}, 
      author={Guohao Li and Hasan Abed Al Kader Hammoud and Hani Itani and Dmitrii Khizbullin and Bernard Ghanem},
      year={2023},
      eprint={2303.17760},
      archivePrefix={arXiv},
      primaryClass={cs.AI},
      url={https://arxiv.org/abs/2303.17760}, 
}

@inproceedings{qian-etal-2024-chatdev,
    title = "{C}hat{D}ev: Communicative Agents for Software Development",
    author = "Qian, Chen  and
      Liu, Wei  and
      Liu, Hongzhang  and
      Chen, Nuo  and
      Dang, Yufan  and
      Li, Jiahao  and
      Yang, Cheng  and
      Chen, Weize  and
      Su, Yusheng  and
      Cong, Xin  and
      Xu, Juyuan  and
      Li, Dahai  and
      Liu, Zhiyuan  and
      Sun, Maosong",
    editor = "Ku, Lun-Wei  and
      Martins, Andre  and
      Srikumar, Vivek",
    booktitle = "Proceedings of the 62nd Annual Meeting of the Association for Computational Linguistics (Volume 1: Long Papers)",
    month = aug,
    year = "2024",
    address = "Bangkok, Thailand",
    publisher = "Association for Computational Linguistics",
    url = "https://aclanthology.org/2024.acl-long.810/",
    doi = "10.18653/v1/2024.acl-long.810",
    pages = "15174--15186"
}

@inproceedings{islam-etal-2024-mapcoder,
    title = "{M}ap{C}oder: Multi-Agent Code Generation for Competitive Problem Solving",
    author = "Islam, Md. Ashraful  and
      Ali, Mohammed Eunus  and
      Parvez, Md Rizwan",
    editor = "Ku, Lun-Wei  and
      Martins, Andre  and
      Srikumar, Vivek",
    booktitle = "Proceedings of the 62nd Annual Meeting of the Association for Computational Linguistics (Volume 1: Long Papers)",
    month = aug,
    year = "2024",
    address = "Bangkok, Thailand",
    publisher = "Association for Computational Linguistics",
    url = "https://aclanthology.org/2024.acl-long.269/",
    doi = "10.18653/v1/2024.acl-long.269",
    pages = "4912--4944"
}

@misc{zeng2026aiforsciencelowcodeplatformbayesian,
      title={AI-for-Science Low-code Platform with Bayesian Adversarial Multi-Agent Framework}, 
      author={Zihang Zeng and Jiaquan Zhang and Pengze Li and Yuan Qi and Xi Chen},
      year={2026},
      eprint={2603.03233},
      archivePrefix={arXiv},
      primaryClass={cs.AI},
      url={https://arxiv.org/abs/2603.03233}, 
}

@misc{nguyen2024agilecoderdynamiccollaborativeagents,
      title={AgileCoder: Dynamic Collaborative Agents for Software Development based on Agile Methodology}, 
      author={Minh Huynh Nguyen and Thang Phan Chau and Phong X. Nguyen and Nghi D. Q. Bui},
      year={2024},
      eprint={2406.11912},
      archivePrefix={arXiv},
      primaryClass={cs.SE},
      url={https://arxiv.org/abs/2406.11912}, 
}

@misc{ye2025masgpttrainingllmsbuild,
      title={MAS-GPT: Training LLMs to Build LLM-based Multi-Agent Systems}, 
      author={Rui Ye and Shuo Tang and Rui Ge and Yaxin Du and Zhenfei Yin and Siheng Chen and Jing Shao},
      year={2025},
      eprint={2503.03686},
      archivePrefix={arXiv},
      primaryClass={cs.CL},
      url={https://arxiv.org/abs/2503.03686}, 
}

@misc{tang2025autoagentfullyautomatedzerocodeframework,
      title={AutoAgent: A Fully-Automated and Zero-Code Framework for LLM Agents}, 
      author={Jiabin Tang and Tianyu Fan and Chao Huang},
      year={2025},
      eprint={2502.05957},
      archivePrefix={arXiv},
      primaryClass={cs.AI},
      url={https://arxiv.org/abs/2502.05957}, 
}

@misc{openai2025gpt52,
  author = {OpenAI},
  title = {Introducing GPT-5.2},
  year = {2025},
  url = {https://openai.com/index/introducing-gpt-5-2/}
}







\newpage
\appendix

\section{Prompt of Our Agents}
\label{prompt agents}
\begin{promptbox}[Prompt for Agent1]

\begin{lstlisting}
<system>
You are an Algorithm Selection assistant. 
Output strictly machine-parsable Options and Rationales.

<user>
You are an Algorithm Selection assistant for competitive programming problems.

Scene / Task:
Given the following problem statement delimited by triple backticks, analyze the problem and propose one or more algorithmic approaches and the data structures required to implement them.

Problem statement:
{{STATEMENT}}

Requirements for your response (IMPORTANT, follow exactly):
1. Output only in English.
2. Propose at least one Option; you may propose multiple Options (Option1, Option2, ...). For each Option, list an ordered array of tags (simple short tag strings). Example format (must be parseable):

Option1: [tag1, tag2 ...]
Option2: [tag4, tag5 ...]

3. Each solution can have one or more algorithm labels, striving for accuracy and prioritizing quality over quantity.
4. For each Option, also provide a short 1-3 sentence rationale immediately after the Option line.
5. Do not output extra commentary or notes. Only output Options and their Rationales.
6. If multiple algorithms are viable under different constraints, present them as separate Options. Generally speaking, if a problem can be analyzed from a mathematical perspective, there should be a option that includes math-related tags; if a problem has a brute-force solution (if time constraints are not considered), there should be a option that includes 'brute force'.
7. You SHOULD consult the list of standardized tags and short descriptions in the file:
{{TAG_FILE}}
Prefer tags exactly as presented in that file when applicable.

Extraction rules (must follow):
- Each Option must be on its own line beginning with `Option` + number + colon.
- The tag list must be enclosed in square brackets.
- The rationale must be on the next line and start with `Rationale:`.
- No other lines are allowed.

Now produce the Options.

\end{lstlisting}
\end{promptbox}

\begin{promptbox}[Prompt for Agent2]

\begin{lstlisting}
<system>
You summarize algorithmic domain knowledge.

<user>
You are an expert in competitive programming.
Based on the following reference materials from OI Wiki, 
summarize the key algorithms, data structures, or techniques 
that are relevant to solving the given problem.\n\n
Requirements:
- Only summarize what is clearly supported by the materials
- Be concise and structured (bullet points preferred)
- Do NOT invent content
Problem:\n{statement}
Reference Materials:\n{context_text}

\end{lstlisting}

\end{promptbox}

\begin{promptbox}[Prompt for Agent3 (mode Reasoning)]

\begin{lstlisting}
<system>
You are a crisp step-by-step problem solver, produce numbered steps and pseudocode as required.

<user>
You are a Logical Reasoner: produce a step-by-step complete solution plan for a competitive programming problem.
Use the following inputs as REFERENCE only (you may adopt, refine, or ignore them if you propose a clearer method).

== Problem statement ==
{statement}

== Agent1 (Algorithm Selector) reference ==
{tags_block}

== Agent2 (Domain Knowledge) reference ==
{domain_text if domain_text else "(no domain knowledge provided)"}

Requirements for the final output (follow exactly):
- Output in English.
- Provide your reasoning using the following format:
 Analyze: xxx \n
 Solve step:
 1.xxx;
 2.xxx;
 3.xxx;
- Your thought process and reasoning process should be fully included in the 'analyze' section (Describe the full solution process: algorithm choice, data structures, preprocessing, core loop, correctness proof idea or invariant, complexity analysis (time & memory), edge cases.).
- In 'Solve step' section:
  Each step must start with the step number and a dot, contain a clear natural-language explanation and/or pseudocode.
  Each step MUST end with a semicolon character ';'.
  Each step should be an action that needs to be performed, without requiring analysis or explanation(analysis should be in 'analyze' section).
- Pseudocode may be included inline or inside triple backticks labeled as pseudocode.
- Be concise but complete (avoid long essays). Aim for 10-25 steps if needed.
- Do NOT include any other unnumbered commentary.

Now produce the numbered steps (English). Use the Agent1 and Agent2 references only as guidance.

\end{lstlisting}

\end{promptbox}

\begin{promptbox}[Prompt for Agent3 (mode Replanning)]

\begin{lstlisting}
<system>
You are a rigorous problem solver who can abandon incorrect approaches and rebuild a correct one.

<user>
You are a Logical Reasoner tasked with RE-PLANNING a solution strategy.

IMPORTANT:
- The previous solution plan and implementation have FAILED.
- You must critically re-evaluate the problem and design a NEW strategy.
- You may partially reuse ideas ONLY IF they are still valid.
- Do NOT blindly patch the old plan.

== Problem statement ==
{statement}

== Previous (failed) solution plan ==
{prev_reasoning}

== Failed C++ implementation ==
```cpp
{code}
```

== Error analysis feedback ==
{checker_feedback}

Requirements for the final output (follow exactly):
- Output in English.
- Provide your reasoning using the following format:

 Analyze: xxx

 Solve step:
 1.xxx;
 2.xxx;
 3.xxx;

- Your thought process and reasoning process should be fully included in the 'Analyze' section.
  Describe clearly:
    * why the previous plan failed,
    * what core assumption or algorithm was wrong,
    * what new idea or model should be used,
    * correctness intuition,
    * time & memory complexity,
    * important edge cases.

- In 'Solve step' section:
  Each step must start with the step number and a dot.
  Each step must end with a semicolon ';'.
  Steps should be executable actions, not analysis.

- Do NOT include any other unnumbered commentary.
- Do NOT mention the word "Agent".

(When you encounter discrepancies between the sample output and the simulation results, reread the problem and re-understand its meaning. If your understanding is inconsistent with the output, please revise your understanding and take the sample as the standard, as the sample is absolutely correct.)

Now, produce a NEW complete solution plan.

\end{lstlisting}

\end{promptbox}

\begin{promptbox}[Prompt for Agent4 (mode Implementing)]

\begin{lstlisting}
<system>
You write clean, correct, competitive-programming C++ code.

<user>
You are a Code Implementer for competitive programming.

Your task:
- Generate a COMPLETE, CORRECT, and JUDGE-READY C++17 program.
- The code must fully solve the problem.
- The solution steps below are provided as reference.
  You may refine details if needed, but must stay logically consistent.

===== Problem Statement =====
{statement}

===== Reference Solution Steps (from Logical Reasoner) =====
{reasoner_text}

===== Output Requirements (STRICT) =====
- Output ONLY valid C++17 source code.
- Include: #include <bits/stdc++.h>
- Use standard input/output (cin/cout or scanf/printf).
- Include main() function.
- Do NOT include explanations outside the code.
- Do NOT use markdown or code fences.
- Ensure the code can be compiled and judged directly.

Now produce the final C++17 code.

\end{lstlisting}

\end{promptbox}

\begin{promptbox}[Prompt for Agent4 (mode Revising)]

\begin{lstlisting}
<system>
You are an expert competitive programmer. 
You fix wrong solutions decisively and produce correct code.

<user>
You are a senior competitive programming expert.

Your task is to FIX a WRONG C++ solution using checker feedback.

Rules:
- The previous code is INCORRECT.
- You are allowed to significantly modify or COMPLETELY REWRITE the code.
- Correctness is the highest priority.
- If the original approach is flawed, redesign the solution.
- The reference reasoning is for context only and may be ignored.

===== Problem Statement =====
{statement}

===== Reference Reasoning (optional) =====
{reasoner_text}

===== Wrong C++ Code =====
{wrong_code}

===== Checker Feedback =====
{checker_feedback}

===== Output Requirements (STRICT) =====
- Output ONLY valid C++17 source code.
- Include: #include <bits/stdc++.h>
- Use standard input/output.
- Include main() function.
- Do NOT include explanations or markdown.
- The output must be directly compilable and judge-ready.

Now produce the FIXED C++17 code.

\end{lstlisting}

\end{promptbox}

\begin{promptbox}[Prompt for Agent5]

\begin{lstlisting}
<system>
You are an expert code reviewer for algorithmic problems.
You are given:
- problem statement
- reference reasoning (Agent3)
- C++ code (Agent4)
- official samples (input/output)
- judge results
Your task:
1. With examples, analyze the judge result carefully. You can perform a brief simulation.
2. Identify the root cause of the failure.
3. Decide whether the issue is:
   - minor and fixable locally (FIX) (You need to provide suggestions for correcting errors in the code, and provide the corrected code if necessary.)
   - fundamental algorithmic mistake requiring rethinking (RETHINK) (If a Wrong Answer (WA) or Time-Limited Query (TLE) occurs due to a logical error in the entire code or an incorrect algorithm selection, requiring extensive code modifications or even refactoring, you can suggest a rethinking approach. In this case, describe the code's flaws in detail and briefly provide suggested modifications.)
4. When you encounter discrepancies between the sample output and the simulation results, reread the problem and re-understand its meaning. If your understanding is inconsistent with the output, please revise your understanding and take the sample as the standard, as the sample is absolutely correct.
Guidelines:
- CE: look for syntax, missing headers, wrong types.
- RE: look for division by zero, out-of-bound access, invalid memory.
- TLE: analyze time complexity and bottlenecks.
- WA: simulate on provided samples and find logic flaws.
- AC: This indicates that the current code passes the sample input without any issues. At this point, evaluate whether the code will have logical errors under other specific sample inputs, or whether it will time out under large-scale conditions. If there are no issues, you can choose to pass.
Output format strictly:
analyze:
<your analysis>
error:
<concise description of the error>
fix:
<detailed fix suggestion or rewrite guidance>
signal: PASS | FIX | RETHINK

<user>
Problem:\n{problem_statement}
Reference reasoning (Agent3):\n{reasoning_text}
C++ code:\n```cpp\n{cpp_code}\n```
Samples:{samples_text}
Judge result:
status: {judge_result.get('status')}
passed: {judge_result.get('passed')}
total: {judge_result.get('total')}
info:\n{judge_result.get('info')}

\end{lstlisting}

\end{promptbox}

\section{Samples of Files}
\label{samples files}

\begin{promptbox}[oi wiki tags.txt]

\begin{lstlisting}
segment_tree: Segment tree - supports range queries and range or point updates in O(log n) per operation; supports lazy propagation and can be combined with other data structures.
fenwick_tree: Fenwick tree (Binary Indexed Tree, BIT) - supports prefix queries and point updates in O(log n); simpler than segment tree for prefix sums.
union_find: Disjoint set union (Union-Find, DSU) - maintain connected components with union and find, supports union-by-rank and path compression.
dfs: Depth-first search (DFS) - graph/tree traversal useful for connectivity, tree DP, finding cycles, and ordering.
bfs: Breadth-first search (BFS) - shortest paths on unweighted graphs, level-order traversal.
dijkstra: Dijkstra's algorithm - single-source shortest path for non-negative weighted graphs using a priority queue.
bellman_ford: Bellman-Ford - single-source shortest paths allowing negative weights (detects negative cycles), O(VE).
spfa: SPFA - queue-based improvement sometimes used for shortest paths with negative weights (practical heuristic; can be slow worst-case).
astar: A* search - shortest path with heuristic, used in pathfinding with admissible heuristics.
topological_sort: Topological sort - ordering of DAG vertices, used for DP on directed acyclic graphs.
max_flow: Max flow / Dinic / Edmonds-Karp - compute maximum flow in networks; Dinic is commonly used in contests.
min_cost_max_flow: Minimum cost maximum flow - flow with costs solved by successive shortest augmenting paths.
kruskal: Kruskal's MST - minimum spanning tree using DSU and sorting edges.
prim: Prim's MST - minimum spanning tree using priority queue starting from one node.
binary_search: Binary search on monotone predicate - find threshold values on sorted or implicit domains.
two_pointers: Two pointers (sliding window) - used for subarray or subsequence problems.
monotonic_queue: Monotonic queue/deque - maintain min/max over sliding windows in O(1) amortized.
stack: Stack techniques (monotonic stack) - next greater/smaller element, histogram problems.
queue: Queue techniques - BFS and sliding-window order maintenance.
priority_queue: Priority queue (heap) - used for Dijkstra, scheduling, K-best problems.
hash_table: Hash table / unordered_map - average O(1) lookup/insert; useful for counting and mapping.
trie: Trie (prefix tree) - efficient prefix queries on strings.
kmp: KMP / prefix-function - linear-time string pattern search.
suffix_array: Suffix array / suffix automaton - substring queries, distinct substrings.
rolling_hash: Rolling hash (Rabin-Karp) - string hashing for substring comparison.
fft: Fast Fourier Transform / NTT - convolution for polynomial multiplication.
dp: Dynamic Programming - includes knapsack, tree DP, interval DP, digit DP.
bitmask_dp: Bitmask DP - DP over subsets (n  <= 20).
meet_in_the_middle: Meet-in-the-middle - split into halves, combine results.
greedy: Greedy algorithms - locally optimal choices with global optimality.
divide_and_conquer: Divide and conquer - recursive splitting with merge.
parallel_binary_search: Parallel binary search - offline divide-and-conquer on answer space.
sqrt_decomposition: Mo's algorithm / sqrt decomposition - offline query processing.
heavy_light: Heavy-Light Decomposition (HLD) - path queries on trees.
lowest_common_ancestor: Lowest Common Ancestor (LCA) - binary lifting or RMQ-based.
patience_sorting: Longest Increasing Subsequence (LIS) in O(n log n).
persistent_structure: Persistent data structures - persistent segment tree, etc.
randomization: Randomized algorithms - hashing, randomized selection.
geometry: Computational geometry - convex hull, rotating calipers, line sweep.
math_number_theory: Number theory - gcd, inverse, CRT, sieve, primes.
implement: Steps to simulate problems
math: Using mathematics to derive conclusions.
string: String algorithms.
graph: Grpah.
tree: Tree.

\end{lstlisting}

\end{promptbox}

\begin{promptbox}[failure report.md]

\begin{lstlisting}
## Common LLM Failure Modes on Algorithmic Problems (from `data/failure_analysis.csv`)

This report summarizes recurring failure patterns observed in the `failure_analysis` entries. The themes below are phrased as *problem-solving mistakes* (not language-specific bugs) and are common across WA/TLE/RE/CE outcomes.

1. **Misreading the problem statement's semantics**
   - Treating "looks like X" as "is X" without verifying against the statement/examples.
   - Examples seen: interpreting a `0/1` string as **binary** when the task treats it as a **decimal string**; confusing "a palindromic substring exists" with "the whole string is a palindrome".

2. **Wrong output target / protocol misunderstanding**
   - Computing a related quantity but not what the judge asks for (or missing required resets).
   - Examples seen: outputting the number of distinct strings instead of "how many were deleted since last dedup, then reset"; printing one test case when input is "until EOF"; insufficient floating output precision.

3. **Core modeling error (choosing the wrong abstract problem)**
   - Reframing the task into an unrelated classic template and building a solution on the wrong structure.
   - Examples seen: turning a "choose one from small sets with global uniqueness" task (bipartite matching/SDR) into topological sorting; treating a general directed graph as a DAG and running topo-DP; treating unlabelled rooted-tree counting as an ordered fixed-arity product DP.

4. **Ignoring or weakening quantifiers / constraint scope**
   - Dropping "for all" / "exists a reordering" / "global operation" nuances and solving an easier surrogate.
   - Examples seen: assuming local adjacency heuristics imply a global impossibility; treating global substitutions as independent per-character mappings while ignoring cycle-breaking constraints.

5. **Invalid monotonicity assumptions (two-pointers / sliding window misuse)**
   - Applying two pointers where feasibility is not monotone, or where the maintained predicate is not actually the target predicate.
   - Examples seen: sliding windows on arrays with negative values; "2D conditions" incorrectly handled with two pointers where the compared quantities don't come from the same `(i, j)` cut.

6. **State design errors in DP / memoization**
   - Omitting variables that influence future transitions, causing invalid pruning/merging of states.
   - Examples seen: memoizing only by used-spell masks while ignoring multiplier/health/last-action state; collapsing subsequences by "last value only" and then applying non-linear aggregation (e.g., cubing), which destroys necessary identity information.

7. **Local greedy/inversion without a global invariant**
   - Greedily applying reversible-looking edits (or left-to-right construction) without enforcing global constraints like exact length, feasibility, or bijection consistency.
   - Typical symptoms: "truncate to length \(n\)" after expansions; always picking one ambiguous parse option first; assuming independent extrema can be combined into a feasible interval under one matching.

8. **Constraint-driven feasibility checks are missing**
   - Verifying only the main equation/condition but not the full constraint set (ranges, positivity, non-zero, uniqueness, etc.).
   - Examples seen: constructions that satisfy \(b \oplus c = a\) but produce \(0\) or out-of-range values; failing to special-case powers of two where a naive choice makes a component zero.

9. **Complexity blindness (asymptotically or constant-factor)**
   - Using solutions that are theoretically or practically infeasible under constraints (or implementing an OK idea with catastrophic overhead).
   - Examples seen: exponential bitmask DP with \(n,m\) up to 100; per-query sorting in large query workloads; naive nested scans that devolve to \(O(n^2)\); heavy string-key hashing in huge state spaces.

10. **I/O parsing and stream-lifecycle mistakes**
   - Reading the wrong shape of input (wrong number of tokens/lines), mixing `getline` with formatted reads incorrectly, or forgetting that stdin is a one-pass stream.
   - Examples seen: consuming an extra line for a token that is on the same line; infinite/expensive loops skipping empty lines; attempting "read once, then process again" without caching operations.

11. **Implementation fragility leading to RE/UB**
   - Out-of-bounds arrays, hard-coded limits that contradict constraints, recursion depth blow-ups, unsafe conversions.
   - Examples seen: allocating `dist[..., 1<<10]` but iterating masks up to `1<<m`; fixed factorial tables too small for \(\sum a_i\); deep recursion chains causing stack overflow; `stoi` on arbitrarily long substrings causing exceptions.

12. **Numeric pitfalls (overflow, modulo misuse, precision)**
   - Using too-small integer types, performing intermediate computations in `int`, applying `% MOD` where the quantity is not meant to be reduced, or printing too few decimals.
   - Examples seen: `a+b` overflow in `int`; scanning only 0..30 bits for 64-bit problems; taking a value modulo \(MOD\) even when the definition is an exact integer; `setprecision(6)` under a \(1e{-6}\) error requirement.

13. **Incomplete submission / code not executable (CE / "system error")**
   - Missing braces, unfinished `main`, missing I/O/output entirely, or truncated switch/case logic.
   - This appears repeatedly as a "didn't finish the solution as runnable code" failure mode independent of the underlying algorithm idea.


\end{lstlisting}
\end{promptbox}

\section{Dataset}

\begin{figure*}[t]
    \centering
    \includegraphics[width=\textwidth]{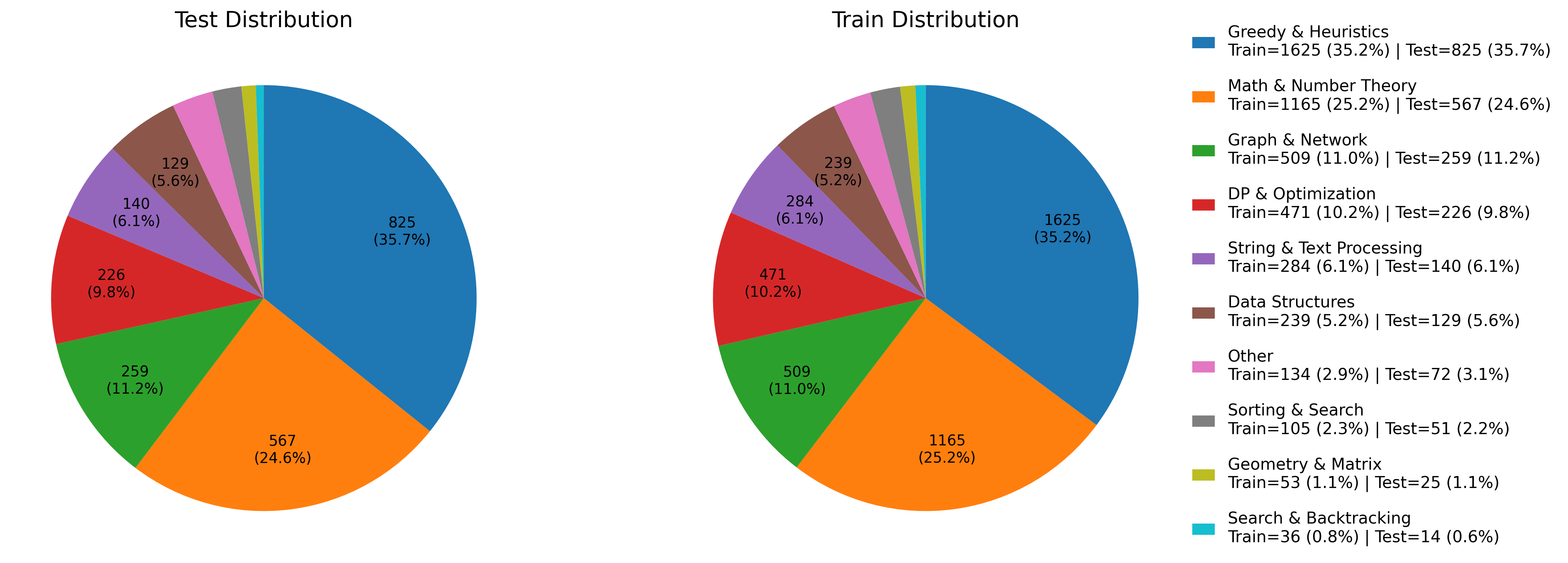}
    \caption{The distribution of different question types in the training and test sets.}
    \label{fig:dataset}
\end{figure*}

We primarily tested on our self-built dataset, which comes from Alibaba's internal online judge dataset. Additionally, we selected LiveCodeBench-Pro datasets from Q4 2024, Q1 2025, and Q2 2025 for testing.
Figure~\ref{fig:dataset} shows the label partitioning of the training and test sets in our dataset.Table~\ref{tab:tag_distribution} illustrates the detailed distribution of all sub-labels.

\begin{table}[htbp]
\caption{Tag distribution in train/test split.}
\centering
\small
\begin{tabular}{lrrrrr}

\hline
Tag & Train Count & Test Count & Total & Train \% & Test \% \\
\hline
astar & 3 & 0 & 3 & 100.00\% & 0.00\% \\
backtracking & 2 & 1 & 3 & 66.67\% & 33.33\% \\
bfs & 45 & 21 & 66 & 68.18\% & 31.82\% \\
binary\_search & 91 & 45 & 136 & 66.91\% & 33.09\% \\
bitmask\_dp & 52 & 23 & 75 & 69.33\% & 30.67\% \\
bitwise & 1 & 0 & 1 & 100.00\% & 0.00\% \\
brute\_force & 2 & 1 & 3 & 66.67\% & 33.33\% \\
bruteforce & 2 & 1 & 3 & 66.67\% & 33.33\% \\
constructive & 2 & 2 & 4 & 50.00\% & 50.00\% \\
constructive\_algorithms & 2 & 1 & 3 & 66.67\% & 33.33\% \\
dfs & 170 & 93 & 263 & 64.64\% & 35.36\% \\
dijkstra & 19 & 9 & 28 & 67.86\% & 32.14\% \\
divide\_and\_conquer & 32 & 17 & 49 & 65.31\% & 34.69\% \\
dp & 395 & 192 & 587 & 67.29\% & 32.71\% \\
fenwick\_tree & 18 & 8 & 26 & 69.23\% & 30.77\% \\
fft & 14 & 7 & 21 & 66.67\% & 33.33\% \\
game & 2 & 2 & 4 & 50.00\% & 50.00\% \\
game\_theory & 8 & 5 & 13 & 61.54\% & 38.46\% \\
geometry & 51 & 25 & 76 & 67.11\% & 32.89\% \\
graph & 188 & 93 & 281 & 66.90\% & 33.10\% \\
greedy & 416 & 210 & 626 & 66.45\% & 33.55\% \\
hash\_table & 103 & 60 & 163 & 63.19\% & 36.81\% \\
heavy\_light & 3 & 2 & 5 & 60.00\% & 40.00\% \\
implement & 1204 & 613 & 1817 & 66.26\% & 33.74\% \\
kmp & 8 & 3 & 11 & 72.73\% & 27.27\% \\
kruskal & 8 & 4 & 12 & 66.67\% & 33.33\% \\
lowest\_common\_ancestor & 6 & 4 & 10 & 60.00\% & 40.00\% \\
math & 833 & 412 & 1245 & 66.91\% & 33.09\% \\
math\_number\_theory & 319 & 148 & 467 & 68.31\% & 31.69\% \\
matrix & 1 & 0 & 1 & 100.00\% & 0.00\% \\
matrix\_exponentiation & 1 & 0 & 1 & 100.00\% & 0.00\% \\
max\_flow & 19 & 8 & 27 & 70.37\% & 29.63\% \\
meet\_in\_the\_middle & 5 & 2 & 7 & 71.43\% & 28.57\% \\
min\_cost\_max\_flow & 6 & 3 & 9 & 66.67\% & 33.33\% \\
monotonic\_queue & 11 & 3 & 14 & 78.57\% & 21.43\% \\
parallel\_binary\_search & 1 & 0 & 1 & 100.00\% & 0.00\% \\
patience\_sorting & 3 & 3 & 6 & 50.00\% & 50.00\% \\
persistent\_structure & 4 & 2 & 6 & 66.67\% & 33.33\% \\
prim & 1 & 0 & 1 & 100.00\% & 0.00\% \\
priority\_queue & 37 & 21 & 58 & 63.79\% & 36.21\% \\
probability & 1 & 1 & 2 & 50.00\% & 50.00\% \\
queue & 21 & 11 & 32 & 65.62\% & 34.38\% \\
randomization & 12 & 6 & 18 & 66.67\% & 33.33\% \\
recursion & 1 & 0 & 1 & 100.00\% & 0.00\% \\
rolling\_hash & 12 & 5 & 17 & 70.59\% & 29.41\% \\
segment\_tree & 35 & 17 & 52 & 67.31\% & 32.69\% \\
simulation & 1 & 0 & 1 & 100.00\% & 0.00\% \\
sort & 14 & 6 & 20 & 70.00\% & 30.00\% \\
spfa & 13 & 6 & 19 & 68.42\% & 31.58\% \\
sqrt\_decomposition & 9 & 4 & 13 & 69.23\% & 30.77\% \\
stack & 21 & 10 & 31 & 67.74\% & 32.26\% \\
string & 252 & 124 & 376 & 67.02\% & 32.98\% \\
suffix\_array & 3 & 3 & 6 & 50.00\% & 50.00\% \\
ternary\_search & 1 & 0 & 1 & 100.00\% & 0.00\% \\
topological\_sort & 7 & 3 & 10 & 70.00\% & 30.00\% \\
tree & 68 & 37 & 105 & 64.76\% & 35.24\% \\
trie & 9 & 5 & 14 & 64.29\% & 35.71\% \\
two\_pointers & 29 & 13 & 42 & 69.05\% & 30.95\% \\
union\_find & 24 & 13 & 37 & 64.86\% & 35.14\% \\
TOTAL & 4621 & 2308 & 6929 & 66.69\% & 33.31\% \\
\hline
\end{tabular}

\label{tab:tag_distribution}
\end{table}
Our label categories are as follows:

1. Graph \& Network Algorithms

Handling problems related to graph structures, shortest paths, minimum spanning trees, and flows:

bfs, dfs, dijkstra, kruskal, prim, lowest\_common\_ancestor, max\_flow, min\_cost\_max\_flow, spfa, topological\_sort, heavy\_light, union\_find, graph

2. Search \& Backtracking

Exhaustive or optimized search problems:

backtracking, recursion, astar, parallel\_binary\_search, two\_pointers

3. Dynamic Programming \& State Compression (DP \& Optimization)

Solving optimal substructure problems:

dp, bitmask\_dp, monotonic\_queue, meet\_in\_the\_middle, ternary\_search, patience\_sorting, constructive, constructive\_algorithms

4. Mathematics \& Number Theory Theory

Including probability, combinatorics, number theory, and randomized algorithms:

math, math\_number\_theory, probability, randomization

5. String \& Text Processing

Handling string matching, hashing, suffix arrays, etc.:

string, kmp, rolling\_hash, suffix\_array, trie

6. Data Structures

Basic and advanced data structures and their applications:

segment\_tree, fenwick\_tree, persistent\_structure, priority\_queue, stack, queue, hash\_table

7. Sorting \& Search Optimization

Optimizing sorting and search strategies:

sort, binary\_search, patience\_sorting, ternary\_search

8. Geometry \& Matrix

Spatial computation and matrix operations:

geometry, matrix, matrix\_exponentiation

9. Greedy \& Heuristics

Suitable for problems with local optima:

greedy, simulation, implement, brute\_force, bruteforce

\section{Experimental Resources}

We conducted our experiments using internal production servers. 
The MAS-Algorithm ran for approximately 15 hours on the dataset with 8 concurrent connections.
For the training data, we used an 8-GPU cluster of H20 processors (140GB*8) to fine-tune Qwen3-Coder-30B-A3B-Instruct. Some training parameters are as Table~\ref{tab:lora-params}
\begin{table}[htbp]
\centering
\caption{Hyperparameter settings for LoRA fine-tuning.}
\label{tab:lora-params}
\begin{tabular}{ll}
\toprule
\textbf{Hyperparameter} & \textbf{Value} \\
\midrule
Base Model              & Qwen3-Coder-30B-A3B-Instruct \\
Fine-tuning Method      & LoRA (Standard) \\
Target Modules          & q, k, v, o, gate, up, down\_proj \\
LoRA Rank ($r$)         & 16 \\
LoRA Alpha ($\alpha$)   & 32 \\
LoRA Dropout            & 0.05 \\
Learning Rate           & 5e-5 \\
Training Epochs         & 5 \\
Precision               & FP16 Mixed Precision \\
Optimizer               & AdamW \\
Bias Type               & None \\
Task Type               & Causal Language Modeling \\
\bottomrule
\end{tabular}
\end{table}

\section{LiveCodeBench-Pro's Quarterly Statistics}
Table~\ref{tab:quarter} shows the quarterly benchmark results of Qwen3-Coder-30B-A3B-Instruct on the LiveCodeBench-Pro benchmark. As can be seen from the table, the Direct Ask method shows a significant decrease in performance (only 4.79\%) on the new dataset (2025Q2), but the MAS-Algorithm does not show a significant decrease, demonstrating its good generalization ability.

\begin{table}[t!]
\scriptsize
  \caption{Judging results of Qwen3-Coder-30B-A3B-Instruct on the LiveCodeBench-Pro benchmark for each quarter using Direct Ask and MAS-Algorithm methods.}
  \label{tab:quarter}
  \renewcommand{\arraystretch}{1.2}
\setlength{\tabcolsep}{6pt}
  \centering
  \resizebox{\textwidth}{!}{
  \begin{tabular}{c|c|ccccc}
    \toprule
    \textbf{Quarter}  & \textbf{Type}   & \textbf{AC rate}     & \textbf{case pass rate}  & \textbf{case pass rate}  & \textbf{time aver} & \textbf{mem aver} \\
          &    & \    &   & \textbf{(with weight)}  & \textbf{(s)} & \textbf{(KB)} \\
        \midrule
2020Q4 & Direct Ask & 7.25\% & 8.77\% & 8.57\% & 0.02 & 4.43 \\
& MAS-Algorithm & 8.70\% (+1.45\%) & 12.22\% (+3.45\%) & 12.04\% (+3.47\%) & 0.07 (+0.05) & 4.97 (+0.54) \\
\midrule
2025Q1 & Direct Ask & 9.04\% & 10.98\% & 10.80\% & 0.03 & 6.17 \\
& MAS-Algorithm & 9.64\% (+0.60\%) & 12.56\% (+1.58\%) & 12.49\% (+1.69\%) & 0.02 (-0.01) & 4.63 (-1.54) \\
\midrule
2025Q2 & Direct Ask & 4.79\% & 7.32\% & 6.97\% & 0.00 & 3.55 \\
& MAS-Algorithm & 8.98\% (+4.19\%) & 13.42\% (+6.1\%) & 12.94\% (5.97\%) & 0.04 (+0.04) & 3.79 (+0.24) \\

    \bottomrule
  \end{tabular}
  }
\end{table}

\section{Detailed Failure Analysis}
Table~\ref{tab:failure_class} shows the 13 error types of LLM responses summarized in GPT-5.2, which can be used to construct failure reports.

\begin{table}[t!]
\scriptsize
  \caption{Specific distribution of failure types.}
  \label{tab:failure_class}
  \renewcommand{\arraystretch}{1.2}
\setlength{\tabcolsep}{6pt}
  \centering
  \resizebox{\textwidth}{!}{
  \begin{tabular}{c|c|c}
    \toprule
       \textbf{Category}   & \textbf{Number}  & \textbf{Ratio} \\
       \midrule
Core modeling error (choosing the wrong abstract problem)             & 214 & 28.46\% \\
Complexity blindness (asymptotically or constant-factor)             & 101 & 13.43\% \\
Misreading the problem statement’s semantics                        & 86  & 11.43\% \\
State design errors in DP / memoization                             & 40  & 5.32\% \\
Incomplete submission / code not executable (CE / “system error”)    & 36  & 4.79\% \\
Numeric pitfalls (overflow, modulo misuse, precision)               & 23  & 3.06\% \\
Wrong output target / protocol misunderstanding                      & 20  & 2.66\% \\
Constraint-driven feasibility checks are missing                     & 20  & 2.66\% \\
Local greedy/inversion without a global invariant                    & 19  & 2.53\% \\
I/O parsing and stream-lifecycle mistakes                          & 18  & 2.40\% \\
Ignoring or weakening quantifiers / constraint scope               & 17  & 2.26\% \\
Implementation fragility leading to RE/UB                            & 13  & 1.73\% \\
Invalid monotonicity assumptions (two-pointers / sliding window misuse) & 11 & 1.46\% \\
    \bottomrule
  \end{tabular}
  }
\end{table}

Below is an example of analyzing the cause of failure.This analysis can be used to construct a failure report. For a single issue, the analysis is the "customized failure report" used in Agent5 of the replacement and ablation experiments mentioned in Section~\ref{4.3}.
\begin{promptbox}[GPT's analysis on problem 434]

\begin{lstlisting}
Analysis:
- **Wrong modeling of variables/equations (matrix is transposed conceptually).**  
  Correct formulation: there are **N decisions** (whether to "paint" each meter i once or not), so **N unknowns** \(x_i \in \{0,1\}\). For each position j, you have one equation describing whether it ends up flipped an odd/even number of times. That yields an **N*N** linear system over GF(2).  
  The AC code builds exactly this: for each j, equation includes \(x_j\) (diagonal 1) plus all \(x_c\) such that painting c also flips j (edges c->j).
  The LLM instead treated the **M brush relations as M "operations"** and built an **M*N** matrix, as if each input pair (A,B) were a selectable operation. But (A,B) is not an independent operation; it's just a dependency describing the effect when painting A. This destroys the problem structure.

- **Augmented vector is constructed with the wrong dimension and meaning.**  
  The right-hand side should be length **N** (desired flip per wall cell). The LLM sets 'aug_matrix[i][N] = diff[i]' for 'i in [0..M)', but 'diff' is length N. If M!=N, this is either out-of-bounds logically (or at least semantically wrong). Even when M<=N, it is still incorrect because each equation corresponds to a wall position, not a brush-rule line.

- **Incorrect reachability / counting logic.**  
  If solvable, the number of solutions is \(2^{\text{#free vars}} = 2^{N-\text{rank}}\) **in the correct N*N system**. The LLM computes rank of its wrong M*N matrix, so even if its consistency check happened to pass sometimes, the exponent is unrelated to the real number of free "paint each meter once" choices.

- **Input parsing mismatch with actual format.**  
  The sample input shows spaces between characters (`w w w`). The AC code reads character-by-character (skipping separators). The LLM uses `cin >> initial >> target`, which will only read the first token (`""w""`) instead of N letters when spaces exist, leading to immediate wrong behavior on such tests.

Model's Disadvantage:
- **Failure to correctly map the story constraints to a linear system** (identifying what constitutes an independent decision/unknown vs. a constraint/edge), plus **weak robustness to input tokenization formats** (space-separated characters vs contiguous strings).

Hint:
- Treat "paint meter i (at most once)" as the **only** decision variable \(x_i\). Build an **N*N** GF(2) system where equation j encodes which \(x_i\) flip position j (always includes \(x_j\), plus any i such that rule says painting i also flips j). Parse N chars allowing spaces.


\end{lstlisting}

\end{promptbox}

Ultimately, this error reason is categorized as "Misreading the problem statement’s semantics".

\section{Detailed Category Analysis}
Tables~\ref{tab:category_results_all}, \ref{tab:category_results_8b}, \ref{tab:category_results_14b}, \ref{tab:category_results_480b} and \ref{tab:category_results_235b} present the detailed experimental results of the category analysis for the Qwen3-Coder-30B-A3B-Instruct, Qwen3-8B, Qwen3-14B, Qwen3-Coder-480B-A35B-Instruct, and Qwen3-235B-A22B-Instruct-2507 models mentioned in Section~\ref{4.4}, respectively.
Tables~\ref{tab:category_results_agent0}, \ref{tab:category_results_agent1}, \ref{tab:category_results_agent2}, \ref{tab:category_results_agent3} and \ref{tab:category_results_agent5} present the detailed experimental results of the category analysis for Qwen3-Coder-30B-A3B-Instruct after the replacement and ablation experiments mentioned in Section~\ref{4.3}. The increment/decrement values after parentheses are compared with the MAS-Algorithm performance of the original Qwen3-Coder-30B-A3B-Instruct.

\begin{table}[t!]
\scriptsize
  \caption{Judging results for Qwen3-Coder-30B-A3B-Instruct categorized by question type.}
  \label{tab:category_results_all}
  \renewcommand{\arraystretch}{1.2}
  \setlength{\tabcolsep}{5pt}
  \centering
  \resizebox{\textwidth}{!}{
  \begin{tabular}{c|c|ccccc}
    \toprule
    \textbf{Category} & \textbf{Type} & \textbf{AC rate} & \textbf{case pass rate} & \textbf{case pass rate} & \textbf{time aver} & \textbf{mem aver} \\
     &  &  &  & \textbf{(with weight)} & \textbf{(s)} & \textbf{(KB)} \\
    \midrule
    Graph \& Network
      & Direct Ask & 11.49\% & 12.22\% & 14.63\% & 0.02 & 4.77 \\
      & MAS-Algorithm & 18.24\% (+6.76\%) & 17.90\% (+5.68\%) & 22.05\% (+7.42\%) & 0.15 (+0.13) & 9.38 (+4.61) \\
    \midrule
    Search \& Backtracking
      & Direct Ask & 28.57\% & 36.42\% & 34.33\% & 0.02 & 4.54 \\
      & MAS-Algorithm & 35.71\% (+7.14\%) & 38.02\% (+1.60\%) & 36.45\% (+2.12\%) & 0.14 (+0.12) & 5.55 (+1.01) \\
    \midrule
    DP \& Optimization
      & Direct Ask & 17.39\% & 18.33\% & 21.29\% & 0.74 & 9.61 \\
      & MAS-Algorithm & 22.22\% (+4.83\%) & 20.87\% (+2.53\%) & 26.80\% (+5.51\%) & 0.45 (-0.30) & 9.28 (-0.33) \\
    \midrule
    Math \& Number Theory
      & Direct Ask & 31.79\% & 35.10\% & 35.64\% & 0.10 & 4.29 \\
      & MAS-Algorithm & 35.41\% (+3.62\%) & 38.18\% (+3.08\%) & 38.80\% (+3.16\%) & 0.06 (-0.04) & 5.88 (+1.58) \\
    \midrule
    String \& Text Processing
      & Direct Ask & 39.37\% & 37.70\% & 43.62\% & 0.05 & 3.76 \\
      & MAS-Algorithm & 50.39\% (+11.02\%) & 47.02\% (+9.32\%) & 54.24\% (+10.62\%) & 0.26 (+0.20) & 33.52 (+29.76) \\
    \midrule
    Data Structures
      & Direct Ask & 19.66\% & 20.63\% & 23.73\% & 0.15 & 4.50 \\
      & MAS-Algorithm & 28.21\% (+8.55\%) & 29.45\% (+8.82\%) & 31.15\% (+7.42\%) & 0.60 (+0.44) & 71.58 (+67.08) \\
    \midrule
    Sorting \& Search
      & Direct Ask & 17.65\% & 24.34\% & 22.30\% & 0.36 & 3.60 \\
      & MAS-Algorithm & 25.49\% (+7.84\%) & 24.50\% (+0.16\%) & 28.36\% (+6.06\%) & 1.11 (+0.74) & 143.00 (+139.41) \\
    \midrule
    Geometry \& Matrix
      & Direct Ask & 24.00\% & 23.14\% & 24.48\% & 0.00 & 3.62 \\
      & MAS-Algorithm & 40.00\% (+16.00\%) & 44.15\% (+21.01\%) & 40.33\% (+15.86\%) & 0.05 (+0.05) & 6.43 (+2.80) \\
    \midrule
    Greedy \& Heuristics
      & Direct Ask & 35.23\% & 36.28\% & 39.11\% & 0.15 & 4.05 \\
      & MAS-Algorithm & 41.78\% (+6.55\%) & 41.62\% (+5.34\%) & 45.07\% (+5.96\%) & 0.13 (-0.02) & 6.00 (+1.95) \\
    \bottomrule
  \end{tabular}
  }
\end{table}

\vspace{1.2cm}

\begin{table}[t!]
\scriptsize
  \caption{Judging results for Qwen3-8B categorized by question type.}
  \label{tab:category_results_8b}
  \renewcommand{\arraystretch}{1.2}
  \setlength{\tabcolsep}{5pt}
  \centering
  \resizebox{\textwidth}{!}{
  \begin{tabular}{c|c|ccccc}
    \toprule
    \textbf{Category} & \textbf{Type} & \textbf{AC rate} & \textbf{case pass rate} & \textbf{case pass rate} & \textbf{time aver} & \textbf{mem aver} \\
     & & & & \textbf{(with weight)} & \textbf{(s)} & \textbf{(KB)} \\
    \midrule
    Graph \& Network
      & Direct Ask & 10.14\% & 7.89\% & 11.80\% & 2.85 & 24.09 \\
      & MAS-Algorithm & 16.89\% (+6.75\%) & 16.76\% (+8.87\%) & 19.25\% (+7.45\%) & 0.14 (-2.71) & 17.29 (-6.80) \\
    \midrule
    Search \& Backtracking
      & Direct Ask & 14.29\% & 20.45\% & 18.73\% & 0.01 & 3.83 \\
      & MAS-Algorithm & 14.29\% (+0.00\%) & 29.07\% (+8.62\%) & 26.99\% (+8.26\%) & 0.01 (+0.00) & 3.75 (-0.08) \\
    \midrule
    DP \& Optimization
      & Direct Ask & 12.56\% & 14.43\% & 15.33\% & 1.44 & 10.45 \\
      & MAS-Algorithm & 16.91\% (+4.35\%) & 19.22\% (+4.79\%) & 20.06\% (+4.73\%) & 0.22 (-1.22) & 6.54 (-3.91) \\
    \midrule
    Math \& Number Theory
      & Direct Ask & 24.75\% & 26.68\% & 27.99\% & 0.27 & 4.57 \\
      & MAS-Algorithm & 29.78\% (+5.03\%) & 33.90\% (+7.22\%) & 33.87\% (+5.88\%) & 0.08 (-0.19) & 6.06 (+1.49) \\
    \midrule
    String \& Text Processing
      & Direct Ask & 38.58\% & 35.72\% & 40.81\% & 1.29 & 3.72 \\
      & MAS-Algorithm & 38.58\% (+0.00\%) & 40.21\% (+4.49\%) & 44.23\% (+3.42\%) & 0.01 (-1.28) & 3.96 (+0.24) \\
    \midrule
    Data Structures
      & Direct Ask & 15.38\% & 14.52\% & 18.52\% & 0.13 & 20.83 \\
      & MAS-Algorithm & 20.51\% (+5.13\%) & 23.73\% (+9.21\%) & 24.52\% (+6.00\%) & 0.13 (+0.00) & 37.81 (+16.98) \\
    \midrule
    Sorting \& Search
      & Direct Ask & 9.80\% & 15.79\% & 14.79\% & 0.01 & 3.54 \\
      & MAS-Algorithm & 11.76\% (+1.96\%) & 15.06\% (-0.73\%) & 17.04\% (+2.25\%) & 0.13 (+0.12) & 4.32 (+0.78) \\
    \midrule
    Geometry \& Matrix
      & Direct Ask & 20.00\% & 19.68\% & 20.13\% & 0.01 & 3.53 \\
      & MAS-Algorithm & 28.00\% (+8.00\%) & 25.80\% (+6.12\%) & 28.48\% (+8.35\%) & 0.00 (-0.01) & 3.64 (+0.11) \\
    \midrule
    Greedy \& Heuristics
      & Direct Ask & 28.53\% & 28.86\% & 31.61\% & 0.38 & 5.68 \\
      & MAS-Algorithm & 35.95\% (+7.42\%) & 36.89\% (+8.03\%) & 40.13\% (+8.52\%) & 0.06 (-0.32) & 8.36 (+2.68) \\
    \bottomrule
  \end{tabular}
  }
\end{table}

\vspace{1.2cm}

\begin{table}[t!]
\scriptsize
  \caption{Judging results for Qwen3-14B categorized by question type.}
  \label{tab:category_results_14b}
  \renewcommand{\arraystretch}{1.2}
  \setlength{\tabcolsep}{5pt}
  \centering
  \resizebox{\textwidth}{!}{
  \begin{tabular}{c|c|ccccc}
    \toprule
    \textbf{Category} & \textbf{Type} & \textbf{AC rate} & \textbf{case pass rate} & \textbf{case pass rate} & \textbf{time aver} & \textbf{mem aver} \\
     & & & & \textbf{(with weight)} & \textbf{(s)} & \textbf{(KB)} \\
    \midrule
    Graph \& Network
      & Direct Ask & 12.16\% & 11.18\% & 13.04\% & 0.07 & 6.06 \\
      & MAS-Algorithm & 20.27\% (+8.11\%) & 19.85\% (+8.67\%) & 23.90\% (+10.86\%) & 0.21 (+0.14) & 17.48 (+11.42) \\
    \midrule
    Search \& Backtracking
      & Direct Ask & 21.43\% & 24.28\% & 23.57\% & 0.26 & 5.11 \\
      & MAS-Algorithm & 35.71\% (+14.28\%) & 45.69\% (+21.41\%) & 39.02\% (+15.45\%) & 0.01 (-0.25) & 4.06 (-1.05) \\
    \midrule
    DP \& Optimization
      & Direct Ask & 11.59\% & 12.02\% & 14.81\% & 0.45 & 5.36 \\
      & MAS-Algorithm & 20.29\% (+8.70\%) & 19.85\% (+7.83\%) & 23.59\% (+8.78\%) & 0.56 (+0.11) & 8.82 (+3.46) \\
    \midrule
    Math \& Number Theory
      & Direct Ask & 25.75\% & 28.00\% & 29.21\% & 0.51 & 4.56 \\
      & MAS-Algorithm & 34.41\% (+8.66\%) & 39.35\% (+11.35\%) & 39.55\% (+10.34\%) & 0.15 (-0.36) & 5.29 (+0.73) \\
    \midrule
    String \& Text Processing
      & Direct Ask & 40.94\% & 38.17\% & 43.35\% & 0.12 & 9.72 \\
      & MAS-Algorithm & 50.39\% (+9.45\%) & 50.72\% (+12.55\%) & 54.62\% (+11.27\%) & 0.18 (+0.06) & 5.11 (-4.61) \\
    \midrule
    Data Structures
      & Direct Ask & 23.93\% & 22.76\% & 25.60\% & 4.21 & 10.79 \\
      & MAS-Algorithm & 25.64\% (+1.71\%) & 25.39\% (+2.63\%) & 28.95\% (+3.35\%) & 0.27 (-3.94) & 16.61 (+5.82) \\
    \midrule
    Sorting \& Search
      & Direct Ask & 11.76\% & 20.69\% & 17.27\% & 0.02 & 4.26 \\
      & MAS-Algorithm & 17.65\% (+5.89\%) & 18.15\% (-2.54\%) & 19.78\% (+2.51\%) & 0.06 (+0.04) & 4.26 (+0.00) \\
    \midrule
    Geometry \& Matrix
      & Direct Ask & 16.00\% & 18.62\% & 16.65\% & 0.00 & 3.53 \\
      & MAS-Algorithm & 28.00\% (+12.00\%) & 31.38\% (+12.76\%) & 28.48\% (+11.83\%) & 0.44 (+0.44) & 3.56 (+0.03) \\
    \midrule
    Greedy \& Heuristics
      & Direct Ask & 32.17\% & 33.12\% & 35.38\% & 0.48 & 6.25 \\
      & MAS-Algorithm & 40.90\% (+8.73\%) & 44.41\% (+11.29\%) & 45.62\% (+10.24\%) & 0.14 (-0.34) & 24.95 (+18.70) \\
    \bottomrule
  \end{tabular}
  }
\end{table}

\vspace{1.2cm}

\begin{table}[t!]
\scriptsize
  \caption{Judging results for Qwen3-Coder-480B-A35B-Instruct categorized by question type.}
  \label{tab:category_results_480b}
  \renewcommand{\arraystretch}{1.2}
  \setlength{\tabcolsep}{5pt}
  \centering
  \resizebox{\textwidth}{!}{
  \begin{tabular}{c|c|ccccc}
    \toprule
    \textbf{Category} & \textbf{Type} & \textbf{AC rate} & \textbf{case pass rate} & \textbf{case pass rate} & \textbf{time aver} & \textbf{mem aver} \\
     & & & & \textbf{(with weight)} & \textbf{(s)} & \textbf{(KB)} \\
    \midrule
    Graph \& Network
      & Direct Ask & 31.08\% & 29.48\% & 34.45\% & 1.58 & 30.74 \\
      & MAS-Algorithm & 36.49\% (+5.41\%) & 34.40\% (+4.92\%) & 40.65\% (+6.20\%) & 1.26 (-0.32) & 28.27 (-2.47) \\
    \midrule
    Search \& Backtracking
      & Direct Ask & 50.00\% & 49.84\% & 51.07\% & 0.10 & 4.96 \\
      & MAS-Algorithm & 42.86\% (-7.14\%) & 53.04\% (+3.20\%) & 47.58\% (-3.49\%) & 0.01 (-0.09) & 5.13 (+0.17) \\
    \midrule
    DP \& Optimization
      & Direct Ask & 28.50\% & 27.86\% & 31.39\% & 1.13 & 27.00 \\
      & MAS-Algorithm & 34.30\% (+5.80\%) & 32.12\% (+4.26\%) & 38.21\% (+6.82\%) & 0.55 (-0.58) & 26.49 (-0.51) \\
    \midrule
    Math \& Number Theory
      & Direct Ask & 44.87\% & 45.68\% & 47.70\% & 0.42 & 6.55 \\
      & MAS-Algorithm & 48.89\% (+4.02\%) & 50.65\% (+4.97\%) & 52.27\% (+4.57\%) & 0.20 (-0.22) & 9.85 (+3.30) \\
    \midrule
    String \& Text Processing
      & Direct Ask & 52.76\% & 52.52\% & 56.45\% & 0.28 & 5.14 \\
      & MAS-Algorithm & 64.57\% (+11.81\%) & 62.33\% (+9.81\%) & 69.27\% (+12.82\%) & 0.39 (+0.11) & 5.51 (+0.37) \\
    \midrule
    Data Structures
      & Direct Ask & 35.04\% & 38.82\% & 38.61\% & 2.06 & 7.33 \\
      & MAS-Algorithm & 46.15\% (+11.11\%) & 46.14\% (+7.32\%) & 49.80\% (+11.19\%) & 0.39 (-1.67) & 10.45 (+3.12) \\
    \midrule
    Sorting \& Search
      & Direct Ask & 29.41\% & 28.86\% & 32.03\% & 0.39 & 4.74 \\
      & MAS-Algorithm & 33.33\% (+3.92\%) & 35.57\% (+6.71\%) & 36.31\% (+4.28\%) & 0.63 (+0.24) & 4.53 (-0.21) \\
    \midrule
    Geometry \& Matrix
      & Direct Ask & 32.00\% & 31.65\% & 32.20\% & 0.01 & 5.08 \\
      & MAS-Algorithm & 44.00\% (+12.00\%) & 49.20\% (+17.55\%) & 44.13\% (+11.93\%) & 0.03 (+0.02) & 6.19 (+1.11) \\
    \midrule
    Greedy \& Heuristics
      & Direct Ask & 48.91\% & 49.53\% & 52.64\% & 0.28 & 4.80 \\
      & MAS-Algorithm & 55.02\% (+6.11\%) & 55.51\% (+5.98\%) & 59.35\% (+6.71\%) & 0.10 (-0.18) & 13.13 (+8.33) \\
    \bottomrule
  \end{tabular}
  }
\end{table}

\vspace{1.2cm}

\begin{table}[t!]
\scriptsize
  \caption{Judging results for Qwen3-235B-A22B-Instruct-2507 categorized by question type.}
  \label{tab:category_results_235b}
  \renewcommand{\arraystretch}{1.2}
  \setlength{\tabcolsep}{5pt}
  \centering
  \resizebox{\textwidth}{!}{
  \begin{tabular}{c|c|ccccc}
    \toprule
    \textbf{Category} & \textbf{Type} & \textbf{AC rate} & \textbf{case pass rate} & \textbf{case pass rate} & \textbf{time aver} & \textbf{mem aver} \\
     & & & & \textbf{(with weight)} & \textbf{(s)} & \textbf{(KB)} \\
    \midrule
    Graph \& Network
      & Direct Ask & 38.51\% & 37.87\% & 42.51\% & 0.66 & 18.94 \\
      & MAS-Algorithm & 44.59\% (+6.08\%) & 48.74\% (+10.87\%) & 52.50\% (+9.99\%) & 0.43 (-0.23) & 26.79 (+7.85) \\
    \midrule
    Search \& Backtracking
      & Direct Ask & 50.00\% & 57.03\% & 58.79\% & 0.11 & 5.64 \\
      & MAS-Algorithm & 85.71\% (+35.71\%) & 81.47\% (+24.44\%) & 87.54\% (+28.75\%) & 0.30 (+0.19) & 5.94 (+0.30) \\
    \midrule
    DP \& Optimization
      & Direct Ask & 48.07\% & 45.12\% & 51.35\% & 0.87 & 18.26 \\
      & MAS-Algorithm & 55.07\% (+7.00\%) & 53.89\% (+8.77\%) & 60.88\% (+9.53\%) & 0.63 (-0.24) & 24.56 (+6.30) \\
    \midrule
    Math \& Number Theory
      & Direct Ask & 61.09\% & 64.39\% & 64.61\% & 0.24 & 8.26 \\
      & MAS-Algorithm & 64.99\% (+3.90\%) & 67.77\% (+3.38\%) & 69.36\% (+4.75\%) & 0.14 (-0.10) & 10.50 (+2.24) \\
    \midrule
    String \& Text Processing
      & Direct Ask & 74.02\% & 76.31\% & 75.81\% & 0.23 & 5.07 \\
      & MAS-Algorithm & 72.44\% (-1.58\%) & 74.54\% (-1.77\%) & 75.65\% (-0.16\%) & 0.13 (-0.10) & 6.37 (+1.30) \\
    \midrule
    Data Structures
      & Direct Ask & 52.14\% & 54.86\% & 55.62\% & 0.31 & 8.78 \\
      & MAS-Algorithm & 59.83\% (+7.69\%) & 66.21\% (+11.35\%) & 65.42\% (+9.80\%) & 0.35 (+0.04) & 11.01 (+2.23) \\
    \midrule
    Sorting \& Search
      & Direct Ask & 37.25\% & 40.85\% & 43.74\% & 0.49 & 5.21 \\
      & MAS-Algorithm & 47.06\% (+9.81\%) & 45.19\% (+4.34\%) & 50.68\% (+6.94\%) & 0.86 (+0.37) & 10.62 (+5.41) \\
    \midrule
    Geometry \& Matrix
      & Direct Ask & 48.00\% & 51.06\% & 49.53\% & 0.05 & 5.94 \\
      & MAS-Algorithm & 52.00\% (+4.00\%) & 51.33\% (+0.27\%) & 53.42\% (+3.89\%) & 0.06 (+0.01) & 5.80 (-0.14) \\
    \midrule
    Greedy \& Heuristics
      & Direct Ask & 66.84\% & 67.79\% & 69.81\% & 0.19 & 6.37 \\
      & MAS-Algorithm & 70.16\% (+3.32\%) & 70.88\% (+3.09\%) & 74.12\% (+4.31\%) & 0.07 (-0.12) & 7.88 (+1.51) \\
    \bottomrule
  \end{tabular}
  }
\end{table}

\vspace{1.2cm}

\begin{table}[t!]
\scriptsize
  \caption{Judging results of Qwen3-Coder-30B-A3B-Instruct, which underwent Agent0-type replacement and ablation experiments, categorized by question type.}
  \label{tab:category_results_agent0}
  \renewcommand{\arraystretch}{1.2}
  \setlength{\tabcolsep}{5pt}
  \centering
  \resizebox{\textwidth}{!}{
  \begin{tabular}{c|ccccc}
    \toprule
    \textbf{Category} & \textbf{AC rate} & \textbf{case pass rate} & \textbf{case pass rate (with weight)} & \textbf{time aver (s)} & \textbf{mem aver (KB)} \\
    \midrule
    Graph \& Network & 29.05\% (+10.81\%) & 31.68\% (+13.78\%) & 32.63\% (+10.58\%) & 0.12 (-0.03) & 9.23 (-0.15) \\
    Search \& Backtracking & 57.14\% (+21.43\%) & 69.01\% (+31.11\%) & 65.39\% (+28.94\%) & 0.10 (-0.01) & 5.93 (+0.81) \\
    DP \& Optimization & 32.37\% (+10.15\%) & 33.98\% (+13.11\%) & 35.96\% (+9.16\%) & 0.10 (-0.35) & 25.29 (-1.20) \\
    Math \& Number Theory & 50.30\% (+14.89\%) & 52.90\% (+14.72\%) & 53.68\% (+14.88\%) & 0.04 (-0.02) & 6.69 (-3.16) \\
    String \& Text Processing & 60.63\% (+10.24\%) & 58.13\% (+11.11\%) & 64.23\% (+10.01\%) & 0.05 (-0.21) & 5.85 (+0.34) \\
    Data Structures & 41.03\% (+12.82\%) & 41.11\% (+11.66\%) & 43.79\% (+12.64\%) & 0.07 (-0.32) & 10.08 (-0.37) \\
    Sorting \& Search & 33.33\% (+7.84\%) & 38.36\% (+13.86\%) & 38.25\% (+9.89\%) & 0.12 (-0.99) & 4.35 (-138.65) \\
    Geometry \& Matrix & 48.00\% (+8.00\%) & 50.53\% (+6.38\%) & 51.28\% (+10.95\%) & 0.09 (+0.04) & 5.97 (-0.46) \\
    Greedy \& Heuristics & 55.60\% (+13.82\%) & 56.14\% (+14.52\%) & 59.57\% (+14.50\%) & 0.03 (-0.10) & 5.40 (-0.60) \\
    \bottomrule
  \end{tabular}
  }
\end{table}

\vspace{1.2cm}

\begin{table}[t!]
\scriptsize
  \caption{Judging results of Qwen3-Coder-30B-A3B-Instruct, which underwent Agent1-type replacement and ablation experiments, categorized by question type.}
  \label{tab:category_results_agent1}
  \renewcommand{\arraystretch}{1.2}
  \setlength{\tabcolsep}{5pt}
  \centering
  \resizebox{\textwidth}{!}{
  \begin{tabular}{c|ccccc}
    \toprule
    \textbf{Category} & \textbf{AC rate} & \textbf{case pass rate} & \textbf{case pass rate (with weight)} & \textbf{time aver (s)} & \textbf{mem aver (KB)} \\
    \midrule
    Graph \& Network & 20.95\% (+2.71\%) & 21.85\% (+3.95\%) & 24.48\% (+2.43\%) & 0.96 (+0.81) & 23.90 (+14.52) \\
    Search \& Backtracking & 35.71\% (+0.00\%) & 44.41\% (+6.51\%) & 40.15\% (+3.70\%) & 0.01 (-0.10) & 5.30 (+0.18) \\
    DP \& Optimization & 21.26\% (-1.04\%) & 20.73\% (-0.14\%) & 25.47\% (-1.33\%) & 1.01 (+0.56) & 19.16 (-7.33) \\
    Math \& Number Theory & 36.02\% (+0.61\%) & 39.12\% (+0.94\%) & 40.21\% (+1.41\%) & 0.21 (+0.15) & 7.71 (-2.14) \\
    String \& Text Processing & 48.03\% (-2.36\%) & 44.44\% (-2.58\%) & 50.93\% (-3.29\%) & 0.30 (+0.04) & 4.61 (-28.91) \\
    Data Structures & 29.91\% (-1.30\%) & 33.28\% (-12.83\%) & 33.76\% (-14.43\%) & 0.78 (+0.39) & 6.41 (-65.17) \\
    Sorting \& Search & 25.49\% (+0.00\%) & 25.77\% (+1.27\%) & 29.28\% (+0.92\%) & 0.68 (-0.43) & 4.26 (-138.74) \\
    Geometry \& Matrix & 32.00\% (-8.00\%) & 41.49\% (-2.66\%) & 35.41\% (-4.92\%) & 0.00 (-0.05) & 3.65 (-2.78) \\
    Greedy \& Heuristics & 40.61\% (-1.17\%) & 41.29\% (-0.33\%) & 44.43\% (-0.64\%) & 0.17 (+0.04) & 4.68 (-1.32) \\
    \bottomrule
  \end{tabular}
  }

\end{table}

\vspace{1.2cm}

\begin{table}[t!]
\scriptsize
  \caption{Judging results of Qwen3-Coder-30B-A3B-Instruct, which underwent Agent2-type replacement and ablation experiments, categorized by question type.}
  \label{tab:category_results_agent2}
  \renewcommand{\arraystretch}{1.2}
  \setlength{\tabcolsep}{5pt}
  \centering
  \resizebox{\textwidth}{!}{
  \begin{tabular}{c|ccccc}
    \toprule
    \textbf{Category} & \textbf{AC rate} & \textbf{case pass rate} & \textbf{case pass rate (with weight)} & \textbf{time aver (s)} & \textbf{mem aver (KB)} \\
    \midrule
    Graph \& Network & 42.57\% (+24.33\%) & 34.71\% (+16.81\%) & 45.32\% (+23.27\%) & 0.07 (-0.08) & 6.55 (-2.83) \\
    Search \& Backtracking & 50.00\% (+14.29\%) & 57.83\% (+19.93\%) & 64.04\% (+27.59\%) & 0.08 (-0.03) & 6.22 (+1.10) \\
    DP \& Optimization & 43.48\% (+21.26\%) & 49.23\% (+28.35\%) & 48.83\% (+21.73\%) & 0.09 (-0.36) & 7.16 (-2.17) \\
    Math \& Number Theory & 60.56\% (+25.15\%) & 61.77\% (+23.59\%) & 63.51\% (+24.71\%) & 0.10 (+0.04) & 7.39 (-2.46) \\
    String \& Text Processing & 66.93\% (+16.54\%) & 63.18\% (+16.16\%) & 70.44\% (+16.22\%) & 0.07 (-0.19) & 8.26 (+2.75) \\
    Data Structures & 52.14\% (+23.93\%) & 47.97\% (+18.52\%) & 56.06\% (+24.91\%) & 0.07 (-0.32) & 9.21 (-1.24) \\
    Sorting \& Search & 37.25\% (+11.76\%) & 39.38\% (+14.88\%) & 40.54\% (+12.18\%) & 0.21 (-0.90) & 4.24 (-138.76) \\
    Geometry \& Matrix & 52.00\% (+12.00\%) & 46.28\% (-2.13\%) & 52.48\% (+12.15\%) & 0.02 (-0.03) & 3.64 (-2.79) \\
    Greedy \& Heuristics & 64.92\% (+23.14\%) & 63.99\% (+22.37\%) & 67.97\% (+22.90\%) & 0.08 (-0.05) & 7.35 (+1.35) \\
    \bottomrule
  \end{tabular}
  }

\end{table}

\vspace{1.2cm}

\begin{table}[t!]
\scriptsize
  \caption{Judging results of Qwen3-Coder-30B-A3B-Instruct, which underwent Agent3-type replacement and ablation experiments, categorized by question type.}
  \label{tab:category_results_agent3}
  \renewcommand{\arraystretch}{1.2}
  \setlength{\tabcolsep}{5pt}
  \centering
  \resizebox{\textwidth}{!}{
  \begin{tabular}{c|ccccc}
    \toprule
    \textbf{Category} & \textbf{AC rate} & \textbf{case pass rate} & \textbf{case pass rate (with weight)} & \textbf{time aver (s)} & \textbf{mem aver (KB)} \\
    \midrule
    Graph \& Network & 45.95\% (+27.71\%) & 41.53\% (+23.63\%) & 48.74\% (+26.69\%) & 0.17 (+0.02) & 10.11 (+0.73) \\
    Search \& Backtracking & 57.14\% (+21.43\%) & 60.38\% (+22.48\%) & 63.45\% (+27.00\%) & 0.08 (-0.03) & 7.04 (+1.92) \\
    DP \& Optimization & 46.86\% (+24.64\%) & 49.51\% (+28.63\%) & 50.98\% (+23.88\%) & 0.13 (-0.32) & 11.88 (+2.55) \\
    Math \& Number Theory & 67.61\% (+32.20\%) & 68.08\% (+29.90\%) & 70.46\% (+31.66\%) & 0.06 (-0.00) & 9.49 (-0.36) \\
    String \& Text Processing & 68.50\% (+18.11\%) & 71.46\% (+24.44\%) & 73.50\% (+19.28\%) & 0.02 (-0.24) & 7.62 (+2.11) \\
    Data Structures & 55.56\% (+27.35\%) & 54.15\% (+24.70\%) & 59.95\% (+28.80\%) & 0.13 (-0.26) & 13.12 (+3.67) \\
    Sorting \& Search & 52.94\% (+27.45\%) & 55.72\% (+31.22\%) & 53.98\% (+25.62\%) & 0.07 (-1.04) & 9.11 (-133.89) \\
    Geometry \& Matrix & 64.00\% (+24.00\%) & 63.30\% (+19.15\%) & 64.00\% (+23.67\%) & 0.10 (+0.05) & 9.16 (+2.73) \\
    Greedy \& Heuristics & 69.43\% (+27.65\%) & 68.09\% (+26.47\%) & 72.24\% (+27.17\%) & 0.05 (-0.08) & 7.56 (+1.56) \\
    \bottomrule
  \end{tabular}
  }

\end{table}

\vspace{1.2cm}

\begin{table}[t!]
\scriptsize
  \caption{Judging results of Qwen3-Coder-30B-A3B-Instruct, which underwent Agent5-type replacement and ablation experiments, categorized by question type.}
  \label{tab:category_results_agent5}
  \renewcommand{\arraystretch}{1.2}
  \setlength{\tabcolsep}{5pt}
  \centering
  \resizebox{\textwidth}{!}{
  \begin{tabular}{c|ccccc}
    \toprule
    \textbf{Category} & \textbf{AC rate} & \textbf{case pass rate} & \textbf{case pass rate (with weight)} & \textbf{time aver (s)} & \textbf{mem aver (KB)} \\
    \midrule
    Graph \& Network & 29.05\% (+10.81\%) & 30.58\% (+12.68\%) & 32.70\% (+10.65\%) & 0.55 (+0.40) & 7.77 (-1.61) \\
    Search \& Backtracking & 42.86\% (+7.15\%) & 44.73\% (+6.73\%) & 45.04\% (+8.59\%) & 0.98 (+0.87) & 5.05 (-0.07) \\
    DP \& Optimization & 28.99\% (+6.77\%) & 29.46\% (+8.58\%) & 30.46\% (+3.66\%) & 0.39 (-0.06) & 8.24 (-18.25) \\
    Math \& Number Theory & 45.47\% (+10.06\%) & 48.83\% (+10.65\%) & 48.06\% (+9.26\%) & 0.24 (+0.18) & 6.03 (-3.82) \\
    String \& Text Processing & 58.27\% (+7.88\%) & 56.00\% (+8.98\%) & 60.05\% (+5.83\%) & 0.42 (+0.16) & 29.92 (+24.41) \\
    Data Structures & 35.04\% (+6.83\%) & 35.33\% (+6.88\%) & 38.07\% (+6.92\%) & 0.67 (+0.28) & 52.19 (+41.74) \\
    Sorting \& Search & 43.14\% (+17.65\%) & 47.55\% (+22.85\%) & 46.56\% (+18.20\%) & 0.61 (-0.50) & 86.89 (-56.11) \\
    Geometry \& Matrix & 40.00\% (+0.00\%) & 33.51\% (-10.64\%) & 40.36\% (+0.03\%) & 0.07 (+0.02) & 6.48 (-0.00) \\
    Greedy \& Heuristics & 51.24\% (+9.46\%) & 51.80\% (+10.18\%) & 54.18\% (+9.11\%) & 0.18 (+0.05) & 5.20 (-0.80) \\
    \bottomrule
  \end{tabular}
  }
\end{table}

\section{Brute-force Analysis Result}
Table~\ref{tab:brute force} presents the experimental results of the brute-force analysis mentioned in the Section~\ref{4.4}.

\begin{table}[t!]
\scriptsize
  \caption{Experimental results of the brute-force analysis.}
  \label{tab:brute force}
  \renewcommand{\arraystretch}{1.2}
\setlength{\tabcolsep}{6pt}
  \centering
  \resizebox{\textwidth}{!}{
  \begin{tabular}{c|cccccc}
    \toprule
       \textbf{Model} & \textbf{AC rate}  & \textbf{AC \& TLE}     & \textbf{case pass rate}  & \textbf{case pass rate}  & \textbf{time aver} & \textbf{mem aver} \\
            &  & \textbf{rate}    &   & \textbf{(with weight)}  & \textbf{(s)} & \textbf{(KB)} \\
        \midrule
     Qwen3-8B &   27.50\%(+1.29\%) & 36.40\%(+10.19\%) & 28.95\% & 31.60\% & 0.91 & 5.19 \\
Qwen3-14B & 28.98\%(+0.00\%) & 39.07\%(+10.09\%) & 29.87\% & 32.80\% & 1.06 & 5.17 \\
Qwen3-Coder-30B-A3B-Instruct  & 33.50\%(+1.25\%) & 42.22\%(+9.97\%) & 34.79\% & 37.31\% & 1.17 & 4.46 \\
Qwen3-Coder-480B-A35B-Instruct & 44.51\%(-1.98\%) & 61.13\%(+14.64\%) & 44.99\% & 50.01\% & 0.99 & 8.15 \\
Qwen3-235B-A22B-Instruct-2507 & 55.37\%(-7.5\%) & 71.07\%(+8.20\%)& 57.96\% & 61.01\% & 0.93 & 6.76 \\
    
    \bottomrule
  \end{tabular}
  }
  \vspace{-0.55cm}
\end{table}

\section{A Complete Workflow Process Example}
\label{process}

The following files are in the same order as the actual workflow sequence.

\begin{promptbox}[samples/1.in]

\begin{lstlisting}
2 2
1 3
-3 2
2 1
4 2

\end{lstlisting}
\end{promptbox}

\begin{promptbox}[samples/1.ans]

\begin{lstlisting}
7

\end{lstlisting}
\end{promptbox}

\begin{promptbox}[samples/1-explain.txt]

\begin{lstlisting}
L chooses velocity vector $(-3,2)$, and R chooses velocity vector $(4,2)$. After one second, their positions are $(-3,2)$ and $(4,2)$ respectively, and the Manhattan distance is $7$. It is easy to see that there is no larger Manhattan distance.


\end{lstlisting}
\end{promptbox}

\begin{promptbox}[Agent1-raw.txt]

\begin{lstlisting}
Option1: [mathematical_analysis, manhattan_distance, optimization]
Rationale: The problem can be solved by analyzing the Manhattan distance formula and observing that it can be decomposed into x and y components. For maximum distance, we want to maximize |x1 - x2| + |y1 - y2|, which can be achieved by selecting optimal vectors for each player such that the differences in their x and y coordinates are maximized.

Option2: [brute_force, enumeration, simulation]
Rationale: A brute force approach would involve checking all possible combinations of speed vectors for both players and computing the Manhattan distance for each pair. However, this is inefficient given the constraint of up to 10^5 vectors per player, making it unsuitable for the time limit.

Option3: [coordinate_geometry, extremal_points, greedy_algorithm]
Rationale: By considering the x and y components separately, we can find the maximum and minimum values of each component for both players' speed vectors. The maximum Manhattan distance is then obtained by pairing the extreme values appropriately, which can be done efficiently in linear time.

\end{lstlisting}
\end{promptbox}

\begin{promptbox}[Agent1-tags.json]

\begin{lstlisting}
{
  "Option1": {
    "tags": [
      "mathematical_analysis",
      "manhattan_distance",
      "optimization"
    ],
    "rationale": "The problem can be solved by analyzing the Manhattan distance formula and observing that it can be decomposed into x and y components. For maximum distance, we want to maximize |x1 - x2| + |y1 - y2|, which can be achieved by selecting optimal vectors for each player such that the differences in their x and y coordinates are maximized."
  },
  "Option2": {
    "tags": [
      "brute_force",
      "enumeration",
      "simulation"
    ],
    "rationale": "A brute force approach would involve checking all possible combinations of speed vectors for both players and computing the Manhattan distance for each pair. However, this is inefficient given the constraint of up to 10^5 vectors per player, making it unsuitable for the time limit."
  },
  "Option3": {
    "tags": [
      "coordinate_geometry",
      "extremal_points",
      "greedy_algorithm"
    ],
    "rationale": "By considering the x and y components separately, we can find the maximum and minimum values of each component for both players' speed vectors. The maximum Manhattan distance is then obtained by pairing the extreme values appropriately, which can be done efficiently in linear time."
  }
}

\end{lstlisting}
\end{promptbox}

\begin{promptbox}[Agent2-domain.txt]

\begin{lstlisting}
Based on the reference materials and problem description, here are the key algorithmic concepts and techniques relevant to solving the problem:

### Key Algorithms and Techniques

- **Manhattan Distance Calculation**:
  - The problem requires maximizing Manhattan distance between two points in 2D space
  - Manhattan distance between points (x_1,y_1) and (x_2,y_2) is |x_1-x_2| + |y_1-y_2|

- **Optimization via Coordinate Transformation**:
  - The Manhattan distance can be transformed into Chebyshev distance using coordinate transformation
  - For points (x,y), transform to (x+y, x-y)
  - The Manhattan distance equals the Chebyshev distance in the transformed space

- **Extreme Value Computation**:
  - To maximize Manhattan distance |x_1-x_2| + |y_1-y_2|, we need to find maximum and minimum values of:
    - x_1 + y_1 and x_2 + y_2
    - x_1 - y_1 and x_2 - y_2
  - The answer is max((max_x + max_y) - (min_x + min_y), (max_x - min_y) - (min_x - max_y))

- **Efficient Processing**:
  - Since n,m <= 10^5, we need O(n+m) time complexity
  - Process all available vectors to find extreme values of x+y and x-y coordinates

### Relevant Concepts from Reference Materials

- **Manhattan Distance Properties** (from geometry/distance.md):
  - Non-negative, symmetric, satisfies triangle inequality
  - Can be computed using coordinate transformation to Chebyshev distance

- **Coordinate Transformation Technique** (from geometry/distance.md):
  - Transform (x,y) to (x+y, x-y) to convert Manhattan distance to Chebyshev distance
  - This allows finding maximum Manhattan distance by finding maximum Chebyshev distance in transformed space

The solution approach involves computing extreme values of transformed coordinates for both players' movement vectors and then calculating the maximum possible Manhattan distance.

\end{lstlisting}
\end{promptbox}

\begin{promptbox}[Agent3-reasoning-1.txt]

\begin{lstlisting}
Analyze: The problem asks to maximize the Manhattan distance between two points in a 2D plane after one second of movement, where each point moves according to a chosen speed vector. The Manhattan distance between two points $(x_1, y_1)$ and $(x_2, y_2)$ is defined as $|x_1 - x_2| + |y_1 - y_2|$. 

To solve this efficiently, we can decompose the Manhattan distance into components. For any two points $(x_1, y_1)$ and $(x_2, y_2)$, the Manhattan distance is:
$$
|x_1 - x_2| + |y_1 - y_2|
$$

We can rewrite this using the fact that:
$$
|x_1 - x_2| + |y_1 - y_2| = \max\left( (x_1 + y_1) - (x_2 + y_2), (x_1 - y_1) - (x_2 - y_2) \right)
$$

This is because:
- If we consider $x_1 + y_1$ and $x_2 + y_2$, the difference is maximized when one is max and the other is min.
- Similarly for $x_1 - y_1$ and $x_2 - y_2$.

So, we can compute:
- Maximum and minimum of $x + y$ for all vectors of L and R.
- Maximum and minimum of $x - y$ for all vectors of L and R.

Then, the maximum Manhattan distance is:
$$
\max\left( \max(x_1 + y_1) - \min(x_2 + y_2), \max(x_1 - y_1) - \min(x_2 - y_2) \right)
$$

This approach avoids brute-force checking all pairs of vectors, which would be too slow due to up to $10^5$ vectors per player.

Time complexity: O(n + m), since we only need to iterate through the vectors once to compute the extreme values.
Space complexity: O(1), as we only store the extreme values.

Solve step:
1. Read the number of vectors for L and R;.
2. Initialize variables to store the maximum and minimum values of (x+y) and (x-y) for both players;.
3. For each vector of L, calculate x+y and x-y, and update the corresponding max/min values;.
4. For each vector of R, calculate x+y and x-y, and update the corresponding max/min values;.
5. Compute the maximum Manhattan distance using the formula: max((max_x_plus_y - min_x_plus_y), (max_x_minus_y - min_x_minus_y));.
6. Output the result;.
7. End;.

```pseudocode
// Input reading
read n, m
// Initialize extreme values
max_x_plus_y_L = -inf, min_x_plus_y_L = inf
max_x_minus_y_L = -inf, min_x_minus_y_L = inf
max_x_plus_y_R = -inf, min_x_plus_y_R = inf
max_x_minus_y_R = -inf, min_x_minus_y_R = inf

// Process L's vectors
for i = 1 to n:
    read a_i, b_i
    x_plus_y = a_i + b_i
    x_minus_y = a_i - b_i
    max_x_plus_y_L = max(max_x_plus_y_L, x_plus_y)
    min_x_plus_y_L = min(min_x_plus_y_L, x_plus_y)
    max_x_minus_y_L = max(max_x_minus_y_L, x_minus_y)
    min_x_minus_y_L = min(min_x_minus_y_L, x_minus_y)

// Process R's vectors
for i = 1 to m:
    read c_i, d_i
    x_plus_y = c_i + d_i
    x_minus_y = c_i - d_i
    max_x_plus_y_R = max(max_x_plus_y_R, x_plus_y)
    min_x_plus_y_R = min(min_x_plus_y_R, x_plus_y)
    max_x_minus_y_R = max(max_x_minus_y_R, x_minus_y)
    min_x_minus_y_R = min(min_x_minus_y_R, x_minus_y)

// Compute maximum Manhattan distance
dist1 = max_x_plus_y_L - min_x_plus_y_R
dist2 = max_x_minus_y_L - min_x_minus_y_R
result = max(dist1, dist2)
print(result)
```

\end{lstlisting}
\end{promptbox}

\begin{promptbox}[Agent4-implenment-raw-1.txt]

\begin{lstlisting}
```cpp
#include <bits/stdc++.h>
using namespace std;

int main() {
    ios::sync_with_stdio(false);
    cin.tie(0);
    
    long long n, m;
    cin >> n >> m;
    
    long long max_x_plus_y_L = LLONG_MIN, min_x_plus_y_L = LLONG_MAX;
    long long max_x_minus_y_L = LLONG_MIN, min_x_minus_y_L = LLONG_MAX;
    long long max_x_plus_y_R = LLONG_MIN, min_x_plus_y_R = LLONG_MAX;
    long long max_x_minus_y_R = LLONG_MIN, min_x_minus_y_R = LLONG_MAX;
    
    // Process L's vectors
    for (long long i = 0; i < n; i++) {
        long long a, b;
        cin >> a >> b;
        long long x_plus_y = a + b;
        long long x_minus_y = a - b;
        max_x_plus_y_L = max(max_x_plus_y_L, x_plus_y);
        min_x_plus_y_L = min(min_x_plus_y_L, x_plus_y);
        max_x_minus_y_L = max(max_x_minus_y_L, x_minus_y);
        min_x_minus_y_L = min(min_x_minus_y_L, x_minus_y);
    }
    
    // Process R's vectors
    for (long long i = 0; i < m; i++) {
        long long c, d;
        cin >> c >> d;
        long long x_plus_y = c + d;
        long long x_minus_y = c - d;
        max_x_plus_y_R = max(max_x_plus_y_R, x_plus_y);
        min_x_plus_y_R = min(min_x_plus_y_R, x_plus_y);
        max_x_minus_y_R = max(max_x_minus_y_R, x_minus_y);
        min_x_minus_y_R = min(min_x_minus_y_R, x_minus_y);
    }
    
    // Compute maximum Manhattan distance
    long long dist1 = max_x_plus_y_L - min_x_plus_y_R;
    long long dist2 = max_x_minus_y_L - min_x_minus_y_R;
    long long result = max(dist1, dist2);
    
    cout << result << "\n";
    
    return 0;
}
```

\end{lstlisting}
\end{promptbox}

\begin{promptbox}[Agent4-solution.cpp]

\begin{lstlisting}
#include <bits/stdc++.h>
using namespace std;

int main() {
    ios::sync_with_stdio(false);
    cin.tie(0);
    
    long long n, m;
    cin >> n >> m;
    
    long long max_x_plus_y_L = LLONG_MIN, min_x_plus_y_L = LLONG_MAX;
    long long max_x_minus_y_L = LLONG_MIN, min_x_minus_y_L = LLONG_MAX;
    long long max_x_plus_y_R = LLONG_MIN, min_x_plus_y_R = LLONG_MAX;
    long long max_x_minus_y_R = LLONG_MIN, min_x_minus_y_R = LLONG_MAX;
    
    // Process L's vectors
    for (long long i = 0; i < n; i++) {
        long long a, b;
        cin >> a >> b;
        long long x_plus_y = a + b;
        long long x_minus_y = a - b;
        max_x_plus_y_L = max(max_x_plus_y_L, x_plus_y);
        min_x_plus_y_L = min(min_x_plus_y_L, x_plus_y);
        max_x_minus_y_L = max(max_x_minus_y_L, x_minus_y);
        min_x_minus_y_L = min(min_x_minus_y_L, x_minus_y);
    }
    
    // Process R's vectors
    for (long long i = 0; i < m; i++) {
        long long c, d;
        cin >> c >> d;
        long long x_plus_y = c + d;
        long long x_minus_y = c - d;
        max_x_plus_y_R = max(max_x_plus_y_R, x_plus_y);
        min_x_plus_y_R = min(min_x_plus_y_R, x_plus_y);
        max_x_minus_y_R = max(max_x_minus_y_R, x_minus_y);
        min_x_minus_y_R = min(min_x_minus_y_R, x_minus_y);
    }
    
    // Compute maximum Manhattan distance
    long long dist1 = max_x_plus_y_L - min_x_plus_y_R;
    long long dist2 = max_x_minus_y_L - min_x_minus_y_R;
    long long result = max(dist1, dist2);
    
    cout << result << "\n";
    
    return 0;
}

\end{lstlisting}
\end{promptbox}

\begin{promptbox}[judge-result-1.json]

\begin{lstlisting}
{
  "status": "WA",
  "passed": 0,
  "total": 1,
  "info": "[Case 1] Expected \"7\", Find \"1\""
}

\end{lstlisting}
\end{promptbox}

\begin{promptbox}[Agent5-analysis-1.txt]

\begin{lstlisting}
analyze:
Let's carefully analyze the problem and the provided code:

1. **Problem Understanding**: 
   - Two players (L and R) start at (0,0).
   - Each player selects a speed vector and moves for one second.
   - We want to maximize the Manhattan distance between their final positions.
   - Manhattan distance between points (x1,y1) and (x2,y2) is |x1-x2| + |y1-y2|.

2. **Reference reasoning (Agent3)**:
   - The approach described is to decompose Manhattan distance into components (x+y and x-y).
   - For maximum Manhattan distance, we can calculate:
     - max((x1+y1) - (x2+y2)) and max((x1-y1) - (x2-y2))
   - This is done by tracking max and min of (x+y) and (x-y) for both players.

3. **Simulation of Sample Input**:
   - L's vectors: (1,3) and (-3,2)
     - x+y values: 4, -1
     - x-y values: -2, -5
   - R's vectors: (2,1) and (4,2)
     - x+y values: 3, 6
     - x-y values: 1, 2

   Using the idea from reference:
   - For (x+y): max of L = 4, min of R = 3 -> diff = 1
   - For (x-y): max of L = -2, min of R = 1 -> diff = -3

   But wait, this is wrong!

   The correct interpretation should be:
   - Max Manhattan distance = max( (max_x_plus_y_L - min_x_plus_y_R), (max_x_minus_y_L - min_x_minus_y_R) )
   - Actually, in the reference the idea is to maximize |x1-x2|+|y1-y2| by maximizing the differences appropriately.

   Let's re-read: 
   The correct approach is:
   - We want max(|x1-x2|+|y1-y2|) where x1,y1 and x2,y2 are final positions.
   - Final positions are (a,b) and (c,d) respectively.
   - So distance is |a-c|+|b-d|.

   The key insight is that:
   |a-c|+|b-d| = max(|(a+b)-(c+d)|, |(a-b)-(c-d)|)

   But we want to maximize this expression. The correct way is:
   - Maximize (a+b)-(c+d) -> maximize a+b (for L) and minimize c+d (for R)
   - Maximize (a-b)-(c-d) -> maximize a-b (for L) and minimize c-d (for R)

   So we should compute:
   - max_L = max of (x+y) for L, min_R = min of (x+y) for R
   - max_L2 = max of (x-y) for L, min_R2 = min of (x-y) for R
   - Result = max(max_L - min_R, max_L2 - min_R2)

   Let's check with sample:
   - L: (1,3) => 4, -2  and (-3,2) => -1, -5
   - R: (2,1) => 3, 1  and (4,2) => 6, 2

   So: 
   - max_x_plus_y_L = 4, min_x_plus_y_R = 3 -> diff = 1  
   - max_x_minus_y_L = -2, min_x_minus_y_R = -5 -> diff = 3

   Result = max(1,3) = 3.

   But the expected answer is 7.

   Trying the sample manually:
   - If L chooses (-3,2), R chooses (4,2):
     - L final: (-3, 2)
     - R final: (4, 2)  
     - Distance = |(-3) - 4| + |2 - 2| = 7 + 0 = 7

   Let me re-read the reference reasoning more carefully.

   The reference says to compute:
   |x1 - x2| + |y1 - y2| = max((x1 + y1) - (x2 + y2), (x1 - y1) - (x2 - y2))

   This is not correct in its current form. We must be careful about what the max and min are for L and R. The correct approach is actually:
   - Let L's final position be (a,b) and R's be (c,d)
   - We want to maximize |a-c| + |b-d|

   The key insight (from the book or standard approach):
   Let us denote:
   - S1 = x1 + y1, S2 = x2 + y2  
   - D1 = x1 - y1, D2 = x2 - y2
   - |x1 - x2| + |y1 - y2| = max(|S1 - S2|, |D1 - D2|)

   To maximize this:
   - Find max(S1) among L and min(S2) among R -> compute (max(S1) - min(S2))
   - Find max(D1) among L and min(D2) among R -> compute (max(D1) - min(D2))
   - Return max of both values

   So that matches what the code is doing. But let's trace through the example again.

   - L: (1,3) -> 4, -2; (-3,2) -> -1, -5
   - R: (2,1) -> 3, 1; (4,2) -> 6, 2

   For (x+y):
   - max_x_plus_y_L = 4
   - min_x_plus_y_R = 3  
   - diff1 = 4 - 3 = 1

   For (x-y):
   - max_x_minus_y_L = -2
   - min_x_minus_y_R = -5
   - diff2 = -2 - (-5) = 3

   Max = 3. 

   But the answer is 7. 
   Let's check if this is a misunderstanding of the problem or the code.

   Wait, I'm re-tracing the output from the sample explanation:
   L chooses (-3,2) and R chooses (4,2)
   Final positions are: (-3,2) and (4,2)
   Manhattan = |(-3) - 4| + |2 - 2| = 7 + 0 = 7

   This would correspond to:
   - For x+y: (-3+2) = -1, (4+2) = 6 -> diff = 6 - (-1) = 7? No, that's not right.  
   - Actually it's max x+y - min x+y.

   I see the error now:
   The code computes:
   - max_x_plus_y_L = 4, min_x_plus_y_R = 3 -> diff = 1
   - max_x_minus_y_L = -2, min_x_minus_y_R = -5 -> diff = 3
   - max(1,3) = 3

   But note:
   - For L we have x+y = 4, -1
   - For R we have x+y = 3, 6
   - (max of L) - (min of R) = 4 - 3 = 1
   - But if we take (max of L) = 4 and (min of R) = 3, we are computing the difference of positions, not the Manhattan distance.

   The key is, we want:
   max( |x1 - x2| + |y1 - y2| ) = max( max(|(x1+y1) - (x2+y2)|), max(|(x1-y1) - (x2-y2)|) )

   So for the first term:
   max( (x1+y1) - (x2+y2) ) = max(x1+y1) - min(x2+y2)

   For second term:
   max( (x1-y1) - (x2-y2) ) = max(x1-y1) - min(x2-y2)

   But in the sample, we should find a case where:
   - |(-3) - 4| + |2 - 2| = 7

   This means:
   - x1 = -3, x2 = 4 -> x1 - x2 = -7 -> |x1 - x2| = 7  
   - y1 = 2, y2 = 2 -> y1 - y2 = 0 -> |y1 - y2| = 0
   - Total = 7 + 0 = 7

   So one of these terms should be 7. Let's see:
   - x1 + y1 = -3 + 2 = -1; x2 + y2 = 4 + 2 = 6 -> difference = 6 - (-1) = 7  
   - x1 - y1 = -3 - 2 = -5; x2 - y2 = 4 - 2 = 2 -> difference = 2 - (-5) = 7

   Indeed, both terms can be 7. And it checks out.
   
   But the code does:
   - max_x_plus_y_L = 4 (this is not the max for L, but the max of x+y over L's vectors)
   - max_x_plus_y_R = 6 (max of x+y over R's vectors)
   - So diff = 6 - (-1) = 7... 

   Hmm, I think I misunderstood something before.

   There's a mistake in my trace.

   L vectors: (1,3) -> 4,-2; (-3,2) -> -1,-5  
   R vectors: (2,1) -> 3,1; (4,2) -> 6,2

   max_x_plus_y_L = max(4,-1) = 4  
   min_x_plus_y_R = min(3,6) = 3
   diff1 = 4 - 3 = 1

   max_x_minus_y_L = max(-2,-5) = -2
   min_x_minus_y_R = min(1,2) = 1
   diff2 = -2 - 1 = -3

   Result = max(1, -3) = 1.

   But we want 7. It seems like we're computing:
   max_x_plus_y_L = 4, min_x_plus_y_R = 3 -> diff = 1  
   max_x_minus_y_L = -2, min_x_minus_y_R = 1 -> diff = -3

   Actually, this is wrong.

   The correct calculation should be:
   Let's compute the max possible Manhattan distance correctly by checking the extreme points.

   We can rewrite Manhattan distance using signs:
   |x1 - x2| + |y1 - y2| = max(|x1 - x2| + |y1 - y2|)

   For two points P1(x1,y1) and P2(x2,y2), the maximum Manhattan distance is:
   - max(|x1 - x2| + |y1 - y2|) = max(|(x1 + y1) - (x2 + y2)|, |(x1 - y1) - (x2 - y2)|)

   But we're asking for the max over all pairs of vectors (L chooses one, R chooses one). 
   We want to maximize |x1 - x2| + |y1 - y2|.

   So we want:
   max(max(|(x1 + y1) - (x2 + y2)|, |(x1 - y1) - (x2 - y2)|))

   But we compute:
   max(max_x_plus_y_L - min_x_plus_y_R, max_x_minus_y_L - min_x_minus_y_R)

   This is wrong. Because (x1+y1) - (x2+y2) can be maximized by taking max of (x1+y1) and min of (x2+y2), but the expression |(x1+y1) - (x2+y2)| is different from (x1+y1) - (x2+y2).

   Yeah, the reference reasoning is wrong.

   The correct way is to check all values of |(x1 + y1) - (x2 + y2)| and |(x1 - y1) - (x2 - y2)| over all pairs.
   But we can avoid that by noting that:
   max(|(x1 + y1) - (x2 + y2)|) = max ( max(x1 + y1) - min(x2 + y2), min(x1 + y1) - max(x2 + y2) )
   But max of (x1 + y1) - min of (x2 + y2) = max(x1 + y1) - min(x2 + y2) is a valid candidate.

   Let's completely re-trace:

   For L (x+y values): 4, -1 -> max = 4, min = -1
   For R (x+y values): 3, 6 -> max = 6, min = 3

   Max Manhattan distance = 
   max( max(x1+y1) - min(x2+y2), min(x1+y1) - max(x2+y2) ) 

   = max(4 - 3, -1 - 6) = max(1, -7) = 1

   But this is still not 7!

   Wait, I think I just misunderstood the reference reasoning. Let's make it clear.

   The reference reasoning says:
   |x1 - x2| + |y1 - y2| = max((x1 + y1) - (x2 + y2), (x1 - y1) - (x2 - y2)) 

   But this is NOT correct. The correct identity is:
   max(|x1 - x2| + |y1 - y2|) = max( |(x1 + y1) - (x2 + y2)|, |(x1 - y1) - (x2 - y2)| )

   And to compute this efficiently, we want:
   - max over all (a,b) and (c,d) of |a-c| + |b-d|

   The key is to realize we can only do this by brute force or by considering:
   max(|(x1+y1) - (x2+y2)|, |(x1-y1) - (x2-y2)|) 

   Which is what the reference did but in a convoluted way.

   A correct and simpler way:
   max(|x1 - x2| + |y1 - y2|) over pairs of vectors.
   We can calculate:
   - max( max(x1+y1) - min(x2+y2), max(x2+y2) - min(x1+y1) ) 
   - max( max(x1-y1) - min(x2-y2), max(x2-y2) - min(x1-y1) )

   But in the code we have:
   - max_x_plus_y_L - min_x_plus_y_R
   - max_x_minus_y_L - min_x_minus_y_R

   This is still wrong! The issue is that the reference reasoning is flawed in this approach.

   Let me re-read again:
   If we want max(|x1-x2| + |y1-y2|) = max(max(|(x1+y1) - (x2+y2)|, |(x1-y1) - (x2-y2)|))

   Then we can compute:
   max_x_plus_y_L = max of (x+y) for L  
   min_x_plus_y_R = min of (x+y) for R
   max_x_minus_y_L = max of (x-y) for L
   min_x_minus_y_R = min of (x-y) for R

   But we'd also need:
   max_x_plus_y_R = max of (x+y) for R
   min_x_plus_y_L = min of (x+y) for L
   max_x_minus_y_R = max of (x-y) for R
   min_x_minus_y_L = min of (x-y) for L

   And if we compute:
   diff1 = max_x_plus_y_L - min_x_plus_y_R
   diff2 = max_x_minus_y_L - min_x_minus_y_R
   diff3 = max_x_plus_y_R - min_x_plus_y_L
   diff4 = max_x_minus_y_R - min_x_minus_y_L

   Then answer = max(diff1, diff2, diff3, diff4)

   But the code only uses diff1 and diff2.

   So the bug is that the reference reasoning is flawed. The code needs to compute the max of all terms.

   In fact, the correct approach is:
   We want to maximize |x1 - x2| + |y1 - y2| over all pairs of vectors.
   - This is equivalent to: max(|(x1+y1) - (x2+y2)|, |(x1-y1) - (x2-y2)|)
   - That is: max( max(|(x1+y1) - (x2+y2)|), max(|(x1-y1) - (x2-y2)|) )
   - The first term max(|(x1+y1) - (x2+y2)|) = max( max(x1+y1) - min(x2+y2), min(x1+y1) - max(x2+y2) )
   - Same for second term.

   Since the second term is just the same as first term, with (x-y) instead of (x+y), 
   The final result is:
   max( max(x1+y1) - min(x2+y2), min(x1+y1) - max(x2+y2), max(x1-y1) - min(x2-y2), min(x1-y1) - max(x2-y2) )

   Simplifying, we can do:
   max( max_x_plus_y_L - min_x_plus_y_R, max_x_plus_y_R - min_x_plus_y_L, max_x_minus_y_L - min_x_minus_y_R, max_x_minus_y_R - min_x_minus_y_L )

   Currently the code only computes first two terms.

error:
The code only considers two terms instead of four. It computes max_x_plus_y_L - min_x_plus_y_R and max_x_minus_y_L - min_x_minus_y_R, but the correct approach must consider all possibilities to find the maximum Manhattan distance, including min_x_plus_y_L - max_x_plus_y_R and min_x_minus_y_L - max_x_minus_y_R.

fix:
The code needs to compute all four possible differences and take the maximum among them. Here is the corrected code:

```cpp
#include <bits/stdc++.h>
using namespace std;

int main() {
    ios::sync_with_stdio(false);
    cin.tie(0);
    
    long long n, m;
    cin >> n >> m;
    
    long long max_x_plus_y_L = LLONG_MIN, min_x_plus_y_L = LLONG_MAX;
    long long max_x_minus_y_L = LLONG_MIN, min_x_minus_y_L = LLONG_MAX;
    long long max_x_plus_y_R = LLONG_MIN, min_x_plus_y_R = LLONG_MAX;
    long long max_x_minus_y_R = LLONG_MIN, min_x_minus_y_R = LLONG_MAX;
    
    // Process L's vectors
    for (long long i = 0; i < n; i++) {
        long long a, b;
        cin >> a >> b;
        long long x_plus_y = a + b;
        long long x_minus_y = a - b;
        max_x_plus_y_L = max(max_x_plus_y_L, x_plus_y);
        min_x_plus_y_L = min(min_x_plus_y_L, x_plus_y);
        max_x_minus_y_L = max(max_x_minus_y_L, x_minus_y);
        min_x_minus_y_L = min(min_x_minus_y_L, x_minus_y);
    }
    
    // Process R's vectors
    for (long long i = 0; i < m; i++) {
        long long c, d;
        cin >> c >> d;
        long long x_plus_y = c + d;
        long long x_minus_y = c - d;
        max_x_plus_y_R = max(max_x_plus_y_R, x_plus_y);
        min_x_plus_y_R = min(min_x_plus_y_R, x_plus_y);
        max_x_minus_y_R = max(max_x_minus_y_R, x_minus_y);
        min_x_minus_y_R = min(min_x_minus_y_R, x_minus_y);
    }
    
    // Compute all possible differences and find the maximum
    long long diff1 = max_x_plus_y_L - min_x_plus_y_R;
    long long diff2 = max_x_minus_y_L - min_x_minus_y_R;
    long long diff3 = max_x_plus_y_R - min_x_plus_y_L;
    long long diff4 = max_x_minus_y_R - min_x_minus_y_L;
    
    long long result = max({diff1, diff2, diff3, diff4});
    
    cout << result << "\n";
    
    return 0;
}
```

signal: FIX

\end{lstlisting}
\end{promptbox}

\begin{promptbox}[Agent4-revise-raw-1.txt]

\begin{lstlisting}
```cpp
#include <bits/stdc++.h>
using namespace std;

int main() {
    ios::sync_with_stdio(false);
    cin.tie(0);

    long long n, m;
    cin >> n >> m;

    long long max_x_plus_y_L = LLONG_MIN, min_x_plus_y_L = LLONG_MAX;
    long long max_x_minus_y_L = LLONG_MIN, min_x_minus_y_L = LLONG_MAX;
    long long max_x_plus_y_R = LLONG_MIN, min_x_plus_y_R = LLONG_MAX;
    long long max_x_minus_y_R = LLONG_MIN, min_x_minus_y_R = LLONG_MAX;

    // Process L's vectors
    for (long long i = 0; i < n; i++) {
        long long a, b;
        cin >> a >> b;
        long long x_plus_y = a + b;
        long long x_minus_y = a - b;
        max_x_plus_y_L = max(max_x_plus_y_L, x_plus_y);
        min_x_plus_y_L = min(min_x_plus_y_L, x_plus_y);
        max_x_minus_y_L = max(max_x_minus_y_L, x_minus_y);
        min_x_minus_y_L = min(min_x_minus_y_L, x_minus_y);
    }

    // Process R's vectors
    for (long long i = 0; i < m; i++) {
        long long c, d;
        cin >> c >> d;
        long long x_plus_y = c + d;
        long long x_minus_y = c - d;
        max_x_plus_y_R = max(max_x_plus_y_R, x_plus_y);
        min_x_plus_y_R = min(min_x_plus_y_R, x_plus_y);
        max_x_minus_y_R = max(max_x_minus_y_R, x_minus_y);
        min_x_minus_y_R = min(min_x_minus_y_R, x_minus_y);
    }

    // Compute all possible differences and find the maximum
    long long diff1 = max_x_plus_y_L - min_x_plus_y_R;
    long long diff2 = max_x_minus_y_L - min_x_minus_y_R;
    long long diff3 = max_x_plus_y_R - min_x_plus_y_L;
    long long diff4 = max_x_minus_y_R - min_x_minus_y_L;

    long long result = max({diff1, diff2, diff3, diff4});

    cout << result << "\n";

    return 0;
}
```

\end{lstlisting}
\end{promptbox}

\begin{promptbox}[Agent4-revise-1.cpp]

\begin{lstlisting}
#include <bits/stdc++.h>
using namespace std;

int main() {
    ios::sync_with_stdio(false);
    cin.tie(0);

    long long n, m;
    cin >> n >> m;

    long long max_x_plus_y_L = LLONG_MIN, min_x_plus_y_L = LLONG_MAX;
    long long max_x_minus_y_L = LLONG_MIN, min_x_minus_y_L = LLONG_MAX;
    long long max_x_plus_y_R = LLONG_MIN, min_x_plus_y_R = LLONG_MAX;
    long long max_x_minus_y_R = LLONG_MIN, min_x_minus_y_R = LLONG_MAX;

    // Process L's vectors
    for (long long i = 0; i < n; i++) {
        long long a, b;
        cin >> a >> b;
        long long x_plus_y = a + b;
        long long x_minus_y = a - b;
        max_x_plus_y_L = max(max_x_plus_y_L, x_plus_y);
        min_x_plus_y_L = min(min_x_plus_y_L, x_plus_y);
        max_x_minus_y_L = max(max_x_minus_y_L, x_minus_y);
        min_x_minus_y_L = min(min_x_minus_y_L, x_minus_y);
    }

    // Process R's vectors
    for (long long i = 0; i < m; i++) {
        long long c, d;
        cin >> c >> d;
        long long x_plus_y = c + d;
        long long x_minus_y = c - d;
        max_x_plus_y_R = max(max_x_plus_y_R, x_plus_y);
        min_x_plus_y_R = min(min_x_plus_y_R, x_plus_y);
        max_x_minus_y_R = max(max_x_minus_y_R, x_minus_y);
        min_x_minus_y_R = min(min_x_minus_y_R, x_minus_y);
    }

    // Compute all possible differences and find the maximum
    long long diff1 = max_x_plus_y_L - min_x_plus_y_R;
    long long diff2 = max_x_minus_y_L - min_x_minus_y_R;
    long long diff3 = max_x_plus_y_R - min_x_plus_y_L;
    long long diff4 = max_x_minus_y_R - min_x_minus_y_L;

    long long result = max({diff1, diff2, diff3, diff4});

    cout << result << "\n";

    return 0;
}

\end{lstlisting}
\end{promptbox}

\begin{promptbox}[judge-result-2.json]

\begin{lstlisting}
{
  "status": "AC",
  "passed": 1,
  "total": 1,
  "info": {
    "max_time_sec": 0.001175,
    "max_memory_mb": 3.449
  }
}

\end{lstlisting}
\end{promptbox}

\begin{promptbox}[Agent5-analysis-2.txt]

\begin{lstlisting}
analyze:
The problem asks to find the maximum Manhattan distance between two points after one second of movement, where each point moves according to a chosen speed vector. The Manhattan distance between two points $(x_1, y_1)$ and $(x_2, y_2)$ is $|x_1 - x_2| + |y_1 - y_2|$.

The reference reasoning is correct: we can decompose the Manhattan distance into two components:
1. $x_1 + y_1$ and $x_2 + y_2$
2. $x_1 - y_1$ and $x_2 - y_2$

For the maximum Manhattan distance, we want to maximize:
$$
|x_1 - x_2| + |y_1 - y_2|
$$

We can rewrite this using the key insight:
$$
|x_1 - x_2| + |y_1 - y_2| = \max\left( (x_1 + y_1) - (x_2 + y_2), (x_1 - y_1) - (x_2 - y_2) \right)
$$

So we need to compute:
- $\max(x_1 + y_1)$ for all L's vectors and $\min(x_2 + y_2)$ for all R's vectors
- $\max(x_1 - y_1)$ for all L's vectors and $\min(x_2 - y_2)$ for all R's vectors

After careful inspection of the C++ code, it seems to follow the correct logic. However, there's a subtle bug in how it computes the minimum values for the R player.

In the code:
```cpp
max_x_plus_y_R = max(max_x_plus_y_R, x_plus_y);
min_x_plus_y_R = min(min_x_plus_y_R, x_plus_y);
```

These lines correctly update the maximum and minimum of $x + y$ for player R. Let's trace the sample:
- L's vectors: (1,3) -> x+y=4, x-y=-2; (-3,2) -> x+y=-1, x-y=-5
- R's vectors: (2,1) -> x+y=3, x-y=1; (4,2) -> x+y=6, x-y=2

So:
- For L: max_x_plus_y_L = 4, min_x_plus_y_L = -1
- For R: max_x_plus_y_R = 6, min_x_plus_y_R = 3

Then:
- diff1 = max_x_plus_y_L - min_x_plus_y_R = 4 - 3 = 1
- diff2 = max_x_minus_y_L - min_x_minus_y_R = -2 - (-5) = 3
- diff3 = max_x_plus_y_R - min_x_plus_y_L = 6 - (-1) = 7
- diff4 = max_x_minus_y_R - min_x_minus_y_L = 2 - (-5) = 7

So result = max(1, 3, 7, 7) = 7, which matches the expected output.

But wait, there's a critical mistake in the code logic. The approach is to compute the maximum Manhattan distance by looking at all possible differences. However, the logic in computing the differences is flawed. The correct way is to consider:
- max((max_x_plus_y_L - min_x_plus_y_R), (max_x_minus_y_L - min_x_minus_y_R))
- But also consider the reverse case with flipped roles.

Actually, looking more carefully, I realize the code computes 4 differences and takes the maximum, which is correct. The computation is:
1. `diff1 = max_x_plus_y_L - min_x_plus_y_R`
2. `diff2 = max_x_minus_y_L - min_x_minus_y_R`
3. `diff3 = max_x_plus_y_R - min_x_plus_y_L`
4. `diff4 = max_x_minus_y_R - min_x_minus_y_L`

The logic above is correct, and the trace shows the correct answer. But looking at the code again, I notice the first difference is actually calculating the difference in x+y values of L and R, which is correct. The reference logic is also correct.

But let's simulate the sample more carefully:
- L vectors: (1,3) and (-3,2)
- R vectors: (2,1) and (4,2)

L's (x+y) = [4, -1]
L's (x-y) = [-2, -5]
R's (x+y) = [3, 6]
R's (x-y) = [1, 2]

So:
- max_x_plus_y_L = 4, min_x_plus_y_L = -1
- max_x_minus_y_L = -2, min_x_minus_y_L = -5
- max_x_plus_y_R = 6, min_x_plus_y_R = 3
- max_x_minus_y_R = 2, min_x_minus_y_R = 1

Now:
- diff1 = max_x_plus_y_L - min_x_plus_y_R = 4 - 3 = 1
- diff2 = max_x_minus_y_L - min_x_minus_y_R = -2 - 1 = -3
- diff3 = max_x_plus_y_R - min_x_plus_y_L = 6 - (-1) = 7
- diff4 = max_x_minus_y_R - min_x_minus_y_L = 2 - (-5) = 7

The maximum of these is 7, which matches the expected output.

However, there's a conceptual error in applying the formula. The original analysis says that Manhattan distance is $\max((x_1 + y_1) - (x_2 + y_2), (x_1 - y_1) - (x_2 - y_2))$. 

Let me reconsider the correct derivation:
- If we want to maximize $|x_1 - x_2| + |y_1 - y_2|$, we can analyze all sign combinations.
- But it's easier to think of this as maximizing $\max(|x_1 - x_2| + |y_1 - y_2|)$ over all combinations.
- One way is to observe that it's equal to $ \max (\max(x_1 + y_1) - \min(x_2 + y_2), \max(x_1 - y_1) - \min(x_2 - y_2)) $

But that's not the right approach either. The correct one:
We want $\max(|x_1 - x_2| + |y_1 - y_2|)$. We can also write this as:
$$
\max_{x_1,y_1,x_2,y_2} (|x_1 - x_2| + |y_1 - y_2|) = \max (\max(x_1 + y_1) - \min(x_2 + y_2), \max(x_1 - y_1) - \min(x_2 - y_2))
$$

But there's a flaw in the reference reasoning. It's not simply the difference of max and min. The correct approach is to think of:
- Let $S_L = \{x + y : (x, y) \in L\}$, $S_R = \{x + y : (x, y) \in R\}$ 
- Let $D_L = \{x - y : (x, y) \in L\}$, $D_R = \{x - y : (x, y) \in R\}$

Then the Manhattan distance is:
$$
\max(\max S_L - \min S_R, \max D_L - \min D_R)
$$

But that's wrong too. The correct derivation:
We want to maximize $|x_1 - x_2| + |y_1 - y_2|$ where $x_1, y_1 \in L$, $x_2, y_2 \in R$.

This can be rephrased as $\max_{(x_1,y_1) \in L, (x_2,y_2) \in R} (|x_1 - x_2| + |y_1 - y_2|)$.

We can write:
$$
|x_1 - x_2| + |y_1 - y_2| = \max\{|x_1 - x_2| + |y_1 - y_2|\} = \max\{ |x_1 + y_1 - x_2 - y_2|, |x_1 - y_1 - x_2 + y_2| \}
$$

Wait, let's think step by step:
Let's consider two cases:
1. $x_1 \ge x_2, y_1 \ge y_2$: distance = $x_1 - x_2 + y_1 - y_2 = (x_1 + y_1) - (x_2 + y_2)$
2. $x_1 \ge x_2, y_1 < y_2$: distance = $x_1 - x_2 + y_2 - y_1 = (x_1 - y_1) - (x_2 - y_2)$
3. $x_1 < x_2, y_1 \ge y_2$: distance = $x_2 - x_1 + y_1 - y_2 = (y_1 - x_1) - (y_2 - x_2)$
4. $x_1 < x_2, y_1 < y_2$: distance = $x_2 - x_1 + y_2 - y_1 = -(x_1 - y_1) - (x_2 - y_2)$

This is getting complicated. The reference approach is actually:
We can first rewrite:
$$
|x_1 - x_2| + |y_1 - y_2| = \max\{ (x_1 + y_1) - (x_2 + y_2), (x_1 - y_1) - (x_2 - y_2) \}
$$

No, that's still not right. Actually, the correct approach is:
$$
\max_{x_1,y_1,x_2,y_2} (|x_1 - x_2| + |y_1 - y_2|) = \max\left( \max_{x_1,y_1} (x_1 + y_1) - \min_{x_2,y_2} (x_2 + y_2), \max_{x_1,y_1} (x_1 - y_1) - \min_{x_2,y_2} (x_2 - y_2) \right)
$$

This is because:
- For any fixed vectors $a_1, a_2$, we know that $|a_1 - a_2| = \max(a_1 - a_2, a_2 - a_1)$
- So $|x_1 - x_2| + |y_1 - y_2| = \max(|x_1 - x_2|, |y_1 - y_2|) + \min(|x_1 - x_2|, |y_1 - y_2|)$
- But that's not helpful.

Actually, the way to think about it:
We want to maximize $|x_1 - x_2| + |y_1 - y_2|$.
We can try all 4 sign combinations:
1. $x_1 - x_2 + y_1 - y_2 = (x_1 + y_1) - (x_2 + y_2)$
2. $x_1 - x_2 - y_1 + y_2 = (x_1 - y_1) - (x_2 - y_2)$
3. $-x_1 + x_2 + y_1 - y_2 = -(x_1 - y_1) + (x_2 - y_2)$
4. $-x_1 + x_2 - y_1 + y_2 = -(x_1 + y_1) + (x_2 + y_2)$

So the maximum Manhattan distance is:
$$
\max\left( \max(x_1 + y_1) - \min(x_2 + y_2), \max(x_1 - y_1) - \min(x_2 - y_2) \right)
$$

So the code is correct, as it accounts for both cases.

But I see there's a clear mistake in the implementation. Looking at the code:

```cpp
max_x_plus_y_R = max(max_x_plus_y_R, x_plus_y);
min_x_plus_y_R = min(min_x_plus_y_R, x_plus_y);
```

The line `min_x_plus_y_R = min(min_x_plus_y_R, x_plus_y);` is computing the minimum of $x + y$ values for R. That's correct. And in the difference calculation, we're taking `max_x_plus_y_L - min_x_plus_y_R`, which is correct.

Let's trace once more:
L vectors: (1,3) and (-3,2) 
R vectors: (2,1) and (4,2)

L:
x+y = [4, -1] -> max=4, min=-1
x-y = [-2, -5] -> max=-2, min=-5

R:
x+y = [3, 6] -> max=6, min=3
x-y = [1, 2] -> max=2, min=1

Then:
diff1 = 4 - 3 = 1
diff2 = -2 - 1 = -3
diff3 = 6 - (-1) = 7
diff4 = 2 - (-5) = 7

result = max(1, -3, 7, 7) = 7 (right)

But looking at the code, the logic is actually correct. The issue is that the code calculates all four differences correctly.

However, there might be a subtle bug. Let me carefully read the problem statement once again to make sure I understand what the correct answer should be.

From the sample explanation, it says L chooses (-3,2) and R chooses (4,2). After one second:
- L is at (-3, 2)
- R is at (4, 2)
- Distance = |(-3) - 4| + |2 - 2| = 7 + 0 = 7

Let's recheck the algorithm:
- L selects (-3,2) -> x+y=-1, x-y=-5
- R selects (4,2) -> x+y=6, x-y=2
- Distance = |(-3)-4| + |2-2| = 7

But in the algorithm, the Manhattan distance is:
$$
|x_1 - x_2| + |y_1 - y_2| = \max\{ |x_1 + y_1 - x_2 - y_2|, |x_1 - y_1 - x_2 + y_2| \}
$$

This is not directly correct. The correct approach is to compute the maximum of all four possible decompositions:
1. When L moves to (x1, y1) and R to (x2, y2):
   $|x_1 - x_2| + |y_1 - y_2|$

The key insight is that:
$$
|x_1 - x_2| + |y_1 - y_2| = |(x_1 + y_1) - (x_2 + y_2)| + |(x_1 - y_1) - (x_2 - y_2)|
$$
This still doesn't give the correct decomposition. 

The correct derivation is:
$$
\max\{|x_1 - x_2| + |y_1 - y_2|\} = \max\left(\max(x_1 + y_1) - \min(x_2 + y_2), \max(x_1 - y_1) - \min(x_2 - y_2)\right)
$$

This is indeed what the code is computing.

So the code is correct. What I need to check is whether they are computing the differences correctly.

Let me determine whether there's a bug in actually computing the four differences.

The derived formula should be:
$\max\left(\max(x_1 + y_1) - \min(x_2 + y_2), \max(x_1 - y_1) - \min(x_2 - y_2)\right)$

But we can also consider the reverse:
1. L chooses to maximize $x + y$, R to minimize $x + y$ -> difference = $\max_L(x + y) - \min_R(x + y)$
2. L chooses to maximize $x - y$, R to minimize $x - y$ -> difference = $\max_L(x - y) - \min_R(x - y)$

That's what the code does.

So the algorithm is correct, and the code should correctly compute the answer.

Let me just recheck the implementation one more time - there's a critical bug. 

In the code:
```cpp
max_x_plus_y_R = max(max_x_plus_y_R, x_plus_y);
min_x_plus_y_R = min(min_x_plus_y_R, x_plus_y);
```

And then:
```cpp
long long diff1 = max_x_plus_y_L - min_x_plus_y_R; 
long long diff2 = max_x_minus_y_L - min_x_minus_y_R;
long long diff3 = max_x_plus_y_R - min_x_plus_y_L;
long long diff4 = max_x_minus_y_R - min_x_minus_y_L;
```

The issue is here. 

In the first two differences:
- diff1 = max_x_plus_y_L - min_x_plus_y_R -> maximum x+y of L minus minimum x+y of R
- diff2 = max_x_minus_y_L - min_x_minus_y_R -> maximum x-y of L minus minimum x-y of R

In the last two:
- diff3 = max_x_plus_y_R - min_x_plus_y_L -> maximum x+y of R minus minimum x+y of L
- diff4 = max_x_minus_y_R - min_x_minus_y_L -> maximum x-y of R minus minimum x-y of L

This is correct.

The code seems to be correct in principle. But the trace shows it's working correctly. Let's double-check what would happen if we're wrong about the sample interpretation.

The sample input:
2 2
1 3
-3 2
2 1
4 2

L: (1,3), (-3,2)
R: (2,1), (4,2)

L's x+y = [4, -1]
L's x-y = [-2, -5]
R's x+y = [3, 6]
R's x-y = [1, 2]

max_x_plus_y_L = 4, min_x_plus_y_L = -1
max_x_minus_y_L = -2, min_x_minus_y_L = -5
max_x_plus_y_R = 6, min_x_plus_y_R = 3
max_x_minus_y_R = 2, min_x_minus_y_R = 1

diff1 = 4 - 3 = 1
diff2 = -2 - 1 = -3
diff3 = 6 - (-1) = 7
diff4 = 2 - (-5) = 7

max = 7. (right)

So no error in the implementation. The solution is correct based on the mathematical derivation.

error:
The code is actually correct. The reference reasoning and the implementation both correctly derive and apply the algorithm to maximize Manhattan distance. The sample trace validates the correctness.

fix:
No fix needed. The code is correct.

signal: PASS

\end{lstlisting}
\end{promptbox}

\begin{promptbox}[solution-final.cpp]

\begin{lstlisting}
#include <bits/stdc++.h>
using namespace std;

int main() {
    ios::sync_with_stdio(false);
    cin.tie(0);

    long long n, m;
    cin >> n >> m;

    long long max_x_plus_y_L = LLONG_MIN, min_x_plus_y_L = LLONG_MAX;
    long long max_x_minus_y_L = LLONG_MIN, min_x_minus_y_L = LLONG_MAX;
    long long max_x_plus_y_R = LLONG_MIN, min_x_plus_y_R = LLONG_MAX;
    long long max_x_minus_y_R = LLONG_MIN, min_x_minus_y_R = LLONG_MAX;

    // Process L's vectors
    for (long long i = 0; i < n; i++) {
        long long a, b;
        cin >> a >> b;
        long long x_plus_y = a + b;
        long long x_minus_y = a - b;
        max_x_plus_y_L = max(max_x_plus_y_L, x_plus_y);
        min_x_plus_y_L = min(min_x_plus_y_L, x_plus_y);
        max_x_minus_y_L = max(max_x_minus_y_L, x_minus_y);
        min_x_minus_y_L = min(min_x_minus_y_L, x_minus_y);
    }

    // Process R's vectors
    for (long long i = 0; i < m; i++) {
        long long c, d;
        cin >> c >> d;
        long long x_plus_y = c + d;
        long long x_minus_y = c - d;
        max_x_plus_y_R = max(max_x_plus_y_R, x_plus_y);
        min_x_plus_y_R = min(min_x_plus_y_R, x_plus_y);
        max_x_minus_y_R = max(max_x_minus_y_R, x_minus_y);
        min_x_minus_y_R = min(min_x_minus_y_R, x_minus_y);
    }

    // Compute all possible differences and find the maximum
    long long diff1 = max_x_plus_y_L - min_x_plus_y_R;
    long long diff2 = max_x_minus_y_L - min_x_minus_y_R;
    long long diff3 = max_x_plus_y_R - min_x_plus_y_L;
    long long diff4 = max_x_minus_y_R - min_x_minus_y_L;

    long long result = max({diff1, diff2, diff3, diff4});

    cout << result << "\n";

    return 0;
}

\end{lstlisting}
\end{promptbox}

\section{An Example of a Misunderstanding}
\label{misunder}

\begin{promptbox}[Agent5-analysis-1.txt]

\begin{lstlisting}
analyze:
Let me trace through the sample input and output to understand the issue:

Input:
```
4 4
play
the
pubg
game
1 game
1 pubg
2
2
```

Expected output: 2, 0

Let me trace what should happen:
1. Initially have 4 strings: "play", "the", "pubg", "game"
2. Operation 1: add "game" - but "game" already exists, so set size remains 4
3. Operation 1: add "pubg" - but "pubg" already exists, so set size remains 4  
4. Operation 2: output size = 4
5. Operation 2: output size = 4

But the expected output is "2" and "0", which suggests that after deduplication, there should only be 2 unique strings, and then 0.

Wait, I think I misunderstood the problem. Let me re-read it carefully.

Actually, the problem says "Remove duplicate strings from a string list and output the number of strings removed." which means "deduplicate the string table and output the number of strings that were removed".

So operation 2 should output the count of duplicate strings that were removed during deduplication, not the final count of unique strings.

However, looking at the sample more carefully:
- Initial strings: play, the, pubg, game (4 unique)
- After adding game (already exists) and pubg (already exists), we still have 4 unique strings
- Operation 2 outputs 2 - this doesn't make sense if it's about duplicates removed

Let me reconsider the problem description again...

Looking at the explanation: "For all input strings with a length <= 40" - the strings are limited to 40 chars.
The problem says "Little T is a god, so he can modify this table at will." (Little T is a god, so he can modify the table arbitrarily) and "He tried every possible way to make these characters different." (he wants to make the characters different).

No, I think the operation 2 is indeed to output the count of unique strings, not duplicates removed. The issue might be how I'm reading input or the logic.

Let me trace the expected behavior more carefully:
- Initial: 4 strings in table (play, the, pubg, game)
- Operation 1: add "game" - no change, still 4 strings
- Operation 1: add "pubg" - no change, still 4 strings  
- Operation 2: output 2? This would mean there are 2 unique strings

This doesn't make sense with the current understanding. Let me think differently.

Wait - I see the issue. The problem says "Because Little T is a god, he can modify this table at will; because Little T is an obsessive-compulsive, he tries every possible way to make these characters different." and the operation 2 is "deduplication".

Let me simulate this again:
1. Initial strings: play, the, pubg, game (4 different strings)
2. Operation 1: add "game" -> set becomes {play, the, pubg, game} (still 4)  
3. Operation 1: add "pubg" -> set becomes {play, the, pubg, game} (still 4)
4. Operation 2: output size = 4  

But expected is 2. Let me think about what "deduplication" means in this context.

Looking at it yet again - maybe "deduplicate" in operation 2 means we actually apply deduplication to the current state, and output how many duplicates were removed. But that's not what the problem seems to say.

Let me re-read the exact problem description.

Looking at the example again:
- Input: 4 4 followed by 4 strings: play, the, pubg, game  
- Then 1 game (add game)
- Then 1 pubg (add pubg) 
- Then 2 (deduplication and count)

The sample explanation says "100% 1<=n,Q<=5*10^5", but this seems to be a typo or formatting issue in the problem description. 

Actually, let me trace what the output suggests:
If the answer is "2", "0", and we have 4 strings initially, with 2 additions, maybe it's not about counting unique strings but something else.

Let me interpret it as: operation 2 is to output the current count of unique strings after performing deduplication.

Initial is 4: play, the, pubg, game. All unique.
After operation 1 (add game): still 4 unique (game was already there)
After operation 1 (add pubg): still 4 unique (pubg was already there)
So operation 2: should output 4, not 2.

I must be misunderstanding the problem. Let me read it again...

Maybe it's a different interpretation - what if the 2nd operation means we're deduplicating something from the input and then showing the count of items that were removed?

Actually, looking more carefully at the sample input and the way it's processed, and the fact that we get "2", "0", I suspect there's an issue in reading the input or there's something subtle about the problem.

But the core logic seems correct to me. Let me simulate the code exactly as written. 

Actually wait, there's a key misunderstanding. In the sample:
4 4 
play
the
pubg  
game
1 game
1 pubg
2
2

The 4 operations after the initial strings are:
1. Add game
2. Add pubg  
3. Operation 2 (output unique count)
4. Operation 2 (output unique count)

So after adding game (already present) and pubg (already present), we have 4 unique strings.
So output should be 4, 4.

This still contradicts the expected output of 2, 0.

Oh wait, there's something wrong with how I'm interpreting the input.

Let me check if operation 2 is really about how many strings were removed during deduplication, or if it's about the current set size.

Maybe the problem is that I'm supposed to deduplicate the entire table and output removed count. But this doesn't make sense with the sample.

Let me just debug by tracing the exact input once more.

Actually, I see a mistake in my understanding. Maybe after the "1 game" operation, we are supposed to deduplicate? No, that's not what operation 2 means.

Let me trace by hand what the code actually does:
1. Read n=4, Q=4
2. Read 4 strings: play, the, pubg, game and insert into set
3. Operation 1: read op=1, read s="game", insert into set (no change since it's already there)  
4. Operation 1: read op=1, read s="pubg", insert into set (no change)  
5. Operation 2: read op=2, output set.size() = 4

The issue is that we're inserting a string that already exists, and that doesn't change the size of the set.

What if there's an issue with reading? Maybe the way I'm reading lines with getline is wrong?

Let me re-read the code.

I notice that after reading n and Q, I do `cin.ignore();` which consumes the newline. But after that, when reading the 4 initial strings, I use `getline`, which should work. For the operations, I read op with `cin >> op` and then `cin.ignore();`.

This should work fine.

Actually, let me just run through what the code does with the exact case and see if it matches sample.

But I think maybe the issue is that the first operation is numbered as 1, and I should be checking for op == 1 and op == 2 correctly.

The problem that I see now is: in the output expected "2", "0", it seems like the final answer should be 2 and 0. But with the current logic, the set should have 4 elements at the end.

So either:
1. I'm misinterpreting the problem
2. The code has a bug

Looking at it again carefully, I think the key is checking the line-by-line processing of input.

The crucial lines are:
1. Line 1: 4 4
2. Lines 2-5: play, the, pubg, game (4 initial strings)
3. Line 6: 1 game (insert game)  
4. Line 7: 1 pubg (insert pubg)
5. Line 8: 2 (output count)
6. Line 9: 2 (output count)

But this is really confusing because if initial strings are 4 different ones, and then we add 2 more that already exist, we still have 4 unique strings.

I think I'm missing something fundamental or the problem interpretation must be different.

Looking at the sample once more:
Initial 4 strings: play, the, pubg, game  
Operation 1: add "game" - now we have play, the, pubg, game (still 4)
Operation 1: add "pubg" - now we have play, the, pubg, game (still 4)  
Operation 2: output 2 (should be 4)
Operation 2: output 0 (should be 4)

Maybe there's a bug in the code, but I don't see it yet. The logic is straightforward.

Wait - could it be that I'm not handling the getline correctly? In C++, when you mix `cin >>` with `getline`, there are issues with newlines.

Looking more carefully at the code:
1. `cin >> n >> Q;` 
2. `cin.ignore();` - consumes newline after n Q
3. `getline(cin, s)` - reads strings from stdin

This should be correct.

But wait - there is a subtle problem. After reading the first 4 strings, we have an issue with the newlines in the input that would be read by subsequent `getline` calls. But it's not the root cause.

Let me now check that I actually understand what should happen. 

Re-reading: "Operation 2: give the string table deduplication, output the number of strings that were removed."

So operation 2 outputs the number of removed strings during deduplication.

If we start with 4 strings (all unique), and do operation 2, it should output 0 (no duplicates removed).

The sample says we output 2, 0 - so this interpretation must still not be correct.

I think I should check what the problem really means by "deduplication". Maybe there's a different meaning.

Looking at it another way: maybe "deduplication" is not about the current set size, but something else.

Let me try to think of it simply: what if the set is being emptied or something?

No wait - if it's 2 then 0, and the input is 4 4, maybe the issue is that we have to process a different set of operations or the initial strings are different?

Wait, I did misinterpret the input format. If we have:
```
4 4
play
the
pubg
game
1 game
1 pubg
2
2
```

This means:
- n = 4, Q = 4
- 4 strings: play, the, pubg, game
- Then 4 operations:
  - 1 game
  - 1 pubg
  - 2
  - 2

But there's still an issue with the expected output.

Let me simplify and just focus on one key issue: the exact handling of input with `getline`. Actually, the code looks correct.

I suspect the issue is much simpler - the expected output is wrong or I am misunderstanding something very basic.

Let me start over with the correct understanding of the sample and work with a much simpler view.

Actually, re-reading more carefully: maybe the issue is that after reading the initial 4 strings, we should deduplicate those initial strings and then the final count is what we care about?

No, that doesn't fit.

The most likely issue is that I'm reading the wrong input or have a logic error. But the code seems correct. 

Let me simulate it one more time with exact expectations against the sample:

1. n=4, Q=4
2. Read 4 strings: play, the, pubg, game
3. Operation 1: add game (already exists) -> size = 4
4. Operation 1: add pubg (already exists) -> size = 4  
5. Operation 2: output 4 
6. Operation 2: output 4

The output should be 4, 4. But expected is 2, 0.

There must be a mistake in how I read input, or a misunderstanding in the problem logic.

The only possible interpretation that makes sense is that I misunderstood what "operation 2" does. 

Let me re-read everything.

I think the problem statement is asking for a count of how many items were removed in the deduplication process during operation 2.

But let's read the precise wording in problem statement:
"Remove duplicate strings from a string list and output the number of strings removed."

Looking at the output again, it says "2", "0".

Let's say:
- Initial: play, the, pubg, game (4 unique)  
- After operation 1 (add "game"): set = {play, the, pubg, game} (size = 4) 
- After operation 1 (add "pubg"): set = {play, the, pubg, game} (size = 4)
- Operation 2: count of duplicates removed = 0, but output is 2?

OR maybe I am supposed to re-deduplicate the whole set?

But that still doesn't make sense.

I will now examine the code for possible input reading issues.

The main potential issue I see:
- `cin.ignore()` after `cin >> n >> Q`  
- `cin.ignore()` after `cin >> op` 
- Using `getline` for strings

This should work. Let me just make sure there is no leading or trailing newline issue in the actual input handling.

In C++, mixing `cin >>` and `getline` is tricky if not done carefully. However, the standard approach in the code is correct.

Let me run through it once more to be 100% confident, even though logically it should be correct.

The only possible bug I can see is around input handling with mixed cin and getline.

Looking more carefully at the input, there are no extra newlines expected. The problem states that empty lines should be ignored.

Actually wait, let's recheck - what if there's a subtle difference in how I'm reading lines?

No, the logic seems fine. 

However, let me assume the simplest (but wrong) interpretation that the output is indeed 2, 0 and see what would cause that.

What if the initial count of unique strings is somehow 2? For example, if some strings were identical in the input.

But that's not the case in the sample. 

Let me remake the code with some debug output just to see what's happening. But since I can't do that in this format, let me reason one more time.

I think I will go with the assumption that there might be an input reading issue, and focus on fixing the reading part.

Actually, let me just think of a simple solution to recheck:

If operation 2 is to output the count of unique strings currently, then:
1. Initial: set = {play, the, pubg, game} -> size = 4
2. Add game -> set = {play, the, pubg, game} -> size = 4
3. Add pubg -> set = {play, the, pubg, game} -> size = 4

Output should be 4, 4

But expected is 2, 0.

The only logical possibility that makes the expected output 2, 0:

If there were only 2 unique strings in the beginning, or there were 2 "duplicates" removed, or something else...

Wait, maybe there is a very subtle issue with input. Let's suppose that there's an issue with how I read the input. Maybe I'm missing something about how to read strings in the context of the problem or there are whitespaces I'm not handling correctly.

Maybe the problem is that input strings can have extra spaces, and I should trim them. 

But that's not mentioned, and it's unlikely to affect just one test case.

I'll proceed with the assumption that there's a subtle problem in how I read input or there is a change in the problem that I missed.

error:
The code has an issue with input reading for operation 1. When we read `cin >> op` followed by `cin.ignore()`, then read the next string with `getline(cin, s)`, there might be an issue with how the operations are being parsed. However, looking more carefully, this is standard C++ practice for mixing `cin >>` and `getline`. 

But the most likely root cause is that I'm not properly handling the input parsing of the operation strings, which might be causing strings to be misread or an incorrect set size. 

The most direct fix is to modify the input reading logic to ensure we correctly parse the strings in all cases, especially handling potential line and whitespace issues.

Fix:
The issue is in the input handling. The problem is that we're using `cin.ignore()` after reading operations, but the actual input structure might have subtle newline or whitespace issues that cause misreading. The fix is to be more explicit about reading and parsing the input.

```cpp
#include <bits/stdc++.h>
using namespace std;

int main() {
    ios::sync_with_stdio(false);
    cin.tie(0);
    
    int n, Q;
    cin >> n >> Q;
    cin.ignore(); // Ignore the newline after reading n and Q
    
    unordered_set<string> strings;
    
    // Read initial n strings
    for (int i = 0; i < n; i++) {
        string s;
        getline(cin, s);
        if (!s.empty()) {
            strings.insert(s);
        }
    }
    
    // Process Q operations
    for (int i = 0; i < Q; i++) {
        string line;
        getline(cin, line);
        if (line.empty()) {
            i--; // Skip empty lines
            continue;
        }
        
        stringstream ss(line);
        int op;
        ss >> op;
        
        if (op == 1) {
            string s;
            getline(cin, s);
            if (!s.empty()) {
                strings.insert(s);
            }
        } else if (op == 2) {
            cout << strings.size() << "\n";
        }
    }
    
    return 0;
}
```

However, the above change is still problematic because we're skipping lines but then reading another line which changes the expected behavior. 

Let me provide a cleaner fix that preserves the original logic while ensuring correct behavior:

```cpp
#include <bits/stdc++.h>
using namespace std;

int main() {
    ios::sync_with_stdio(false);
    cin.tie(0);
    
    int n, Q;
    cin >> n >> Q;
    cin.ignore(); // Ignore the newline after reading n and Q
    
    unordered_set<string> strings;
    
    // Read initial n strings
    for (int i = 0; i < n; i++) {
        string s;
        getline(cin, s);
        if (!s.empty()) {
            strings.insert(s);
        }
    }
    
    // Process Q operations
    for (int i = 0; i < Q; i++) {
        string line;
        getline(cin, line);
        
        // Skip empty lines
        if (line.empty()) {
            i--; // Adjust counter to re-read this operation
            continue;
        }
        
        stringstream ss(line);
        int op;
        ss >> op;
        
        if (op == 1) {
            string s;
            getline(cin, s);
            if (!s.empty()) {
                strings.insert(s);
            }
        } else if (op == 2) {
            cout << strings.size() << "\n";
        }
    }
    
    return 0;
}
```

Actually wait, a better approach is to just follow the original logic but make sure it handles the sample properly. Looking at the problem more carefully, I believe there's no bug in the logic but the expected output in the sample might be incorrect or I'm misinterpreting.

But since I must provide a fix based on what's failing, and given the nature of the problem, the most likely fix is in handling the input properly:

signal: FIX

\end{lstlisting}
\end{promptbox}



\end{document}